\ifdefined\pdfoutput\pdfoutput=1\fi
\documentclass[10pt,a4paper,twocolumn]{article}
\usepackage[text={174mm,258mm},columnsep=12pt,centering]{geometry}
\usepackage{times}
\usepackage[scaled=.95]{helvet}
\usepackage{graphicx}
\usepackage{amsmath,amssymb,amsfonts}
\IfFileExists{dsfont.sty}{\usepackage{dsfont}}{%
  \DeclareRobustCommand{\mathds}{\mathbf}%
}
\usepackage{float,multirow,booktabs}
\usepackage{stfloats}
\usepackage[nospread,noshrink]{flushend}
\usepackage[table,xcdraw,dvipsnames]{xcolor}
\usepackage{caption}
\usepackage[authoryear,round,sort]{natbib}
\usepackage{placeins}
\usepackage[colorlinks,bookmarksopen,bookmarksnumbered,
  linkcolor=blue,citecolor=blue,urlcolor=blue]{hyperref}
\hypersetup{
  pdftitle={NeuSE: Neural SE(3)-equivariant embedding for long-term object-based simultaneous localization and mapping},
  pdfauthor={Jiahui Fu, Yilun Du, Kurran Singh, Joshua B. Tenenbaum, John J. Leonard},
  pdfsubject={Object-based simultaneous localization and mapping},
  pdfkeywords={SLAM, object representation, perception, change detection, long-term spatial understanding}
}
\bibpunct{(}{)}{;}{a}{,}{,}
\makeatletter
\renewcommand{\footnoterule}{%
  \kern-3\p@
  \hrule\@width\columnwidth
  \kern2.6\p@}
\renewcommand{\section}{\@startsection{section}{1}{\z@}%
  {-8pt plus -2pt minus -1pt}{5pt}{\normalfont\large\bfseries}}
\renewcommand{\subsection}{\@startsection{subsection}{2}{\z@}%
  {-6pt plus -1pt minus -1pt}{3pt}{\normalfont\normalsize\bfseries}}
\renewcommand{\subsubsection}{\@startsection{subsubsection}{3}{\z@}%
  {-5pt plus -1pt minus -1pt}{-0.5em}{\normalfont\normalsize\itshape}}
\renewcommand{\paragraph}{\@startsection{paragraph}{4}{\z@}%
  {-4pt plus -1pt minus -1pt}{-0.5em}{\normalfont\normalsize\itshape}}
\renewcommand{\@maketitle}{%
  {\noindent\sffamily\bfseries\fontsize{18}{20}\selectfont\raggedright
   \@title\par}
  \vspace{14pt}
  {\noindent\sffamily\bfseries\fontsize{12}{14}\selectfont\raggedright
   \@author\par}
  \vspace{22pt}
}
\makeatother
\AtBeginDocument{%
  \setlength{\abovedisplayskip}{5pt plus 1pt minus 1pt}%
  \setlength{\belowdisplayskip}{5pt plus 1pt minus 1pt}%
  \setlength{\abovedisplayshortskip}{2pt plus 1pt}%
  \setlength{\belowdisplayshortskip}{4pt plus 1pt minus 1pt}%
}

\definecolor{pink}{cmyk}{0, 0.7808, 0.4429, 0.1412}
\definecolor{blue}{RGB}{96,118,193}
\definecolor{green}{RGB}{34,139,34}
\definecolor{orange}{RGB}{255,141,65}
\definecolor{dark_blue}{RGB}{135,164,185}
\definecolor{darkgreen}{RGB}{106,164,139}
\definecolor{darkpink}{RGB}{218,112,214}
\definecolor{grey}{RGB}{100,100,100}
\definecolor{yellow}{RGB}{255,219,111}
\definecolor{purple}{RGB}{147,112,219}

\DeclareMathOperator*{\argmax}{arg\,max}
\newcommand{\sethree}{$\text{SE}(3)$}

\newcommand{\bsethree}{$\text{\textbf{S}E}(3)$}
\newcommand{\sothree}{$\text{SO}(3)$}

\newcommand\BibTeX{{\rmfamily B\kern-.05em \textsc{i\kern-.025em b}\kern-.08em
T\kern-.1667em\lower.7ex\hbox{E}\kern-.125emX}}

\begin{document}

\title{NeuSE: Neural SE(3)-equivariant embedding
for long-term object-based simultaneous localization and mapping}
\date{}

\author{%
Jiahui Fu, Yilun Du, Kurran Singh,
Joshua B. Tenenbaum, and John J. Leonard}

\twocolumn[%
\maketitle
\begin{minipage}{\textwidth}
\begin{abstract}
We present NeuSE, a novel \textbf{Neu}ral \bsethree-Equivariant \textbf{E}mbedding for objects,  and illustrate how it supports object-based Simultaneous Localization and Mapping (SLAM) for consistent spatial understanding with long-term scene changes.  NeuSE is a set of latent object embeddings created from partial object observations. It serves as a compact point cloud surrogate for complete object models, encoding the full shape, scale,  and transform information about an object. In addition, the inferred latent code is both \sethree~and scale equivariant, enabling strong generalization to objects of both unseen sizes and different \sethree~poses. This makes NeuSE particularly effective in real-world scenarios where objects may vary in size or spatial configuration. With NeuSE, relative frame transforms can be \emph{directly} derived from inferred latent codes. Our proposed SLAM paradigm, using NeuSE for object shape, size, and pose characterization, can operate independently or in conjunction with typical SLAM systems. It directly infers {\sethree} camera pose constraints that are compatible with general SLAM pose graph optimization, while maintaining a lightweight, object-centric map that adapts to real-world changes. Our evaluation is conducted on synthetic and real-world sequences with changes in both controlled and uncontrolled settings, featuring multi-category objects of various shapes and sizes. Our approach demonstrates improved localization capability and change-aware mapping consistency when working either independently or as a complement to common SLAM pipelines. 
Additional results are presented in the \href{https://www.youtube.com/watch?v=fZ4_2P1fQBQ}{\mbox{supplementary video}}.
\end{abstract}

\noindent\textbf{Keywords:} Simultaneous localization and mapping, object representation,
perception, change detection, long-term spatial understanding.
\end{minipage}
\vspace{0.4em}
]

\begingroup
\renewcommand{\thefootnote}{}
\footnotetext[0]{All authors are with the MIT Computer Science and Artificial
Intelligence Laboratory, Cambridge, MA 02139, USA.\par
\textbf{Corresponding author:} Jiahui Fu,
\href{mailto:jiahuifu@mit.edu}{jiahuifu@mit.edu}.}
\endgroup

\section{Introduction}
The ability to conduct consistent object-level reasoning is crucial for many high-level robotic tasks, especially those involving repetitive traversal in the same environment, such as household cleaning and object retrieval. In a constantly evolving world, robots are expected to accurately locate themselves and their target while keeping an updated map of the environment, ensuring that a specific ``blue coffee mug'' can always be retrieved regardless of its location since its last use.

\begin{figure}[!htb]
    \centering
    \includegraphics[width=0.95\linewidth]{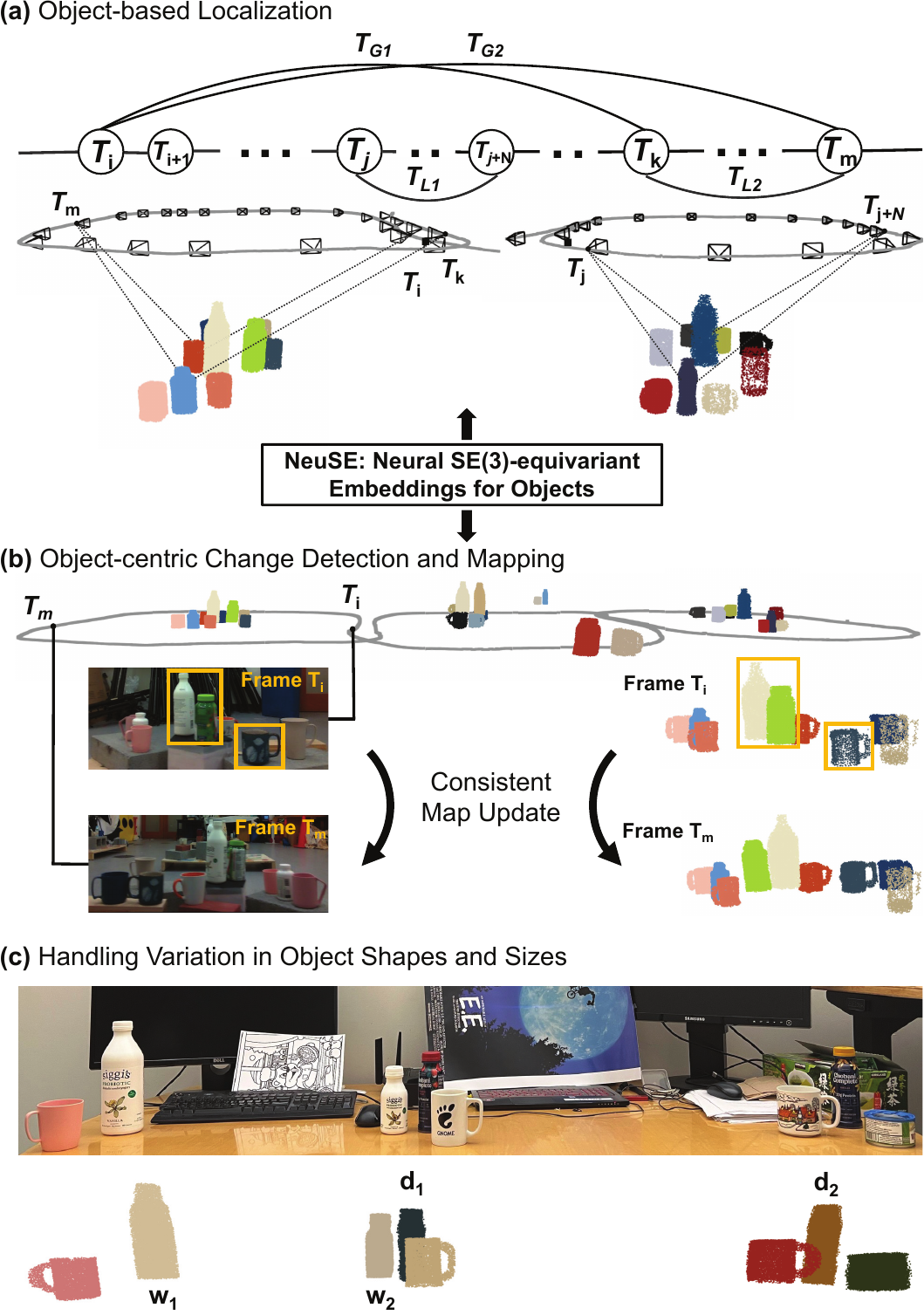}
    \caption{ \textbf{Schematic of consistent spatial understanding with NeuSE}. An object-centric map of multiple object categories constructed from real-world data is shown for illustration.
\textbf{(a)} NeuSE acts as a compact point cloud surrogate for objects, encoding full object shapes and sizes and transforming \sethree-equivariantly with the object in the physical world. Latent codes of bottles and mugs from different frames can be effectively associated (dashed line) for \emph{direct} computation of inter-frame transforms, which are then added to constrain camera pose (${T}_i$) optimization both locally ($T_{Li}$) and globally ($T_{Gi}$). \textbf{(b)} The system performs change-aware object-level mapping, where changed objects (highlighted in \textcolor{orange}{orange}) are updated alongside unchanged ones with full shape reconstructions in the object-centric map. \textbf{(c)}  With \sethree~and scale equivariance, NeuSE effectively handles multiple instances of identically shaped objects (dark bottles $d_1$ and $d_2$), as well as objects with diverse shapes (bottles, mugs, and cans) and sizes (big white bottle $w_1$ and small white bottle $w_2$).}
    \label{overview}
\end{figure}
Traditional Simultaneous Localization and Mapping (SLAM) approaches~\citep{campos2021orb, engel2014lsd, 5336495} see the world through a static set of low-level geometric primitives extracted from observations, making them less amenable to human-like reasoning about the world. In the absence of semantic information, these unordered collections of points, lines, or planes are not completely compatible with object-level interpretation, making them susceptible to false correspondence matches when faced with scene changes over time.

As the world changes and operates under the minimal unit of objects, objects serve as an intuitive source for assisting localization and an object-centric map can act as a lightweight and flexible reflection of the latest environment layout. To facilitate the communication between objects and typical SLAM systems, previous works have experimented with various object representations to guide back-end optimization, ranging from pre-defined object model libraries~\citep{salas2013slam++,7487378}, semantic segmentation masks~\citep{mccormac2017semanticfusion,runz2018maskfusion,mccormac2018fusion++,xu2019mid}, to parameterized geometry~\citep{8440105,hosseinzadeh2019real,yang2019cubeslam, wang2024icra}. However, they are confined to either a limited number of objects, or a loss of geometric details due to partial reconstruction or simplification of object shapes.  

Recently, neural implicit representations have been  introduced~\citep{sucar2020nodeslam,wang2021dsp,zhu2022nice,sucar2021imap, rosinol2023nerf}  to SLAM as object or scene representations, working with probabilistic rendering loss to help constrain camera localization. However,  since both object and scene representations are not directly interpretable, an expensive rendering procedure is necessary to convert representations to explicit reconstructions, which can then be iteratively optimized to reflect {\sethree} camera pose constraints embedded within observations. This incurs extra training and computation overhead and these camera constraints are sensitive to starting pose initializations.

To more effectively leverage the shape description power of neural representations and bypass the undesirable iterative refinement, we depart from the dominant ``\emph{render-optimize}'' convention in previous works by directly embedding geometric constraints into object representations. We introduce NeuSE, a novel category-level \textbf{Neu}ral \bsethree-Equivariant \textbf{E}mbedding for objects.  Given a partial observation of an object, NeuSE constructs a latent code corresponding to a canonical point cloud that represents the shape and transform of the full object. This latent point cloud captures the object's translation and rotation, transforming equivariantly with the object when an \sethree~transform is applied to it. Additionally, it embeds the object's size, scaling equivariantly with changes in the observed size (both types of equivariance are illustrated in \autoref{se3_example}). Finally, it encodes the full shape of the object through the rotation-invariant portion of the latent point cloud.

By constructing this latent representation of an object, we can directly distinguish  between objects by their latent shapes, scales, and relative spatial relations. Simultaneously, the \sethree~equivariance of this latent representation allows relative frame transformations to be \emph{directly} computed from the corresponding latent codes of an object when it is observed in different frames. To account for the pose ambiguity arising from symmetrical geometry, we further train NeuSE's latent point cloud to conform to the object's geometric ambiguity. In this way, working with NeuSE's latent canonical point cloud of an object is akin to working with the full object model, only with operations applied to a compact latent point cloud surrogate with known correspondences, which can be directly inferred from partial observations. Furthermore, the learned \sethree~and scale equivariance in NeuSE allows inferred latent codes to generalize effectively to objects with distinct poses, shapes, and sizes beyond those seen during training, which helps promote NeuSE's generalization and adaptability for robust spatial understanding.

In this paper, we present NeuSE and further demonstrate how it supports object SLAM targeting spatial understanding with long-term scene inconsistency (see \autoref{overview}). By using NeuSE for object shape, scale, and pose characterization,  we unify the representations of major SLAM modules, for example, data association and pose constraint derivation, around one versatile latent code. Our proposed approach can either work as a standalone system or complement common SLAM systems by \emph{directly} inferring {\sethree}  camera pose constraints compatible with general SLAM pose graph optimization and maintaining a lightweight, object-centric map with change-aware mapping ability.

\begin{figure}[!htb]
    \centering
    \includegraphics[width=\linewidth]{figures/2.pdf}
    \caption{\textbf{System overview}. The input to the system is a depth image and segmentation masks of objects of interest, generated from the corresponding RGB images, which together create partial object point clouds. The system’s six modules then process these partial point clouds to interact with and update object and frame information in the database. The output of the system can be retrieved from the database as camera pose estimates for input frame sequences and the latest object-centric map of the environment, consisting of full shape reconstructions of the objects of interest.}
    \label{block_diagram}
\end{figure}

We provide a system overview of the proposed NeuSE-based object SLAM pipeline in \autoref{block_diagram}, which consists of six functional modules and a database for storing object and frame information. The system takes as input both depth images and segmentation masks of objects of interest, where the masks are obtained by segmenting the corresponding RGB images. These inputs generate partial point clouds of the observed objects, serving as the sole data source for the system. The six modules perform three main functions: NeuSE extraction, NeuSE odometry generation and pose graph update, and change detection. These modules operate in sequence, continuously interacting with and updating the object and frame information in the database. The system's output, which can be retrieved from the database, includes the estimated camera poses for the current image stream and the most up-to-date object-centric map, consisting of full shape reconstructions of the objects of interest in the environment.

Our main contributions are as follows: {\bf (1)} We introduce NeuSE, a canonical latent point cloud representation that encodes the full object shape, scale, and transformation,  transforming and scaling equivariantly with the physical object.  {\bf (2)} We propose a NeuSE-based object SLAM paradigm targeting long-term scene inconsistency, enabling NeuSE-predicted object-level localization and change-aware mapping. {\bf (3)} We evaluate our approach on synthetic and real-world sequences, in both controlled and uncontrolled settings with objects of varying shapes and sizes, demonstrating improved localization performance and adaptable mapping capability when working as a standalone system or jointly with common SLAM pipelines.
\section{Related work}
\subsection{Object-based SLAM}
SLAM++~\citep{salas2013slam++} introduced object-based SLAM by incorporating camera-object constraints with objects from a predefined model database.  Attempts~\citep{mccormac2017semanticfusion,runz2018maskfusion,mccormac2018fusion++,xu2019mid}  were made to  leverage semantic segmentation for instance-level dense reconstructions. Furthermore, simple parameterized geometry, for example, ellipsoids adopted by~\citet{8440105} and ~\citet{hosseinzadeh2019real} and cuboids by~\citet{yang2019cubeslam}, were  explored to guide the joint optimization of the object shape parameters and camera poses. For environments with moving objects, \citet{strecke2019_emfusion} proposed an object-level SLAM approach that utilized local Signed Distance Function (SDF) object volumes for tracking moving objects and performing camera localization. A recent work, VOOM by \citet{wang2024icra}, described landmarks in a hierarchical manner that combined high-level objects, represented as dual quadrics, with low-level feature points to create a visual object odometry and mapping framework. The proposed pipeline enabled efficient object optimization and association, leading to improved localization accuracy and demonstrating a certain level of robustness in dynamic scenes. Efforts have also been made to integrate neural shape priors into the object SLAM pipeline. NodeSLAM~\citep{sucar2020nodeslam}  adopted a class-level optimizable object shape descriptor and used RGB-D images for joint estimation of object shapes, poses, and camera trajectory through iterative probabilistic rendering optimization.  DSP-SLAM~\citep{wang2021dsp}, on the other hand, used DeepSDF~\citep{8954065} for object representation and optimized the object code, camera poses, and sparse landmark points altogether through a similar rendering loss in RGB, stereo, or stereo+LiDAR modalities. As the rendering process is parameterized as a neural network with no interpretable meaning, both methods require iterative optimization with a proper initialization to obtain the {\sethree} transform constraint that aligns with the real-world observation.  This results in added training and computational expenses, making the adoption of neural representations a complex process.                   
\subsection{Neural implicit representations for robotics}
Neural implicit representations have emerged as a promising tool to encode the underlying 3D geometry of objects and scenes~\citep{8954065, 8953655, Ortiz:etal:iSDF2022}. Different works have explored how neural implicit representations can be used in various fields, including change detection~\citep{9981246}, localization~\citep{9712211,moreau2022lens},  SLAM~\citep{zhi2019scenecode,sucar2021imap,zhu2022nice,10160950, rosinol2023nerf}, and manipulation~\citep{ shen2022acid, li20223d, simeonov2022se, pmlr-v205-lin23c, pmlr-v205-kerr23a, 10160423, ryu2023equivariant,  khargonkar2023neuralgrasps, driess2023learning}. 

Notably, some works extended the original representation by integrating {\sothree} or {\sethree} equivariance for tasks such as reconstruction~\citep{Deng_2021_ICCV}, point cloud registration~\citep{zhu2022correspondence, lin2022coarse}, and manipulation~\citep{9812146} as well as scale equivariance~\citep{lei2023efem, yang2024equivact}.  \citet{zhu2022correspondence}  learned \sothree-equivariant features to perform correspondence-free point cloud registration, while \citet{lin2022coarse} used \sethree-equivariant representations to obtain and refine  the registration result globally and locally. \citet{9812146} learned \sethree-equivariant object representations for manipulation and estimated relative transforms through optimization. These methods target point clouds known to be associated with the same object, which can suffer from performance degradation for partially overlapped point clouds \citep{zhu2022correspondence, lin2022coarse} or require iterative refinement to recover the desired relative transform~\citep{9812146}.

In the context of  SLAM, most works, other than the object-based methods listed in the previous section, utilized scene-level neural implicit representations to be jointly optimized with camera poses. iMap~\citep{sucar2021imap} showed that a multilayer perceptron (MLP) could serve as the scene representation for real-time RGB-D SLAM. NICE-SLAM~\citep{zhu2022nice}, built on top of iMap, further introduced a hierarchical grid-based neural encoding, enabling RGB-D SLAM on a larger scale. In terms of monocular SLAM, Nerf-SLAM~\citep{rosinol2023nerf}  relied on an indirect loss for pose estimation and produced higher-quality reconstructions by supervising the radiance field with depth information. These methods, like their object-based counterparts, still require undesirable iterative optimization with photometric or depth loss for localization, while being hard to adapt to changes with the scene represented as one single code. 

Our NeuSE-based SLAM paradigm differs from prior SLAM works with neural representations by further explicitly imposing \sethree~and scale equivariance onto the vanilla neural object representations. To handle unknown data associations, in contrast to previous works on point cloud registration or manipulation with equivariant representations, we can directly use the implicit shapes and scales captured by latent representations. This allows partial point clouds to be robustly associated regardless of viewing angle differences. With additional regularization for objects with pose ambiguity, we ultimately achieve \emph{direct} inference of {\sethree} camera pose constraints from partial object representations. This eliminates the need for the computationally expensive ``render-optimize'' process and offers a lightweight yet flexible solution to object SLAM problems with long-term changes. 
\section{Category-level neural SE(3)-equivariant embedding (NeuSE) for objects}
\subsection{SE(3) and scale equivariance}
We propose to represent each object in a scene by using a corresponding \sethree-equivariant latent embedding with scale equivariance. Precisely,  given a point cloud ${\bf{P}} \in \mathbb{R}^{N \times 3}$, we represent it with a lower dimensional latent embedding (``a canonical latent point cloud'') ${\bf z} \in \mathbb{R}^{D\times 3}$, inferred using a neural network encoder $f$ so that ${\bf z} = f(\bf{P})$. The underlying latent embedding is equivariant, so that for any {\sethree} transform {\bf T}:
\begin{equation}
\label{se3eq}
    {\bf{T}z} = f(\bf{T}P),
\end{equation}
that is, the latent embedding $\bf z$ transforms equivariantly w.r.t. the point cloud $\bf{P}$. 

Simultaneously, given a uniform scaling factor $s$:
\begin{equation}
\label{s3}
    {s{\bf z}} = f(s{\bf P}),
\end{equation}
where the latent embedding ${\bf z}$ also transforms equivariantly with the scaling factor $s$.
Such equivariance allows us to treat the latent ${\bf z}$ as a processed surrogate of the point cloud ${\bf{P}}$.
\begin{figure}[!htb]
    \centering
    \includegraphics[width=\linewidth]{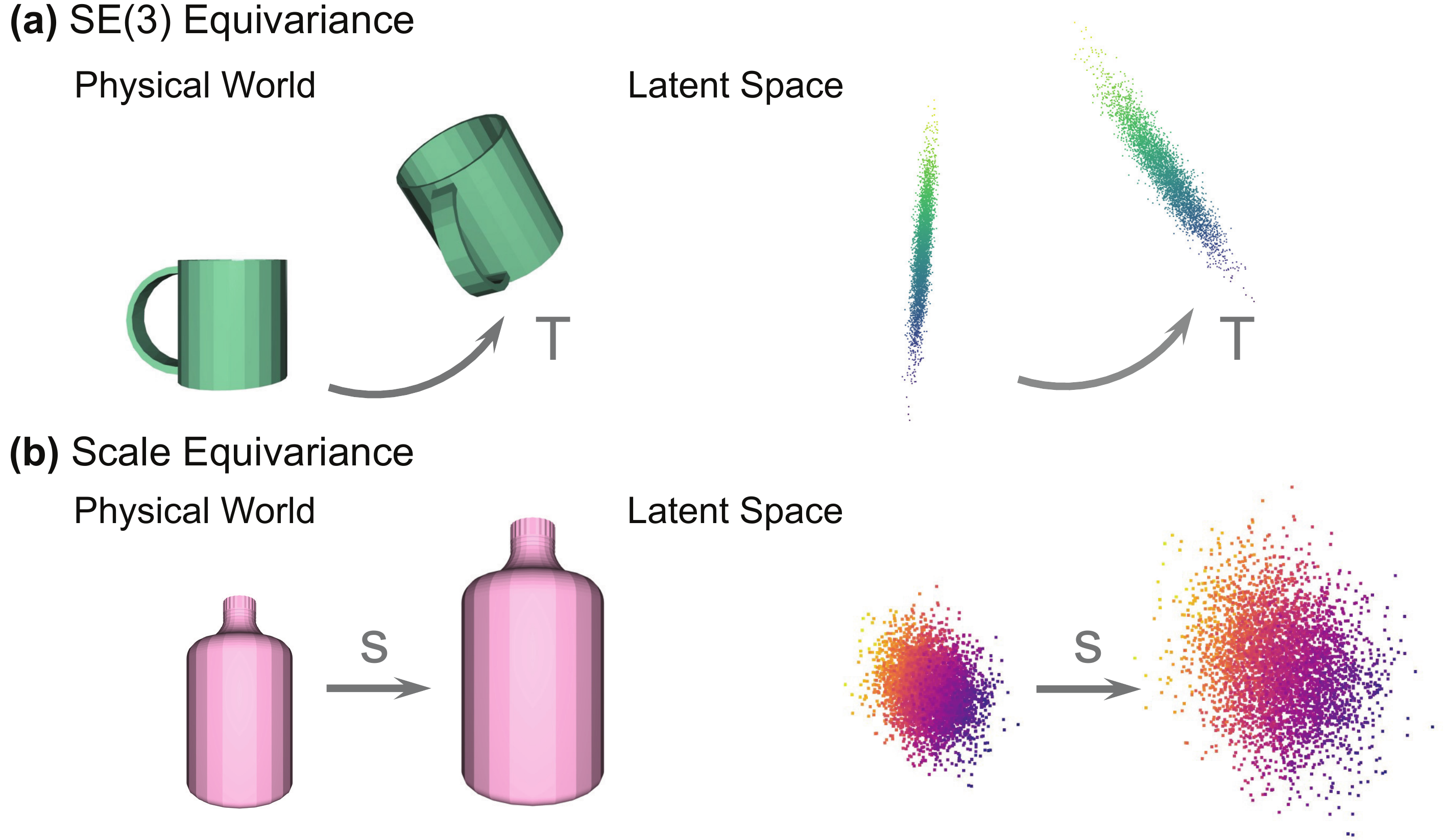}
\caption{\textbf{Illustration of \sethree~and scale equivariance.} Equivariance means that the latent representation, though not visually resembling the physical object, transforms in the latent space in the same way as the object does in the real world. \textbf{(a)} With \sethree~equivariance, the latent code undergoes the same \sethree~transform $\mathbf{T}$ applied to the \textcolor{darkgreen}{mug} in the physical world. \textbf{(b)} With scale equivariance, the latent code scales by the same factor $s$ as the \textcolor{darkpink}{bottle} in the physical world.}
    \label{se3_example}
\end{figure}

In \autoref{se3_example}, we present two examples explaining both types of equivariance, that is, how latent representations transform and scale within the latent space, corresponding to the physical transformations of objects in the real world. Although these latent embeddings do not visually resemble the actual objects, they maintain consistent behaviors in latent space, following the objects' transformations in the physical world. This consistency offers the following three benefits:

\subsubsection{Latent pose constraints.}  The underlying latent embedding space operates under the same \sethree~action as the point cloud. Thus, we may express pose constraints between matched objects directly in the latent space as opposed to the full point cloud space of objects. As the latent space is both low-dimensional and canonical, pose constraints may be more efficiently computed using the closed-form solution developed by \citet{Horn:87}. 

\subsubsection{Implicit pose representation.} The object latent code implicitly captures the underlying {\sethree} transform of an object. This circumvents the need to explicitly specify 6-DOF poses of objects when computing pose constraints, which may not always be accessible and can be ill-defined for objects with symmetrical ambiguity.

\subsubsection{Implicit shape representation.} The object latent code richly encodes the underlying shape, size, and features of an object, which enables robust data association across viewing angle disparity.

\subsection{Constructing equivariance}
To infer \sethree-equivariant latent codes with scale equivariance, NeuSE uses a {\sothree}-equivariant encoder function \citep{Deng_2021_ICCV} $f_{\theta}({\bf{P}) = z}$ that maps a partial object point cloud $\bf P$ into a global latent point cloud $\bf z$, and a decoder function $\Phi(\bf{x}, f_{\theta}(\bf{P}))$ that maps an input query point $\bf{x}$ to its predicted occupancy value according to $\bf z$:
\begin{equation}
\begin{aligned}
   & f_{\theta}({\bf{P) = z}} :  \mathbb{R}^{n\times 3}\rightarrow\mathbb{R}^{k\times 3} \\
   & \Phi({\bf{x}, f_{\theta}({\bf{P}}))} = \Phi {\bf(\bf{x}, {\bf{z}}}) : \mathbb{R}^{3}\times \mathbb{R}^{k\times 3} \rightarrow [0,1].
\end{aligned}
\label{vanilla}
\end{equation}

We modify the encoder function from \citet{Deng_2021_ICCV} to achieve scale equivariance by normalizing the cross-product encoding of the point cloud using the point cloud's norm. The subsequent vector neuron layers are then scale equivariant by construction.

By feeding $\Phi(\cdot,\cdot)$ with a query point cloud $\mathcal{X}$ obtained via uniform sampling within a large bounding box centered around $\bf P$, the full shape point cloud $\mathcal{S}$ of the object can be reconstructed from the predicted occupancy values with $ \mathcal{S} = \{{\bf x}| \Phi({\bf x},f_{\theta}({\bf x}\vert {\bf{P}}))>v_0, {\bf x}\in \mathcal{X}\}$, where $v_0$ is the threshold to mark whether a point location is occupied.

\subsection{Learning SE(3) equivariance across viewing angles} 
While equation~\eqref{se3eq} holds for identical but transformed point clouds, capturing the same object point cloud as the robot moves is nearly impossible due to constantly changing camera views and occlusion patterns, resulting in varying partial observations of the same object. To maintain consistent equivariant behaviors between the latent representation and the associated physical object, \sethree~equivariance must hold not only for one single partial point cloud from a specific view but also across the varying partial point clouds observed of the object.

To achieve this, we construct \sethree~equivariance  from partial point cloud $\bf P$ separately through rotation and translation equivariance.  

For rotation equivariance, as our encoder is rotation equivariant,  when a point cloud is rotated by $\bf R$, the inferred latent code will be equivariantly rotated by $\bf R$:
\begin{equation}
    f_{\theta}({\bf RP}) = {\bf{Rz}}, {\bf R}\in \text{SO}(3).
\label{so3equi}
\end{equation}

Since $\bf P$ is a partial observation of the complete object geometry, we treat this partial center $\bf \overline{P}$ as an initial estimate of the actual object translation so as to learn an approximately translation equivariant latent $\bf{z}$. We first infer ${\bf z}_{0}$ for the zero-centered partial point cloud $\mathbf{P}-\overline{\mathbf{P}}$. The final latent code $\bf z$ for point cloud $\bf P$ is obtained by adding back the partial center ${\bf z} = \overline{\bf P} + {\bf z}_{0}$. Hence, to infer an \sethree-equivariant $\bf z$, the final formulation of  equation~\eqref{vanilla} is accordingly written as:
\begin{equation}
\begin{aligned}
   & f_{\theta}({\bf{P}-\bf{\overline{P}}) = \bf{z}}_{0} :  \mathbb{R}^{n\times 3}\rightarrow\mathbb{R}^{k\times 3} \\
   & {\bf{z}}  ={\bf{z}}_{0} + \bf {\overline{P}}, \mathbf{z}' = \mathbf{z}-\overline{\mathbf{z}} \\
   & \Phi({\bf{x}, f_{\theta}({\bf{P}}))} = \Phi {\bf(\bf{x}-\overline{z}, {\bf{z'}}}) : \mathbb{R}^{3}\times \mathbb{R}^{k\times 3} \rightarrow [0,1],
\end{aligned}
\label{se3}
\end{equation}
where $\bf \overline{z}$ is the center of $\bf z$. The translational equivariance on $\bf z$  is imposed by training the center of ${\bf z}_{0}$ to learn the offset between $\bf \overline{P}$ and the true object center (translation).  Ultimately, for the same object observed partially with camera views ${\bf T}_1$ and ${\bf T}_2$, the \sethree~transform ${ {\bf{T}}_{1,2}=\bf(R,t)}$ between the two latent point clouds ${\bf z}_1$ and ${\bf z}_2$, which is expected to be close to ${\bf T}_2^{-1}{\bf T}_1$, can be obtained by:
\begin{equation}
   {\bf{T}}_{1,2}={\bf(R, t)} = \Psi({\bf z}_1,{\bf z}_2),
\label{horn}
\end{equation}
where $\Psi(\cdot,\cdot)$ is Horn's method~\citep{Horn:87} providing a closed-form solution of the relative {\sethree} transform between two point clouds with known correspondence.
\subsection{Dealing with pose ambiguity}
\sethree~equivariance is desirable for revealing the relative transform between the two frames where the same object is observed. However, shape symmetry can result in ambiguity in the inferred transform, causing our latent codes to be fallible when the transform selected is one of many possible candidates rather than the correct one.

To make our representations applicable to a broader range of objects, we therefore propose separate training objectives for object shapes with and without ambiguity w.r.t. the camera viewing frustum.
\subsubsection{Unambiguous objects.}  For objects without pose ambiguity (e.g., mugs with a handle), the transform $\bf (R,t)$ obtained from equation~\eqref{horn} should be unique and thus approximate the true inter-frame camera transform. We therefore simply minimize the $L_2$ distance between the estimated transform $(\hat {\bf R}_{3\times 3}, \hat {\bf t}_{3})$ and the ground truth $({\bf R}_{3\times 3}, {\bf t}_{3})$:
\begin{equation}
L_{transform}^{uab} = \vert\vert({\bf \hat{R}{R}}^T)-{\bf I}_{3\times 3}\vert\vert_F + \vert\vert {\bf{\hat{t}}} - {\bf{t}} \vert\vert_2,
\end{equation}
where $\vert\vert \cdot \vert\vert_F$ is the Frobenius norm.

\subsubsection{Ambiguous objects.} We limit ``ambiguous objects'' to objects with pose ambiguity from their shapes (e.g., upright wine bottles), but not the ones that may appear ambiguous due to occlusion (e.g., mugs with their handles obscured). 

Since ambiguous objects have multiple or infinitely many valid transforms that can meet the current observation, the exact single correct transform can never be learned. We instead wish that the derived transform will always lead to similar object shapes when transforming the object's point cloud from one frame to another. In a nutshell, we require the latent code $\bf z$ to implicitly learn the distribution of the possible transforms. 

Hence, given the full object point clouds in two frame coordinates,  ${\bf P}_{o1}$ and ${\bf P}_{o2}$ (readily available as we train fully in simulation),  we enforce that the Chamfer distance between the two point clouds should be small after aligning them with the predicted transform:
\begin{equation}
    \begin{aligned}
        L_{amb}  = &CD({\bf T}_{1,2}{\bf P}_{o,1}, {\bf{P}}_{o,2})\\
           CD({\bf P}_1,{\bf P}_2) = &\frac{1}{\vert {\bf P}_1\vert}\sum_{{\bf x} \in {\bf P}_1} \min_{\bf{y} \in {\bf P}_2} \vert \vert {\bf x-y}\vert \vert _2+ \\
        &\frac{1}{\vert {\bf P}_2\vert}\sum_{{\bf y} \in {\bf P}_2} \min_{\bf{x} \in {\bf P}_1} \vert \vert {\bf x-y}\vert \vert_2.
    \end{aligned}
\end{equation}

\begin{figure}[!htb]
    \centering
    \includegraphics[width=0.9\linewidth]{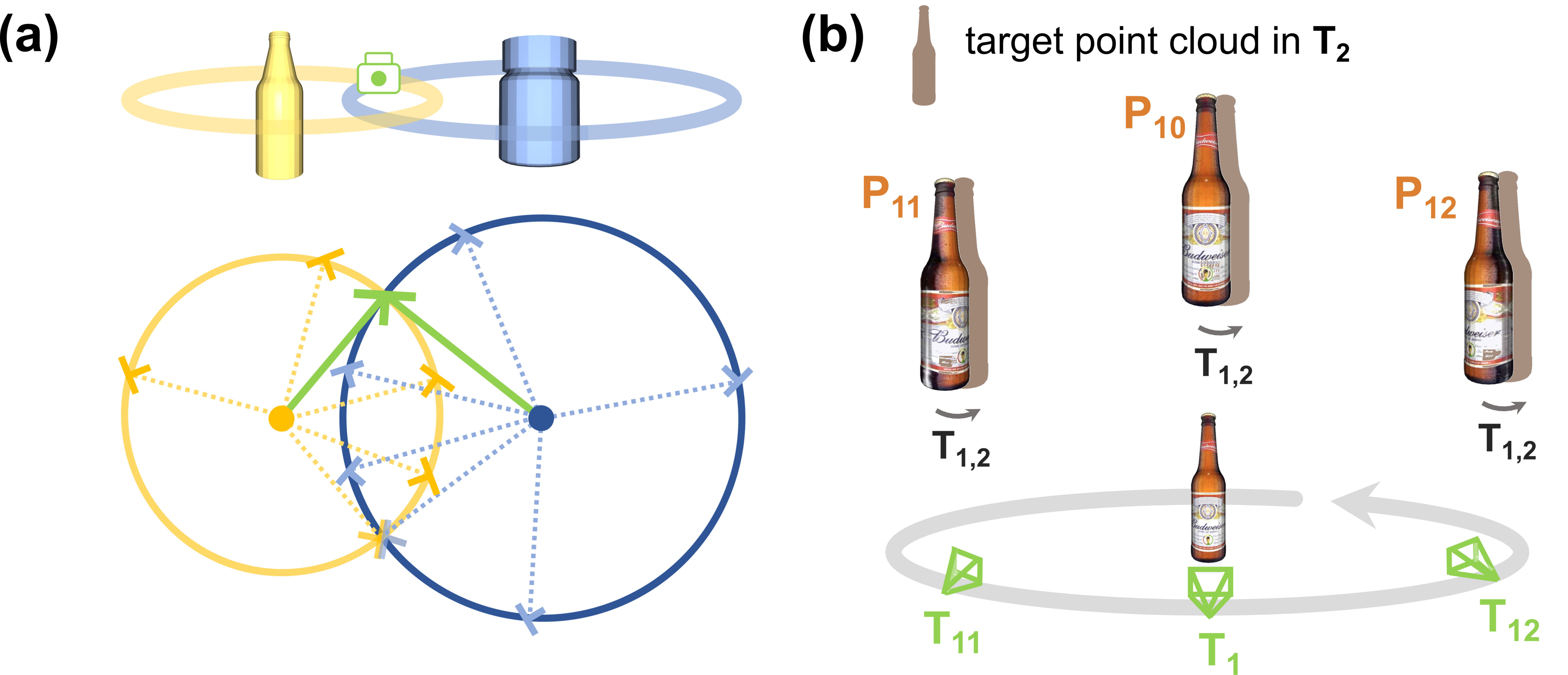}
    \caption{\textbf{(a)  Breaking pose ambiguity with covisible ambiguous objects}. Motions around a bottle's axis of symmetry result in seemingly identical observations, making it impossible to determine inter-frame transforms. However, with two covisible bottles, the intersection (\textcolor{green}{green}) of their camera pose distributions (\textcolor{yellow}{yellow} and \textcolor{blue}{blue}) for the current  observation reveals the true camera pose, where inter-frame transforms can then be determined without ambiguity. \textbf{(b)  Latent symmetry}. The canonicalized latent embedding should be invariant to camera motion (${\bf T}_{1i}$) around the object's axis of symmetry, inducing consistently small Chamfer distance between the transformed bottle (${{\bf T}_{1,2}\bf P}_{1i}$) and the \textcolor{brown}{target point cloud}.}
    \label{chamfer}
\end{figure}

We can recover the exact transform that simultaneously justifies all current object observations by intersecting the distributions of possible transforms for multiple ambiguous objects (see \autoref{chamfer}(a) for the reasoning of the base two-object case involving two bottles), or further refine the predicted one when working together with unambiguous objects. Note here we do not account for the rare degenerate case of colinear axes of symmetry for all visible objects.

Furthermore, to facilitate the learning of the underlying distribution, we further augment the original (${\bf P}_{10}$, ${\bf P}_{20}$) pair to include extra samples in the distribution. Given camera views ${\bf T}_1$ and ${\bf T}_2$, we fix ${\bf T}_2$ and generate $N$ random transforms ${\bf T}_{1i}$s that allow for camera movement around the object's axis of symmetry with observations that appear identical to those from ${\bf T}_1$ (\autoref{chamfer}(b)). The resulting $N$ object point clouds in corresponding camera frames, ${\bf P}_{1i}$s, 
should  retain similar shapes to ${\bf P}_{2}$ using the predicted transform. Hence, the ultimate training objective for ambiguous objects is:
\begin{equation}
        L_{transform}^{amb}  = \sum_{i=0}^{N} CD({\bf T}_{1,2}{\bf P}_{1i}, {\bf{P}}_{20}),
\end{equation}
where $N$ and values of ${\bf T}_i$ are determined by the type, for example, cylindrical ($360^\circ$) or cubical ($180^\circ$), of object ambiguity. Here in our experiment, for cylindrical bottles, we set $N=180$ and draw transforms from $[0^\circ,360^\circ]$ circulation around the cylindrical bottles while for rectangular objects, we directly apply the $180^\circ$ transform w.r.t. its axis of symmetry. 

Finally, the target inter-frame transform can be similarly obtained using equation~\eqref{horn}, with the two latent codes $\bf z$s formed by concatenating all corresponding ${\bf z}_i$s of covisible objects in each frame.
\subsection{Shape consistency across viewing angles}
Since ${\bf z}_{0}$ is \sothree-equivariant, its rotation invariant part, ${\bf{s}}\in \mathbb{R}^k$, encoding the object's \emph{full shape}, can then be extracted as ${\bf s}=\{s_i\}_{i=1}^{i=k}, s_i = \vert\vert ({\bf{z}}_{0})_i \vert\vert_2$, where we term $\bf s$ as the shape descriptor.

As $\bf s$ is expected to encode the full object shape, it should remain consistent across partial point clouds of the same object, regardless of  viewing angle variation. However, $\bf s$ by construction only ensures consistency for identical point clouds and their rotated counterparts, and does not inherently guarantee consistency across varying, incomplete observations of the object.

Following~\citet{9981246}, we adopt the batch-hard shape similarity loss $L_{b\_shape}$, enforcing $\bf s$ to be consistently similar across viewing angles of the same object while discriminatively far apart for different objects. 

$L_{b\_shape}$ takes the form of the triplet loss as [\emph{anchor, positives, negatives}]. To allow for a variety of viewing angle combinations during training, we populate each training batch $B$ with $M$ partial observations for each of the $N$ randomly drawn objects. Samples of the same object instance serve as mutual $\emph{anchors}$ and $\emph{positives}$, $({\bf{A}}_i$, ${\bf{P}}_i)$, with samples not from the current shape instance being the $\emph{negatives}$, ${\bf{N}}_i$. $L_{b\_shape}$ is calculated in a ``batch-hard'' fashion, that is, it only uses the most dissimilar $(\bf{A},\bf{P})$ and the most similar $(\bf{A},\bf{N})$ for each anchor to guide the training. With $D(\cdot,\cdot)$ as the cosine similarity, the final batch-hard shape similarity loss is formulated as:
\begin{equation}
\begin{aligned}
        L_{b\_shape} = \frac{1}{\vert B \vert} \sum_{i=1}^{N}\sum_{j=1}^{M}(&-\min_{k\in[1,M]}D(o_{ij},o_{ik}) 
         \\ &+\max_{m\neq i}D(o_{ij},o_{mn})),
\end{aligned}
\label{triplet}
\end{equation}
where $o_{ij}$ is the $j$th observation of object $i$ within the batch.
\subsection{Capturing relative scale} 
\label{sect:scale}
As ${\bf z}_{0}$ is scale equivariant, the $L_2$ norm of the full latent code $n = \vert\vert {\bf{z}}_{0} \vert\vert_2$ provides a relative scale of a point cloud ${\bf P}$. In particular, given a point cloud ${{\bf P}_2} = s{\bf P}$, the norm of the inferred latent code satisfies the expression $n_2 = sn$ and thus reflects the scale of the point cloud.

However, given two point clouds ${\bf P}_1$ and ${\bf P}_2$ of different shapes, the corresponding norms $n_1$ and $n_2$ can be drastically different, even if the underlying sizes are similar, as such cases are not captured by the scale equivariance. To ensure that the norms $n_i$s are of comparable magnitude corresponding to the size of a point cloud ${\bf P}_i$, we compute for each point cloud a scalar $d_i$ representing the approximate scale, defined as the length of the diagonal of the minimal bounding box around ${\bf P}_i$. We then explicitly enforce that the inferred norm $n_i$ is close in value to $d_i$:
\begin{equation}
\begin{aligned}
        L_{scale} = \| n_i - d_i \|_2.
\end{aligned}
\label{eqn:scale}
\end{equation}

This ensures that the norm of the inferred latent code accurately captures scale information in the object shape.
\subsection{Training in simulation}
\label{training_data}
\subsubsection{Training objective.} NeuSE is trained with partial object point clouds and corresponding 3D occupancy voxel grids of objects' complete geometry. The full model $[f_{\theta},\Phi]$ predicts the complete 3D occupancy values at query object locations, which is then evaluated by the standard cross-entropy classification loss $L_{occ} = \mathcal L(\Phi({\bf{p}}, f_{\theta}({\bf{P}}),v))$ with each of the point location $\bf{p}$ in the sampled query point cloud $\mathcal{X}$ and its corresponding true occupancy value $v$.

The ambiguous and unambiguous object categories are trained separately, with respective $L_{transform}$ and shared $L_{occ}$, $L_{shape}$, and $L_{scale}$. The final training objective is then defined as:
\begin{equation}
    L = L_{occ}+L_{transform}+L_{b\_shape} + L_{scale}.
\label{loss}
\end{equation}

The training samples are organized following $L_{b\_shape}$'s formulation, where $L_{occ}$ and $L_{scale}$ are evaluated for each sample in $B$ and $L_{transform}$ for any two observations of the same object. With this composition of the training data, the model is expected to see various pairs of viewing angles and learn to predict the relative transform between two frames within a certain range of separation.

\subsubsection{Data generation.}\label{data_gen}
NeuSE is trained fully in simulation with RGB-D images rendered with Pybullet~\citep{coumans2021}. We place a randomly posed principal object on the table, along with 2-4 (for unambiguous objects) and 1-2 (for ambiguous objects) objects arbitrarily selected from the trained categories to simulate a typical cluttered environment. To ensure a variety of viewing angles, for each multi-category object layout, we uniformly sample a fixed number of camera locations over the hollow cubical space centered around the table. The cubical space is set to be $[d_n,d_f]$ away from the table within the table plane and $[d_l,d_h]$ away from the table in the vertical direction, thus accounting for observations from near, far, low, and high viewpoints.

\section{NeuSE-based object SLAM with long-term scene inconsistency}
NeuSE enables robust data association across viewing angles and further serves as a lightweight, alternative ``sensor'' for providing cross-frame camera pose constraints. We propose a NeuSE-based localization strategy in tandem with a change-aware object-centric mapping procedure to enable robust robotic operation in scenes with long-term changes.
\subsection{System formulation and update}
Our object-based SLAM problem is formulated as a pose graph consisting of only keyframe camera pose vertices, where an edge exists to constrain the two vertices if there are inter-frame transform measurements available from NeuSE or any other sources. The measurement error between vertices $i$ and $j$ for each edge is defined as $ {\bf e}_{ij} = log({\bf Z}_{ij}\hat {\bf{T}}_{j}^{-1}\hat {\bf T}_{i})^{\vee}$, where ${\bf Z}_{ij}$ is the odometry measurement from arbitrary sources between frame $i$ and $j$, and $\bf \hat T$ is the current estimate of $\bf T$. 

The system maintains a library of keyframes with the latest camera pose estimates obtained via pose graph updates, as well as NeuSE latent codes of the observed objects in the frame coordinate. The camera pose of the current frame is recovered as the smoothed estimate of pose constraints from associated objects and external sources between the frame itself and the nearest keyframe. 

The objects in the system are recorded by their per-keyframe visibility, change status, a partial point cloud from their last keyframe observation (for query point generation during rendering), and the latest shape descriptor from initialization or mapping updates.

For localization, the system works only with latent codes in the local camera frame, while their world-frame counterparts are used for mapping operations. When an object is first observed, its world-frame latent code is initialized and then updated as needed by averaging the back-projected latent codes of the same object using the latest camera pose estimates recorded in the keyframe library.

 \begin{figure}[!htb]
    \centering
    \includegraphics[width=\linewidth]{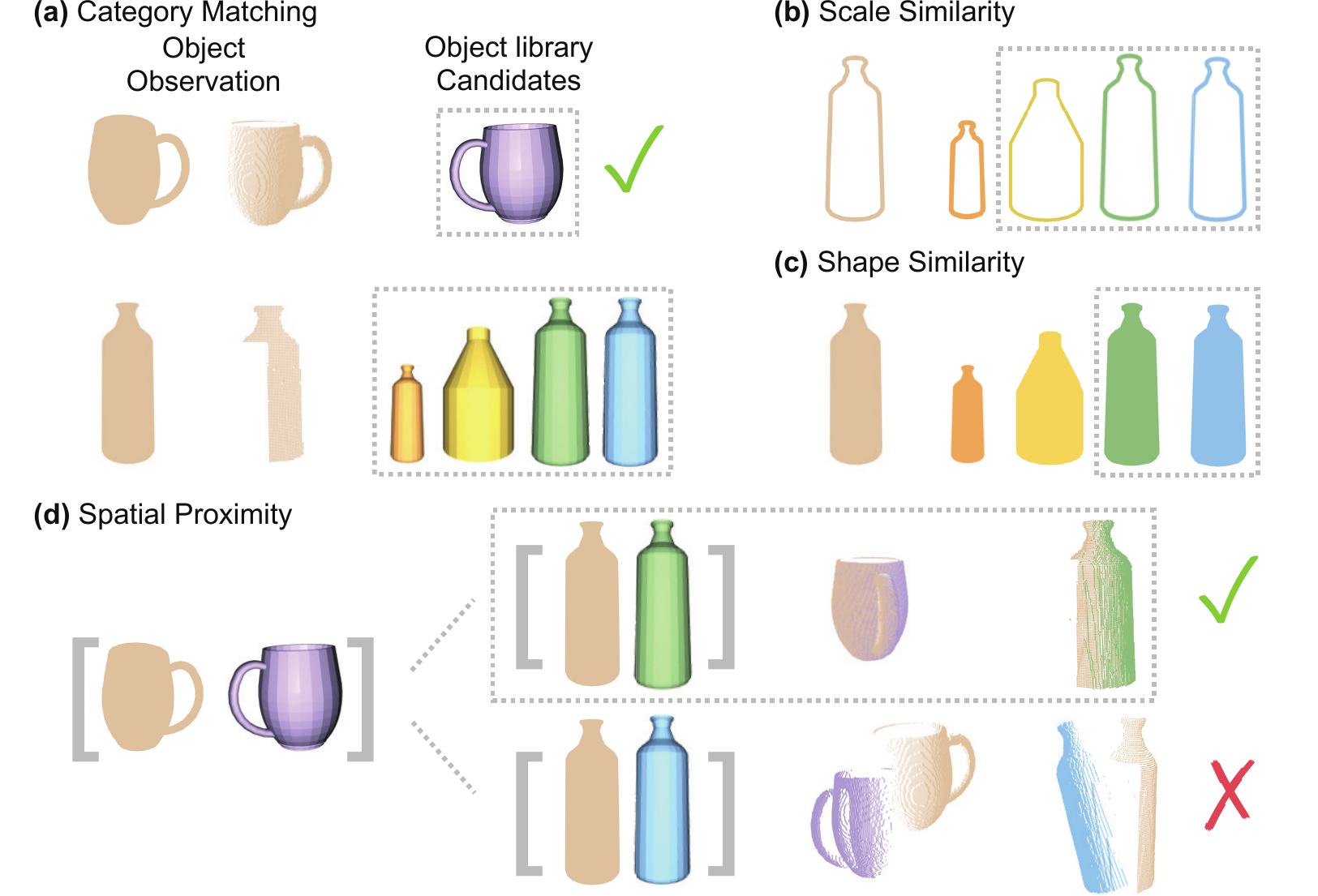}
    \caption{\textbf{Data association. }With NeuSE as the object representation, we utilize semantic labels, scale, shape, and spatial proximity for effective data association. After associating the unambiguous mug observation with the correct \textcolor{purple}{purple mug} in the database, the bottle observation undergoes four steps to eliminate incorrect candidates in the \textcolor{grey}{dashed gray box} and identify the correct \textcolor{green}{green bottle}: \textbf{(a)} Semantic matching selects the bottle-labeled instances from the object library; \textbf{(b)} Scale comparison rules out the smaller \textcolor{orange}{orange bottle}; \textbf{(c)} Shape comparison eliminates the \textcolor{yellow}{yellow bottle} with a different shape; and \textbf{(d)} spatial proximity comparison identifies the correct \textcolor{green}{green bottle}, as the transform derived from the combination of the \textcolor{purple}{purple mug} and \textcolor{green}{green bottle} results in minimal center difference for their partial point clouds.} 
    \label{data_association}
\end{figure}

\subsection{Data association}
NeuSE-predicted inter-frame transforms are only valid when computed from latent codes belonging to the same object. Our data association scheme first leverages semantic labels to select candidates within the query observation's category, and then progressively uses scale, shape similarity, and spatial proximity to ensure that pose constraints are generated only between latent codes with reliable object associations (see \autoref{data_association}).

\subsubsection{Semantic category matching.}  To improve the efficiency of object matching, we first use semantic labels to eliminate candidates from other categories, focusing only on potential matches within the same category, as illustrated in \autoref{data_association}(a).

\subsubsection{Scale similarity.} Leveraging NeuSE's scale equivariance, we filter data association candidates based on object sizes. We compute the norm of the latent code for the current observation and search for objects in the library within a $|\delta_{size}|$ scale difference. This reduces the candidates to those of similar sizes to the object represented by the query partial point cloud,  thereby reducing the computational load in subsequent steps, as shown in \autoref{data_association}(b).

\subsubsection{Shape similarity.} As described in \autoref{data_association}(c), when multiple candidates are of similar sizes to the current object,  we then extract the shape descriptor from the latent code and calculate its cosine shape similarity (as adopted in equation~\eqref{triplet}) with all object candidates. Objects with a shape similarity score greater than $\delta_{shape}$ are considered potential data association candidates $\mathcal{O}_c$. Again, if no similarity scores exceed $\delta_{shape}$, a new object instance is initialized and added to the object library.

\subsubsection{Spatial proximity.} Spatial proximity involves examining the Euclidean distance between the partial point cloud centers of the current object and its candidates in $\mathcal{O}_c$. The current partial center is projected to the world frame to compare the distance with that of the latest keyframe where its candidate was last seen. As shown in \autoref{data_association}(d), the transform for projection is computed using Horn's method (equation~\eqref{horn}) between corresponding concatenated latent codes. The candidate with the smallest distance, if below $\delta_{prox}$, is deemed a successful match and included in $\mathcal{O}_{matched}$ for further pose constraint generation. Otherwise, the current object remains unassociated and is grouped into $\mathcal{O}_{unmatched}$.

The procedure is first performed on unambiguous objects, followed by ambiguous objects.  For unambiguous objects, we compute the transform directly from the concatenated latent codes using Horn's method, selecting the candidate combination that minimizes the partial center distances of all observed unambiguous objects in the frame. For ambiguous objects, we utilize the transform from associated unambiguous objects if available. If not, we perform an exhaustive search of all paired combinations of covisible object candidates in previous keyframes, similar to  the process for unambiguous objects. We compute the inter-frame transform from the concatenated latent codes, selecting the combination that results in the smallest partial center distance for all object observations in the current frame.

Hence, we divide all covisible objects $\mathcal{O}$ in one frame into three groups: (1) $\mathcal{O}_{matched}$, which has objects with shape, scale, and spatial consistency, and is adopted for pose constraint generation, (2) $\mathcal{O}_{unmatched}$, which consists of scene changes or temporally ambiguous observations, and is processed by change detection, and (3) new objects never seen before.
\subsection{Pose graph optimization}
\begin{figure}[!htb]
    \centering
    \includegraphics[width=\linewidth]{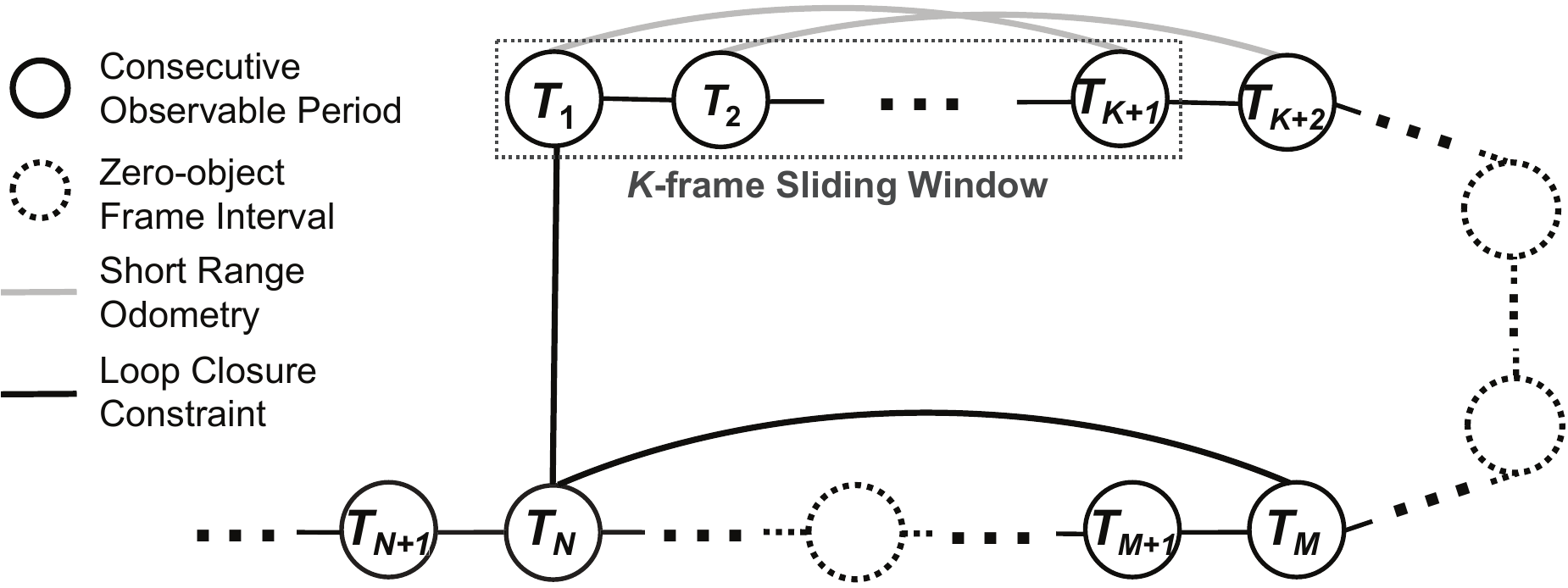}
    \caption{\textbf{Pose graph optimization}. With objects observed in periods of consecutive frames, we derive from corresponding latent codes (1) short-range odometry constraints (\textcolor{gray}{gray}) within a local $K$-frame sliding window, and (2) global loop closure constraints (black) between the current (${\bf T}_N$) and the first frame of each of its previous consecutive observable periods (${\bf T}_1$ and ${\bf T}_M$), which are working  jointly to constrain the pose graph optimization.  } 
    \label{pose_graph}
\end{figure}

With objects successfully associated across frames, we compute NeuSE-predicted transforms among frames to constrain the pose graph both locally and globally (see \autoref{pose_graph}). 

\subsubsection{Keyframe selection.}Keyframes are selected based on the presence of new objects and proximity to previous keyframes. New objects trigger the selection of a frame as a keyframe, and frames located at least 0.04~m away from the previous keyframe based on accumulated odometry are also chosen. Additional keyframes may be added after change detection for frames with changes.

\subsubsection{Short-range odometry.} To reduce local drift in frames with persistently observed objects, short-range NeuSE-predicted pose constraints are applied to a sliding window optimization of $K$ keyframes. For each newly added keyframe, we search its preceding $K-1$ keyframes  and identify the common objects observed between the current and previous frames. The inter-frame transform constraint is computed based on the concatenated latent codes of the shared objects (or a single unambiguous object) and then added as an edge to the pose graph.

\subsubsection{Long-range loop closing.} Global loop closing is activated when an object is detected again in a frame after its last consecutive observable period. The common objects between the current frame and the initial frames of all its previous observation periods are identified, and relative transform constraints are derived from the concatenated NeuSE latent codes. These constraints are then added to the pose graph, which initiates a global optimization process using the latest pose estimates from the local sliding-window optimization as the starting point.

\subsection{Change-aware object-centric mapping}
\label{change_detection}
Change detection is performed frame by frame on objects in $O_{unmatched}$ that match in shape but are identified as spatially apart based on latent codes, providing a foundation for consistent long-term mapping. 

Since changes are often gradual and occupy only a small portion of the object clutter in long-term scenes, here change detection is performed by comparing the relative layout of the query unmatched object $o_{ui} \in \mathcal{O}_{unmatched}$ with all objects $o_{mi} \in \mathcal{O}_{matched}$ in the matched set serving as anchors. We argue that the relative object position disparity is more robust to camera pose drift compared to the absolute position difference, as all objects observed in the same frame will drift concurrently in the world frame while maintaining their relative layout. 

\begin{figure}[!htb]
    \centering
    \includegraphics[width=\linewidth]{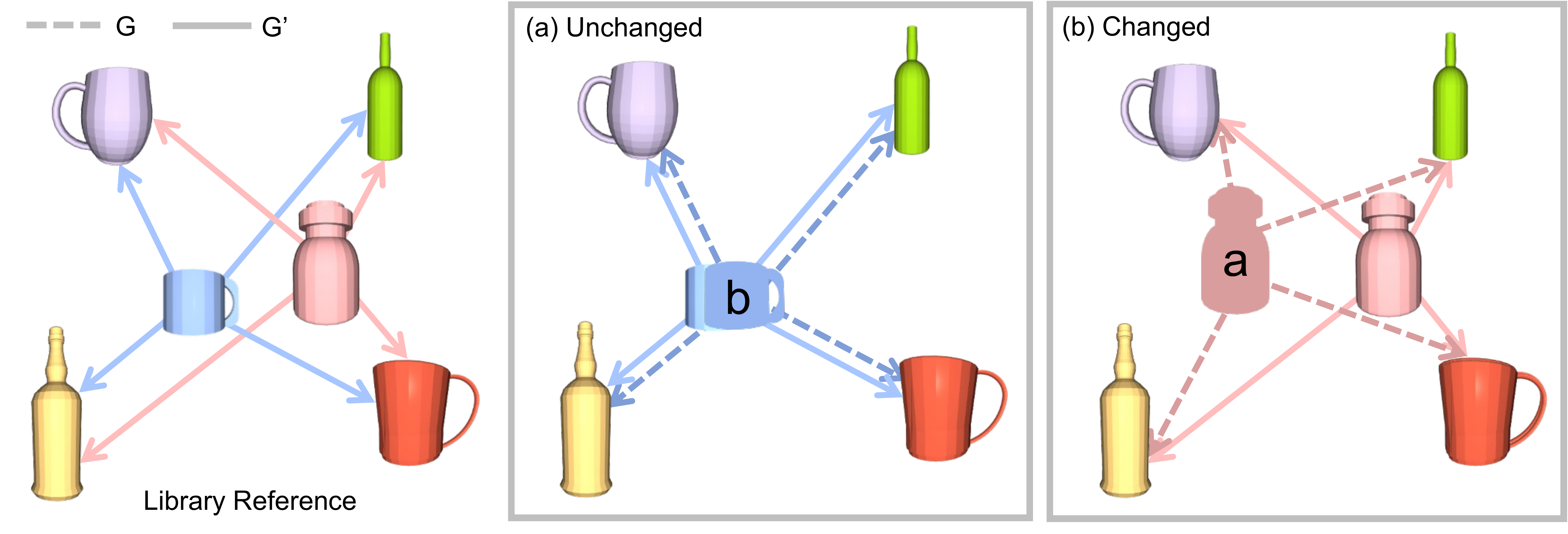}
    \caption{\textbf{Object layout comparison through graph matching}. Object graphs are constructed for the current frame ($G$) and the library ($G^\prime$). For object \textcolor{blue}{$a$} and \textcolor{pink}{$b$}, which are similar in shape to the \textcolor{blue}{blue mug} and \textcolor{pink}{pink bottle} in the library, respectively, the inter-object distances between them and the anchor objects in the four corners are computed and compared. \textbf{(a)} All corresponding edges (dashed and solid lines) with anchor objects have similar oriented lengths, indicating that the mug is unchanged but was seen with an occluded handle, leading to a false ambiguous transform by the latent code. \textbf{(b)} There are no similar edges, indicating a different layout with the bottle moved.}
    \label{graph_matching}
\end{figure}

 We represent the local layout with a directed object graph $G$ constructed with $\mathcal{O}_{unmatched}$ and $\mathcal{O}_{matched}$. Each vertex of $G$ represents an object $o$ with its shape descriptor and the true object center as $(\bf s,c)$.  The center $\bf c$ is computed from the full object reconstruction using the decoding steps in equation~\eqref{se3} and back-projected to the world frame using the latest camera pose estimate. Edges are established between objects $o_{ui} \in \mathcal{O}_{unmatched}$ and all anchor objects $o_{mj} \in \mathcal{O}_{matched}$, indicating the oriented  distance between their centers $E=\{{\bf e}_{ij} \vert {\bf e}_{ij}={\bf c}_{ui}-{\bf c}_{mj},\forall {o}_{ui} \in \mathcal{O}_{unmatched}, {o}_{mj} \in \mathcal{O}_{matched}\}$. 
 
 We build the local and reference object graphs, $G$ and $G^\prime$, respectively, for $\mathcal{O}$ in the current frame and their associated or shape-similar counterparts in the system library (see \autoref{graph_matching}). After a quick alignment of the two graphs using the centers of anchor objects, for each pair of edges  $({\bf e}_{ij},{\bf e}_{i'j'})$ connecting vertices of similar shapes (determined by the scale and then $\bf s$ from data association), we compare their edge value disparities to assess whether this is a changed layout:
\begin{equation}\label{edge}
\begin{aligned}
   \sum\limits_{j}\mathds{1}(\vert {\bf e}_{ij}-{\bf e}_{i'j'} \vert \leq \delta_e) 
   = \begin{cases}0, & \text{changed}\\\geq 1, & \text{unchanged}.
\end{cases}
\end{aligned}
\end{equation}

An object $o_i$ is marked as \emph{unchanged} if at least one pair of edges is found to be closer than a threshold $\delta_e$. This indicates that its inter-spatial relationship with at least one of the anchor objects is consistent. If no edges are found to be close, the object is marked as \emph{changed} and its change status and partial point cloud are updated in the object library. Here, we define an object to be ``removed'' from the scene if it has never been shape-matched during frame periods with global loop closure. 

Therefore, we are able to maintain a lightweight, object-centric map that accurately reflects full object reconstructions from NeuSE predictions. By using objects as the basic building blocks of the map, we can update changes seamlessly by replacing the old latent code with the new one during the decoding stage, avoiding cumbersome and artifact-prone point- or voxel-wise modifications commonly used in traditional low-level geometric maps.

\section{Experiment setup}
In our experiments for evaluation, we first assess NeuSE's efficacy in characterizing object shapes and poses from partial object observations. We then investigate how NeuSE can be utilized for robust spatial understanding across scenes.

Specifically, we aim to answer three questions: \textbf{(1)} Can NeuSE provide an effective object representation with \sethree~and scale equivariance, enabling accurate shape and pose descriptions even under challenging conditions such as varying viewpoints and occlusions? \textbf{(2)} Can NeuSE-based object SLAM perform reliable localization independently or enhance existing results when integrated with other SLAM measurements, particularly in the presence of temporal scene inconsistency? \textbf{(3)} Can the proposed approach build a consistent object-centric environment map with timely updates to reflect scene changes?

We therefore trained NeuSE fully in simulation and evaluated the proposed algorithm directly on both synthetic and real-world sequences, featuring \emph{unseen} objects from the trained categories with \emph{greater size variation} than the training data. We first verified the proposed \sethree-~and scale-equivariant object characterization of NeuSE using controlled synthetic sequences and further demonstrated its applicability to long-term object-based SLAM on both synthetic and real-world traversals, where multi-category objects of diverse sizes were added, removed, or switched places.

Given the limited availability of object model collections for training and the scarcity of public datasets featuring appropriate object-level scene changes, we generated our own synthetic and real-world testing sequences. These sequences included multiple categories of daily objects with gradual layout changes in both controlled experimental settings and less structured lab-office environments.

We selected objects from five categories: mugs, bottles, bowls, cans, and trash bins, to capture a diverse range of shapes, sizes, and spatial ambiguities, such as cylindrical or rectangular shapes. This diversity allows for a comprehensive evaluation of the effectiveness and generality of our latent code design.
\subsection{Implementation details}
To train NeuSE's occupancy network, for each object category, we generated training samples using the corresponding models from ShapeNet~\citep{shapenet2015} rendered in Pybullet, each containing 60,000 RGB-D partial observations with segmentation masks. We followed the sample generation strategy proposed in Section~\ref{data_gen}: 2,000 object layouts were created in Pybullet with the target object category mixing with all categories of interest, from each of which 30 views  were uniformly sampled with $[d_n,d_f]=[0.3,4]$~(m) and $[d_l,d_h]=[-0.4,0.4]$~(m). We trained our model on two NVIDIA RTX 3090 GPUs using a learning rate of $5 \times 10^{-4}$ with the Adam optimizer.  The latent code size was $k=512$ and the occupancy threshold for reconstruction was $v_0=0.5$.  The training batch was populated with eight object shapes, each with 15 partial observations, by setting $M = 15$ and $N = 8$. 

For the object SLAM system, we used Mask R-CNN~\citep{8237584} to generate segmentation masks of interest. For data association and change detection, we used $(\delta_{size},\delta_{shape})=(0.01,0.95)$, $(\delta_{prox},\delta_{e})= (0.03,0.02)$~(m) for the synthetic sequences, $(\delta_{size},\delta_{shape})=(0.02,0.90)$, $(\delta_{prox},\delta_{e})= (0.03,0.02)$~(m) for the controlled environment and $(\delta_{size},\delta_{shape})=(0.02,0.90)$, $(\delta_{prox},\delta_{e})= (0.03,0.03)$~(m) for the hand-held sequences. We set the sliding window size to $K=10$ and adopted the factor graph representation for SLAM pose graph optimization. The local sliding-window optimization was solved with a Levenberg-Marquardt fixed-lag smoother, and the global pose graph was solved with iSAM2~\citep{5979641}, both using implementations from GTSAM~\citep{gtsam}.
\section{SE(3) and scale equivariance of object representations}
We first assess the \sethree~and scale equivariance of object representations in NeuSE, which are crucial for its effective application in robotic spatial understanding tasks. In principle, NeuSE is \sethree-~and scale-equivariant by construction when given the same transformed object point cloud. However, in real-world robotic executions, partial point clouds of an object vary across viewing angles, making it important to demonstrate that NeuSE's equivariance is learned to still hold across different observations of the same object.

To test the robustness of NeuSE’s \sethree~and scale equivariance under partial observations, we conducted controlled experiments in simulation. In these experiments, we applied \sothree~rotations, three-dimensional translations, and scaling to objects, observing them from the same view before and after each transformation $\mathbf{T}$. We then followed the definition of each type of equivariance and compared the distance between the resulting object latent codes ${(\mathbf{z}, \mathbf{z'})}\in \mathbb{R}^{k\times 3}$ using a distance measure $D(\mathbf{Tz}, \mathbf{z'})$ to determine whether NeuSE, in its latent space, captured the transformation applied to the object in the physical world.

The testing data were generated in PyBullet using object instances from ShapeNet, covering all five chosen categories. For each equivariance experiment, that is, rotation, translation, and scaling, we created 100 samples per category, for a total of 500 samples. In each sample, an object was randomly selected from a category, and a randomly generated transformation was applied to obtain object observations before and after the change from the same camera view.

\begin{figure}[!htb]
    \centering
    \includegraphics[width=\linewidth]{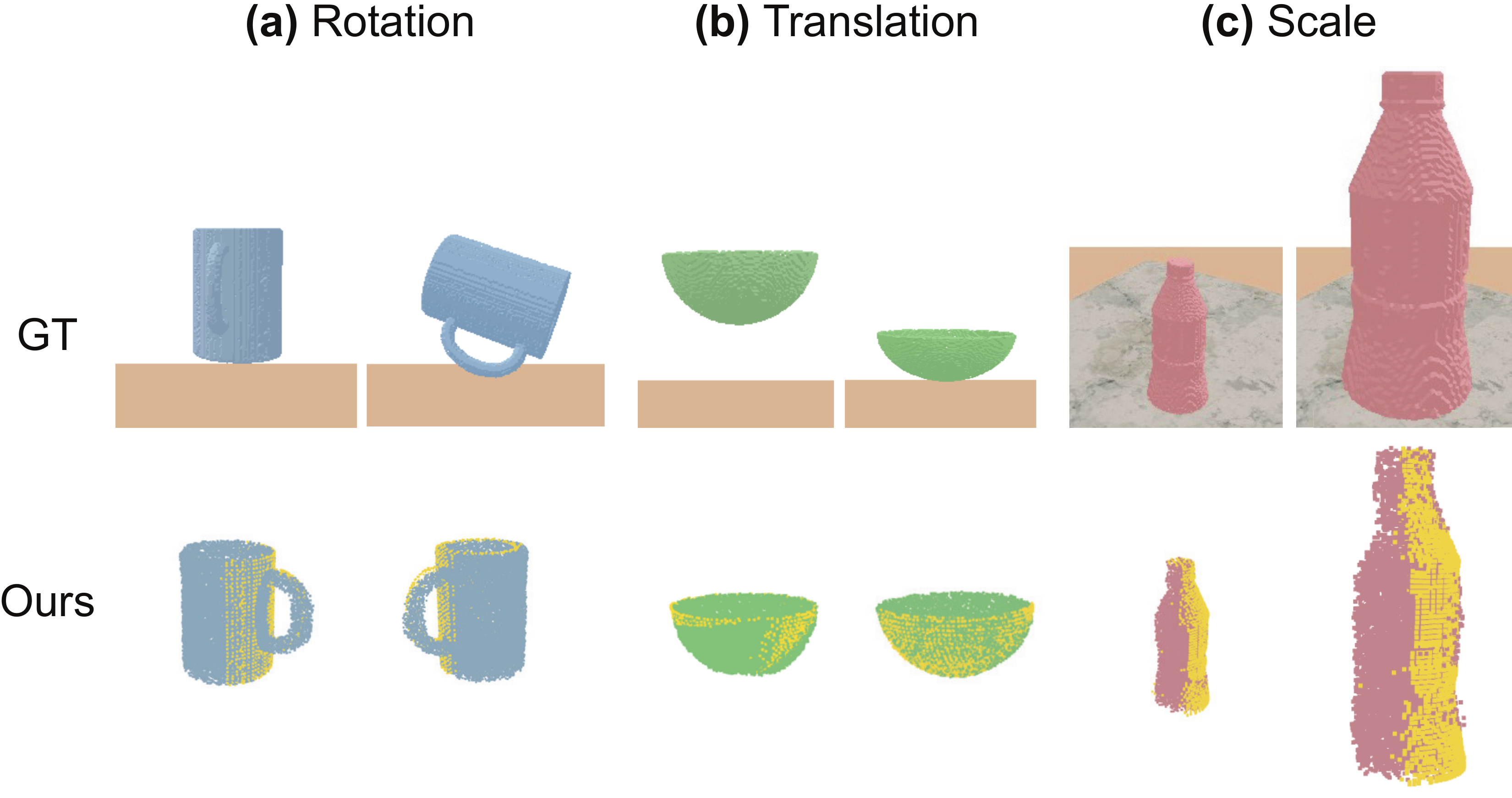}
    \caption{\textbf{Robust reconstruction with equivariance.} Examples of robust object reconstructions (Ours) compared to the ground truth (GT). Yellow points represent partial point cloud observations, and the colored reconstructions are shown from viewing angles that emphasize the occluded parts completed by NeuSE. Across the three transformations applied to the example objects: \textbf{(a)} \sothree~rotation on the {\color{dark_blue!70!black} mug}, \textbf{(b)} 3D translation on the \textcolor{green}{bowl}, and \textbf{(c)} scaling on the \textcolor{pink}{bottle}, NeuSE effectively encodes and reconstructs the full object shape, consistently aligning with partial observations from both the original and transformed states.} 
    \label{equivariance_reconstruction}
\end{figure}

We quantitatively present the distribution, along with the mean and standard deviation of the errors of the NeuSE-inferred distribution, in \autoref{equivariance}, and provide qualitative examples of NeuSE's reconstructions of different object instances before and after transformation in \autoref{equivariance_reconstruction}.
 \begin{figure}[!tb]
    \centering
    \includegraphics[width=0.95\linewidth]{figures/8.pdf}
    \caption{\textbf{Equivariance error distribution. }Scatter plots showing the error distributions from the testing data across five object categories in three equivariance experiments. \textbf{(a)} Most normalized RMSE values across the full rotation range $(0^\circ, 180^\circ)$ are bounded within a range close to zero, indicating small differences between the latent codes of the original and rotated objects, thus demonstrating strong \sothree-equivariance. \textbf{(b)} The RMSE values between predicted and ground truth translations are concentrated below 0.01~m, reflecting NeuSE's high translation encoding accuracy. \textbf{(c)} For scale changes ranging from one to nearly seven times the original size, the errors between the estimated and ground truth scale ratios average around 0.011, with most error values ranging between -0.01 and 0.03,  suggesting NeuSE's scale equivariance across object sizes.}
    \label{equivariance}
\end{figure}

\subsection{Rotation equivariance}\looseness=2
Following the definition of rotation equivariance in equation~\eqref{so3equi}, \autoref{equivariance}(a) illustrates the distribution of the Normalized Root Mean Squared Error (RMSE) between the latent codes ${\bf z}$ and ${\bf z}'$, representing the object before and after rotation. Given that the metric distance between these two latent codes is measured in the latent space, it is challenging to interpret their proximity  solely  based on the magnitude of this distance. To provide a more interpretable measure, we adopt the normalized RMSE, which quantifies the relative difference between the latent codes w.r.t. their average norms.

The normalized RMSE is defined as:
$$
\text{Normalized RMSE} = \frac{\sqrt{\frac{1}{k} \sum_{i=1}^{k} (  ({\bf Rz})_i - {\bf z'}_i )^2}}{mean(\left\| {\bf z} \|_2+ \| {\bf z'} \|_2 \right)},
$$
where ${\bf R}\in \mathbb{SO}(3)$ is the applied random rotation. 

As shown in \autoref{equivariance}(a), with the magnitudes of sampled rotations covering nearly the entire $(0^\circ, 180^\circ)$ range and a mean normalized RMSE of $0.83\%$, we observe a consistent trend of low normalized RMSE values across different rotation angles for all object categories, indicating NeuSE's equivariance with physical-world rotations. Notably, the variation in normalized RMSE remains within an acceptable range of $(0.4\%, 1.2\%)$, further demonstrating NeuSE's consistency in encoding the intrinsic geometric structure of objects in the latent space after various degrees of rotation. This is further supported by \autoref{equivariance_reconstruction}(a), which presents an example of a full mug reconstruction before and after applying the rotation. Despite the significant change in the visible portion caused by the rotation (the yellow point region), the full mug reconstruction still aligns well with the visible part and accurately recovers the complete mug shape with the correct orientation.
\subsection{Translation equivariance}The other component of \sethree~equivariance is translation equivariance. Given an applied translation ${\bf t}\in \mathbb{R}^3$, we measure the RMSE between the two latent codes before and after applying the transformation as:
$$
\text{RMSE} = \sqrt{\frac{1}{k} \sum_{i=1}^{k} (({\bf z}+{\bf t})_i - {\bf z'}_i )^2}.
$$

In \autoref{equivariance}(b), we plot the RMSE against the translation magnitude, where the translation is controlled to be within 0.4~m. This limit is based on the maximum potential distance we set between two keyframes within the sliding window of the pose graph. Most of the points, well within the standard deviation band, are concentrated below 0.01~m, regardless of the object category or translation magnitude. This demonstrates NeuSE's ability to robustly encode the complete object shape, effectively inferring the offset between the observed partial object center and the true object center, despite variations in the observable parts caused by translation. This finding is further supported by \autoref{equivariance_reconstruction}(b). Whether the bowl is initially observed in a closer, upper position with only the bowl body visible (left), or later seen in a more distant, lower position with the rim also visible (right), NeuSE consistently reconstructs the complete bowl shape, which aligns reliably with the yellow observable portion.

\subsection{Scale equivariance}To evaluate scale equivariance, we apply a scaling transformation to change the object size (scaling the bounding box diagonal length, as adopted in equation~\eqref{eqn:scale}), by a factor $s$ within the range of $[1.03, 6.75]$. This results in overall object sizes ranging from 0.05~m to 0.75~m, which is similar to the size range of daily objects used in our real-world experiments. To quantify the change in NeuSE's representation before and after scaling, we compute the difference between the NeuSE-inferred scaling ratio and the ground truth ratio $s$ as:
$$
\text{Estimated Ratio Error} = \frac{\left\| {\bf z'} \right\|_2}{\left\| {\bf z} \right\|_2}-s.
$$

As shown in \autoref{equivariance}(c), NeuSE effectively generalizes across different object categories and size variations, maintaining a small and consistent estimated ratio error even with scaling up to 6.75 times the original size. For smaller scaling factors (e.g., below 1.5), the inferred ratios closely match the ground truth, remaining well within the standard deviation band and close to zero. This demonstrates NeuSE’s accuracy in encoding object sizes. Even with larger scaling transformations that introduce significant changes to partial observations, NeuSE kept the prediction error mostly below 0.05, even for size increases ranging from 3 to 6.75 times the original size. \autoref{equivariance_reconstruction}(c) further illustrates NeuSE’s ability to generalize across scales with an example of the bottle category. Despite the object being scaled to approximately three times its original size, NeuSE reliably reconstructs both the small and bigger bottles, demonstrating its robustness in capturing object shapes across varying scales.

\section{Spatial understanding with NeuSE: localization with temporal scene inconsistency}
Having demonstrated the equivariant properties of object representations in NeuSE, we proceed to assess whether these properties enable NeuSE to support robust spatial understanding for robots. Specifically, we explore how NeuSE, by leveraging its inherent equivariance, can facilitate reliable localization in environments with long-term scene inconsistency and achieve change-aware object-centric mapping.

Building on NeuSE's desirable equivariance properties, in this section, we first demonstrate its application to enhance robot localization in environments with long-term changes. To evaluate the effectiveness of the proposed NeuSE-based object SLAM paradigm, we collected testing sequences in both simulation and real-world environments involving the five selected object categories. These sequences featured multiple instances of objects with varying shapes and sizes, with camera motion ranging from \emph{steady} movement with a robot-mounted camera in a controlled environment to more dynamic, \emph{less constrained} movement using a hand-held camera in larger, less structured lab-office spaces. In this way, we included various scenarios with varying levels of scene complexity in the testing sequences.

We report the performance of our approach and baselines on localization with temporal scene inconsistency in both simulated and real scenes. All results were obtained on a laptop with an Intel Core i7-9750H CPU and an Nvidia GeForce RTX 2070 GPU. NeuSE network inference takes 6~ms per object, with inter-frame pose constraint calculation taking 1~ms. One-time rendering for object-centric map construction costs 30~ms per object with 20,000 query points. With data association included, the speed is approximately 20~fps for generating object-level inter-frame pose constraints with our NeuSE-based front end, making it possible for NeuSE to be integrated as an external ``constraint sensor'' with real-time operating speed. The final overall localization speed of our change-aware SLAM system is around 4~fps in the current, more crowded controlled experiment setting, with sequential processing of observed objects, no software optimization, or major tuning of the back-end iSAM2 solver. 
\subsection{Synthetic sequences }
The ability to easily render observation trajectories in simulation allows us to extensively evaluate  NeuSE's capabilities for conducting change-aware localization and mapping and study their sensitivity to various factors in simulation. In addition, in the multi-category setting in simulation, all object categories, except mugs, possess some cylindrical or rectangular ambiguity, enabling the validation of our design on employing covisible ambiguous objects to support localization. 
\begin{figure}[!htb]
    \centering
    \includegraphics[width=\linewidth]{figures/10.pdf}
    \caption{\textbf{Synthetic data overview}.  Table layouts with object changes and the ground truth camera trajectories of \textbf{(a)} the mug-bottle, 10-table setting, and \textbf{(b)} the multi-category, 12-table setting.}
    \label{sim_data_overview}
\end{figure}

\begin{table*}
    \centering
    \caption{RMSE of ATE and translational RPE on synthetic sequences. Best results are marked in bold.}
    \label{rmse_sim}
    \small
    \scalebox{0.85}{
    \begin{tabular}{lcccccccc}
        \toprule
          &  \multicolumn{4}{c}{\bfseries Planar} &  \multicolumn{4}{c}{\bfseries Non-planar} \\
          \cmidrule(lr){2-5} \cmidrule(lr){6-9} 
          &  Mug-only & Mug + Bottle & Multi (\sethree-Only) & Multi  & Mug-only & Mug + Bottle & Multi (\sethree-Only) & Multi \\
         \midrule
          \textbf{RMSE of ATE (m)}\\
          $1^{\text{st}}$ traversal &  0.072 & 0.043 & 0.034 & \textbf{0.021} & 0.058 & 0.045 & 0.028 & \textbf{0.026}\\
          $2^{\text{nd}}$ traversal & 0.096 & 0.071 & 0.049 & \textbf{0.041} & 0.077 & 0.033 & 0.041 & \textbf{0.029} \\
          Full & 0.116 & 0.065 & 0.052 & \textbf{0.047} & 0.091 & 0.053 & 0.042 & \textbf{0.036}\\
          \midrule
          \textbf{RMSE of Trans RPE (m/f)}\\
          Full & 0.026 & 0.017 & \textbf{0.014} & \textbf{0.014} & 0.024 & 0.016 & 0.015 & \textbf{0.012}\\
        \bottomrule
    \end{tabular}
    }
\end{table*}

\subsubsection{Data preparation.}
The environment was rendered in PyBullet with objects placed on multiple tabletops across a $10~\text{m}\times 15~\text{m}$ area. To thoroughly evaluate NeuSE's \sethree~and scale equivariance, and its capability to handle covisible ambiguous objects, two object sets of increasing complexity were used, varying in instance density, shape, size, and occlusion patterns: (1)  50 unseen mug and bottle instances distributed across 10 tables (\autoref{sim_data_overview}(a)) adopted in ~\cite{fu2023neural}, and (2) approximately 75 unseen multi-category instances, including mugs, bottles, bowls, cans, and trash bins, placed on 12 tables (\autoref{sim_data_overview}(b)). To further assess NeuSE's \sethree~equivariance, two different object layouts were generated: (1) a roughly planar layout with all objects standing upright, and (2) a non-planar, ``hilly'' layout, with nearly half of the objects laid down and arbitrarily oriented on tabletops. The camera followed a preset closed-loop trajectory, recording RGB-D images and segmentation masks for both layouts. This resulted in two sequences featuring \emph{uninterrupted} object observations for each of the two object sets, where most objects were revisited from approximately opposite views with low overlap. For each sequence, objects were added, removed, or relocated, resulting in 9 to 10 changes along the trajectory.
\begin{figure}[!htb]
    \centering
    \includegraphics[width=\linewidth]{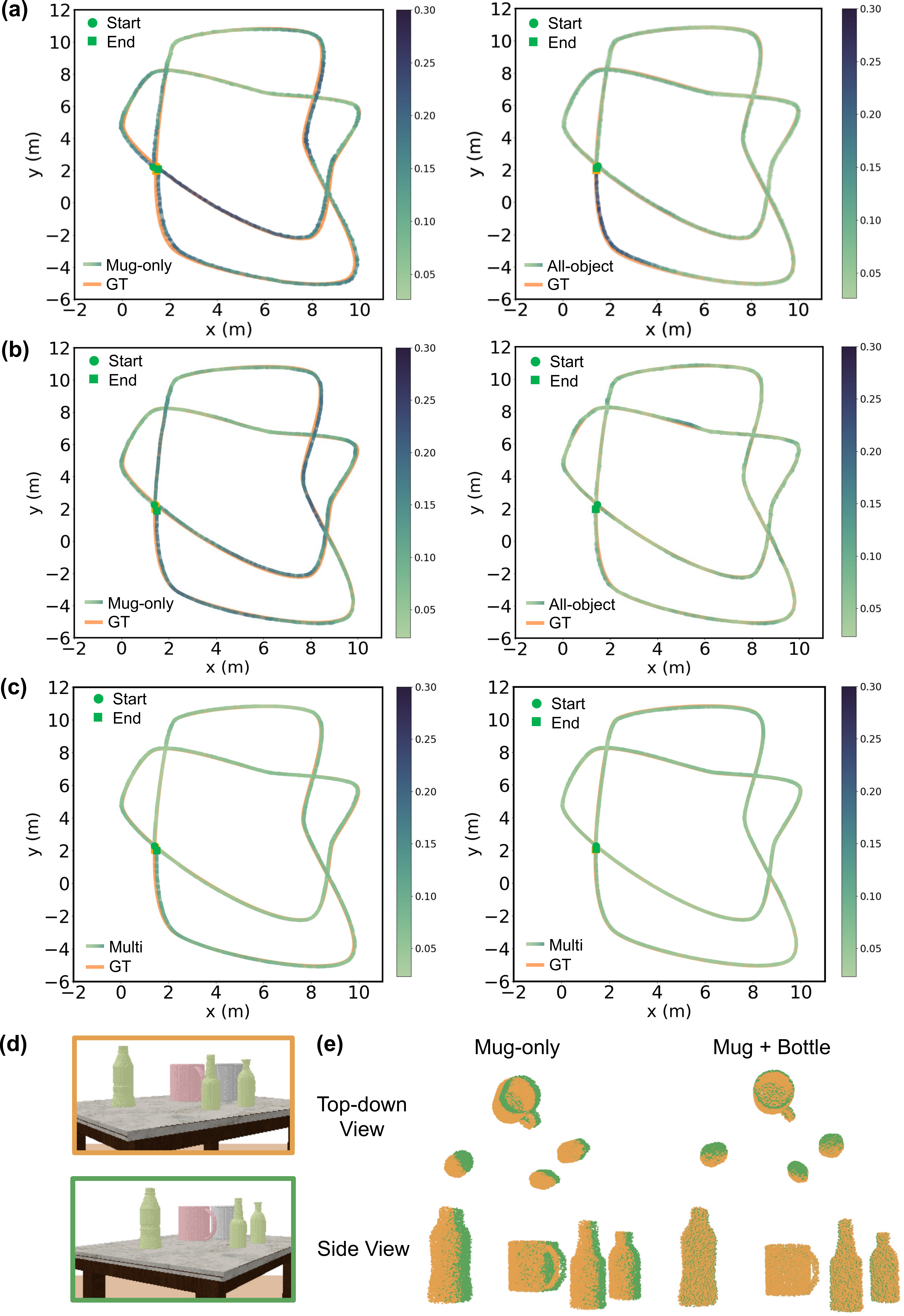}
\caption{\textbf{(a)--(c): Comparison of estimated and ground truth (GT) trajectories on synthetic sequences.} \textbf{(a)} Mug + Bottle planar sequences. \textbf{(b)} Mug + Bottle non-planar sequences. \textbf{(c)} Multi-category planar (left) and non-planar (right) sequences. Trajectory color variation indicates ATE distribution. The decreasing ATE values, shown by lighter trajectory colors from \textbf{(a)} to \textbf{(c)}, demonstrate the efficacy of NeuSE's \sethree-equivariance in handling randomly oriented objects from \textbf{(a)} to \textbf{(b)} and highlight the benefit of including ambiguous objects for improved inter-frame transform prediction. \textbf{(d) and (e): Ambiguous objects for inter-frame transform prediction. }In \textbf{(d)}, object point clouds are transformed from the \textcolor{orange}{orange frame} to the \textcolor{green}{green frame} using transforms derived from the \textcolor{pink}{pink mugs} alone, or together with \textcolor{green}{green bottles}. The improved point cloud alignment in \textbf{(e)}, using all the objects (Mug + Bottle) compared to Mug-only, demonstrates the effectiveness of using covisible ambiguous objects to enhance transform prediction accuracy.}
    \label{chamfer_res}
\end{figure}

\subsubsection{Localization with scene changes and object variation.}

To assess NeuSE's capability to handle occlusion and viewing angle variation, and to validate our approach of including ambiguous objects to improve localization accuracy, we compare three data variants for both planar and non-planar object configurations: (1) localizing using only mugs in the mug-bottle sequences (Mug-only), (2) localizing using all objects in the mug and bottle sequences (All-object), and (3) localizing using all objects in the multi-category sequences (Multi). 

Furthermore, since the objects in the multi-category sequences have a wider size range ($0.05-0.75$~m) compared to those in the training data ($0.3-0.6$~m), we also report the performance of the \sethree-only variant of NeuSE without scale equivariance (Multi(\sethree-only)) from~\cite{fu2023neural} for the planar and non-planar multi-category sequences. This provides a clearer examination of the benefits of scale equivariance in enhancing NeuSE's object characterization ability, particularly in dense object layouts. 

In this way, building upon the \sethree-only results from~\cite{fu2023neural} on the mug-bottle sequences for configurations (1) and (2), we further explore the impact of both object ambiguity and scale equivariance on the accuracy of camera pose constraint generation.

For the few frames with no objects for data association or pose generation, we maintained system operation with odometry measurements corrupted from ground truth by a zero-mean Gaussian noise with $\sigma=0.003$~rad for rotation and $\sigma=0.05$~m for translation. The RMSE of the translational Relative Pose Error (RPE) and Absolute Trajectory Error (ATE) are used as evaluation metrics.

We report quantitatively in \autoref{rmse_sim} the RMSE of both the translational RPE and the ATE of the estimated camera poses for the four testing sequences, showcasing NeuSE's consistent performance both locally and globally. We further visualize the ATE distribution along the way in \autoref{chamfer_res}. 

The  RPE and ATE values in \autoref{rmse_sim} show that (1) NeuSE is a reliable ``constraint sensor'' for producing consistent short- and long-range camera pose constraints, and (2) our system is capable of producing a globally consistent trajectory, despite various object sizes, occlusion patterns, viewing angle disparities, and object changes along the traversal. 

Specifically, we observe from \autoref{rmse_sim} that the proposed object SLAM approach performs better on the non-planar object layout, fully showing the efficacy of our \sethree-equivariant representations in handling randomly oriented objects. This can be attributed to our training data generation strategy, which includes various views and occlusion patterns to learn robust geometric features of object shapes across viewing angles. Further, the lying-down mugs in the sequence help reduce shape ambiguity by providing more valid observations for generating camera pose constraints, as their handles are more frequently visible when pointing upwards than in the usual sideways direction. With the \sethree-equivariant property of NeuSE, our approach can learn from upright observations to benefit the processing of laid down objects, thus enabling generalization to new scenarios with various object orientations.

\begin{table}
    \centering
    \caption{Data association accuracy comparison with point clouds (PC-based) and NeuSE (Ours). Best results are in bold.}
    \label{da_accuracy}
    \small
    \resizebox{\columnwidth}{!}{ 
    \begin{tabular}{lcccc}
        \toprule
         & \multicolumn{2}{c}{\textbf{Planar}} & \multicolumn{2}{c}{\textbf{Non-planar}} \\
         \cmidrule(lr){2-3} \cmidrule(lr){4-5}
         &  \textbf{PC-based} & \textbf{Ours} &  \textbf{PC-based} & \textbf{Ours} \\
        \midrule
        ${1^\text{st}}$ traversal & 83.6\% & \textbf{91.2\%}  & 79.7\% & \textbf{90.3\%} \\
         Full & 37.1\% & \textbf{88.6\%} & 36.6\%& \textbf{89.1\%} \\
        \bottomrule
    \end{tabular}
    }
\end{table}

Validation of using ambiguous objects for pose constraint generation is confirmed by the decreasing RPE and ATE RMSE values in \autoref{rmse_sim} with increasing object number and orientation complexity from the mug-only to multi-category and planar to non-planar scenarios. In addition, in \autoref{chamfer_res}(d) and (e), with object point clouds in (d) transformed from the upper (orange) to the lower (green) frame using transforms derived from only the pink mug and together with green bottles, the improved point cloud alignment in (e) for  Mug + Bottle compared to Mug-only demonstrates the viability of leveraging covisible ambiguous objects for improving transform estimation accuracy. 

Furthermore, from \autoref{rmse_sim} we can also conclude that the incorporation of scale equivariance greatly enhances NeuSE's ability to generalize across objects of different sizes, as indicated by NeuSE's lower ATE and RPE RMSE values compared to its \sethree-only counterpart. Scale equivariance facilitates the effective transfer of learned geometric knowledge to objects with similar shapes but varying sizes, thereby allowing NeuSE to accurately infer previously unseen object sizes with comparable precision and contributing to its broader generalization capability.

\subsubsection{NeuSE for robust data association.} In the last section, we demonstrate NeuSE's effectiveness as a \emph{standalone} ``constraint sensor'' for generating consistent camera pose constraints, due to its successful data association throughout the trajectory with continuous object observations. However, real-world SLAM systems do not always have uninterrupted object observations and often have external odometry measurements that are prone to noise and drift. 

\begin{figure}[!htb]
    \centering
    \includegraphics[width=\linewidth]{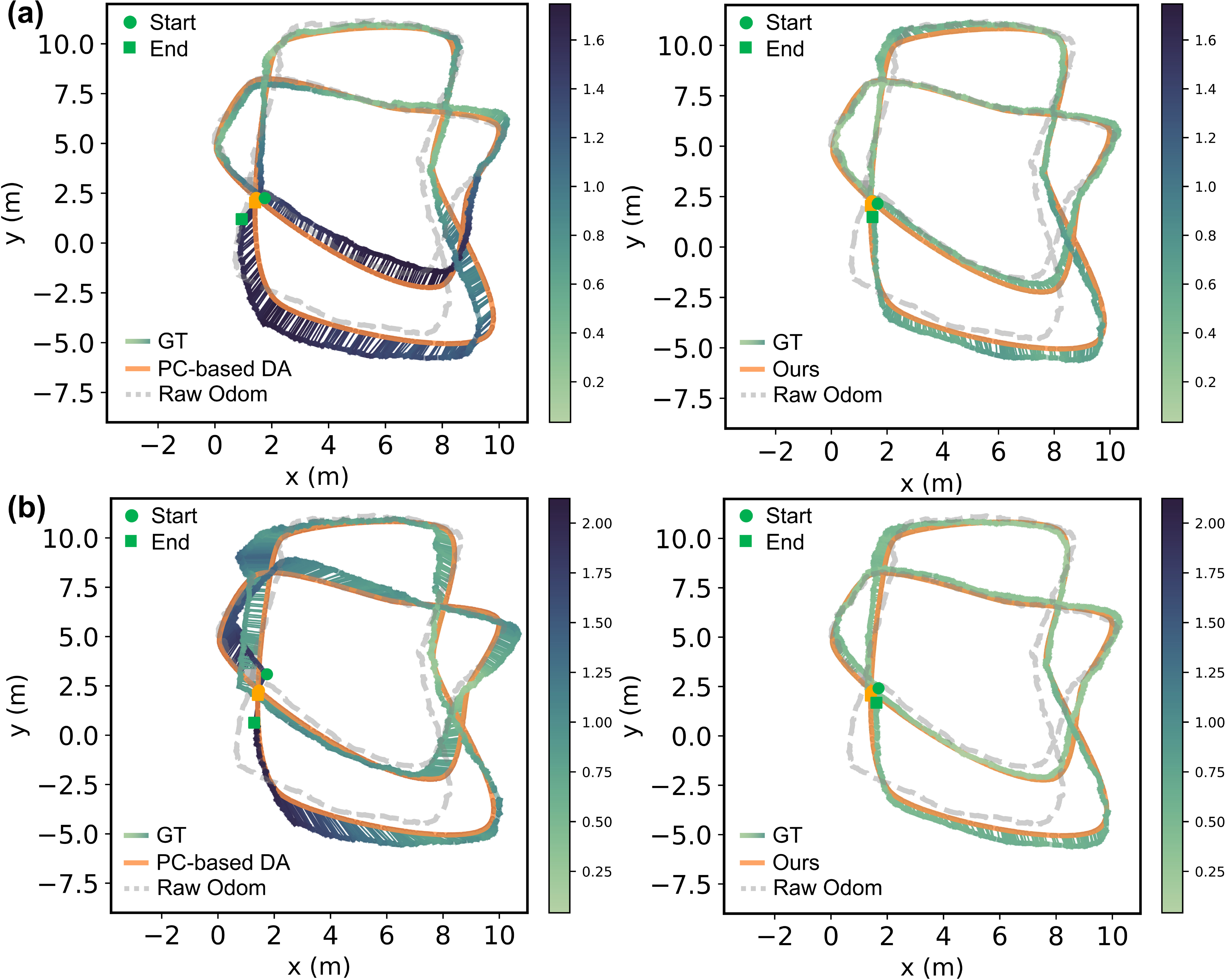}
    \caption{\textbf{Comparison of estimated trajectories on synthetic sequences with point cloud-based (PC-based) and NeuSE-based (Ours) data association.} \textbf{(a)} Results on planar multi-category sequence. \textbf{(b)} Results on non-planar multi-category sequence. Trajectory color variation indicates ATE distribution. Our approach with NeuSE-based data association consistently improves external noisy odometry by encoding full object shapes to handle partial observations, outperforming the point cloud-based method in both sequences. The point cloud-based approach struggles more on the non-planar sequence due to the more complex occlusion patterns and the reduced observation overlap from a mix of upright and laid-down objects.}
    \label{da_traj_viz}
\end{figure}

In this section, we test the ability of NeuSE for robust data association when dealing with such noisy odometry measurements. We generated noisy odometry data from the ground truth camera trajectories for both the multi-category planar and non-planar sequences (shown as the gray dashed trajectories in \autoref{da_traj_viz}). With ground truth data association available in the synthetic data, we apply our NeuSE-based object SLAM pipeline to these noisy odometry measurements and compare the data association accuracy (of partially observed object point cloud observations) and camera pose estimation results with those obtained using the commonly adopted point cloud overlap data association scheme. Our goal is to assess whether the proposed NeuSE-based SLAM framework will have robust data association and localization performance under noisy odometry, in addition to the robustness to viewing angle variations and occlusions already shown in the previous section.

For the point cloud overlap baseline, we modify the proposed NeuSE-based object SLAM pipeline by replacing NeuSE's data association step with a method that finds the associated object, $O_k^*$, with the maximum overlap between the accumulated point clouds of previously seen objects, ${\bf P}_{O_i}$, and the current partial point cloud ${\bf P}$ projected into the world frame. This projection uses the latest camera pose estimate and external odometry measurements to transform the partial point cloud from the local to the world frame. The nearest neighbor search is then performed as follows:
$$
\begin{aligned}
 O_{k^*} &=  \mathbb{I}(R_{k^*}>\delta_{overlap}), \\
 R_{k^*} &= \argmax_{k \in \{1, 2, \dots, K\}} \frac{N_{nn,k}}{|\mathbf{P}|},\\
 N_{nn,k} &= \sum_{\mathbf{p}^{(i)} \in {\bf P}} \mathbb{I}\left( \min_{\mathbf{p}_{O_k}^{(j)} \in \mathbf{P}_{O_k}} \|\mathbf{p}^{(i)} - \mathbf{p}_{O_k}^{(j)}\|_2 \leq \delta_p \right).
\end{aligned}
$$
Here, $\delta_p=0.01$~m is the threshold to determine whether a point in the partial point cloud has a nearest neighbor in the reference one. We compute the nearest neighbor overlap ratio $R_{k}$ relative to the size of the observation point cloud for each of the $K$ object point clouds in the library and find the maximum overlap ratio $R_{k^*}$. If $R_{k^*}$ exceeds the overlap threshold $\delta_{overlap}=0.1$, the associated object is found.

\begin{table*}
\centering
\caption{RMSE of ATE and translational RPE of camera trajectory estimates with point cloud-based and NeuSE-based data association. Best results are marked in bold.}
\label{rmse_da}
\small
\scalebox{1}{
\begin{tabular}{lcccccc}
    \toprule
      &  \multicolumn{3}{c}{\bfseries Planar} &  \multicolumn{3}{c}{\bfseries Non-planar} \\
      \cmidrule(lr){2-4} \cmidrule(lr){5-7} 
      &  Raw Odometry & PC-based & Ours  & Raw Odometry & PC-based & Ours\\
     \midrule
      \textbf{RMSE of ATE (m)}\\
      \quad\quad\quad\quad $1^{\text{st}}$ traversal   &  0.406 & 0.301 & \textbf{0.236} & 0.412 &0.416 & \textbf{0.164}\\
      \quad\quad\quad\quad  $2^{\text{nd}}$ traversal  & 0.453& {0.629} & \textbf{0.398} & 0.456 & 0.628 & \textbf{0.339} \\
      \quad\quad\quad\quad Full                   & 1.064 & 0.917 & \textbf{0.389} & 1.083 & 1.01 & \textbf{0.348}\\
      \midrule
      \textbf{RMSE of Trans RPE (m/f)}\\
      \quad\quad\quad\quad Full & 0.035 & 0.054 & \textbf{0.027} & 0.035 & 0.065 & \textbf{0.024}\\
    \bottomrule
\end{tabular}
}
\end{table*}

We present the data association accuracy for the point cloud overlap baseline (PC-based) and our NeuSE-based method (Ours) in \autoref{da_accuracy}, along with the camera trajectory estimation results in \autoref{rmse_da} and \autoref{da_traj_viz}. 

As shown in \autoref{da_accuracy}, the NeuSE-based data association maintains consistently high accuracy throughout the trajectory, from the first traversal before object changes to the end after changes occur. In contrast, the point cloud-based method experiences a significant drop in accuracy, from slightly above 80\% to around 37\% after the second traversal. This decline is partly due to the low object observation overlap between the first and second traversals, as many objects are observed from nearly opposite views (e.g., objects on the second through ninth tables in \autoref{sim_data_overview}(b)). Additionally, the point cloud-based approach is highly sensitive to external odometry noise, as inaccurate odometry can project partial point clouds to incorrect locations, leading to false associations through overlap comparison.  In contrast, NeuSE achieves accurate data association with no reliance on external odometry and robustness to low observation overlap, leveraging \sethree~equivariance for potential projection transformations while embedding full shape and scale information within a fixed-size latent code.

As a result, NeuSE also leads to a more lightweight data association process. After two rounds of candidate filtering based on latent code scale and shape similarity, most observations are narrowed down to 1-2 potential candidates. A final spatial proximity comparison between the projected partial point cloud centers, requiring little computation, allows NeuSE to complete data association and camera constraint generation at the reported speed of 20~Hz. On the other hand, the point cloud-based method accumulates a growing number of increasingly large object point clouds over time, incurring higher memory usage and longer computation time, particularly due to the expensive nearest neighbor search for overlap computation. By the end of the second traversal, when revisiting 75 objects, each object observation can take up to 0.07 seconds for nearest neighbor comparison with objects in the library, with each library object’s voxel-downsampled point cloud containing about 2,500 points. This creates a significant speed bottleneck, reducing data association performance to 1-3~Hz in our current setting.

The benefit of NeuSE-based data association is further reflected in camera pose estimation.  As shown in \autoref{rmse_da}, NeuSE significantly outperforms the point cloud-based method, yielding much lower ATE and RPE values. Moreover, the point cloud-based method performs worse on the non-planar sequence than the planar one, due to the more complex occlusion patterns and the reduced observation overlap from a mix of upright and laid-down objects. This indicates the advantage of NeuSE's full-shape encoding ability in handling partial observations. The second column of \autoref{da_traj_viz} further emphasizes this, showing how NeuSE corrects heavily drifted odometry in the lower half of the trajectory (around tables 2, 4, 6, and 8 in \autoref{sim_data_overview}(b)), resulting in a 65\% improvement in ATE RMSE values and better alignment with the ground truth. This demonstrates NeuSE's ability to perform robust data association and the potential to enhance camera localization with noisy external odometry.

\subsubsection{Robustness to occlusion and depth noise.} 
In mobile robotic tasks, partial object observations arise not only from changes in viewing angles but also from occlusions by nearby objects. Since NeuSE relies on partial point clouds for object inference, and hence for data association and camera pose constraint generation (as discussed above), the occlusion level and the quality of depth measurements are crucial. To assess the impact of occlusion and depth noise, we conduct ablation studies by systematically varying their levels and observing how NeuSE's equivariance responds in localization and shape reconstruction performance.

\begin{figure}[!htb]
    \centering
    \includegraphics[width=\linewidth]{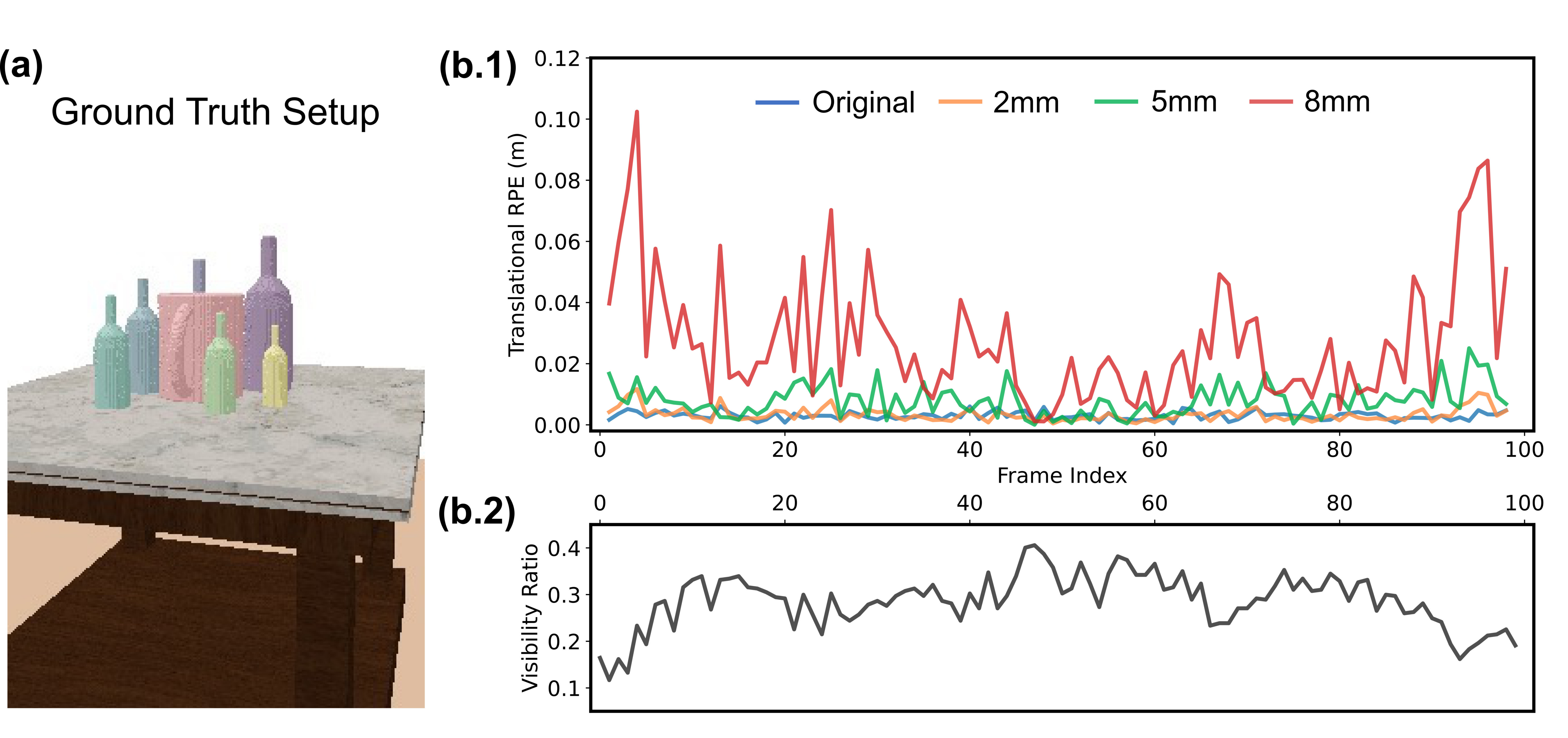}
    \caption{\textbf{(a)} Ground truth scene setup: A \textcolor{pink}{mug} is centered with six bottles of increasing size around it. The camera orbits from the largest \textcolor{purple}{purple bottle} to the smallest \textcolor{yellow}{yellow bottle}, creating varying mug visibility along the trajectory. \textbf{(b)}  Translational RPE curves \textbf{(b.1)} against mug visibility ratio \textbf{(b.2)}, plotted with a shared x-axis of frame index. NeuSE exhibits robustness to occlusion at lower noise levels, while a higher 8~mm noise level produces a U-shaped RPE curve, indicating increased sensitivity to occlusion under noisier depth data.}
    \label{occlusion_data_overview}
\end{figure}

\begin{figure}[!htb]
    \centering
    \includegraphics[width=\linewidth]{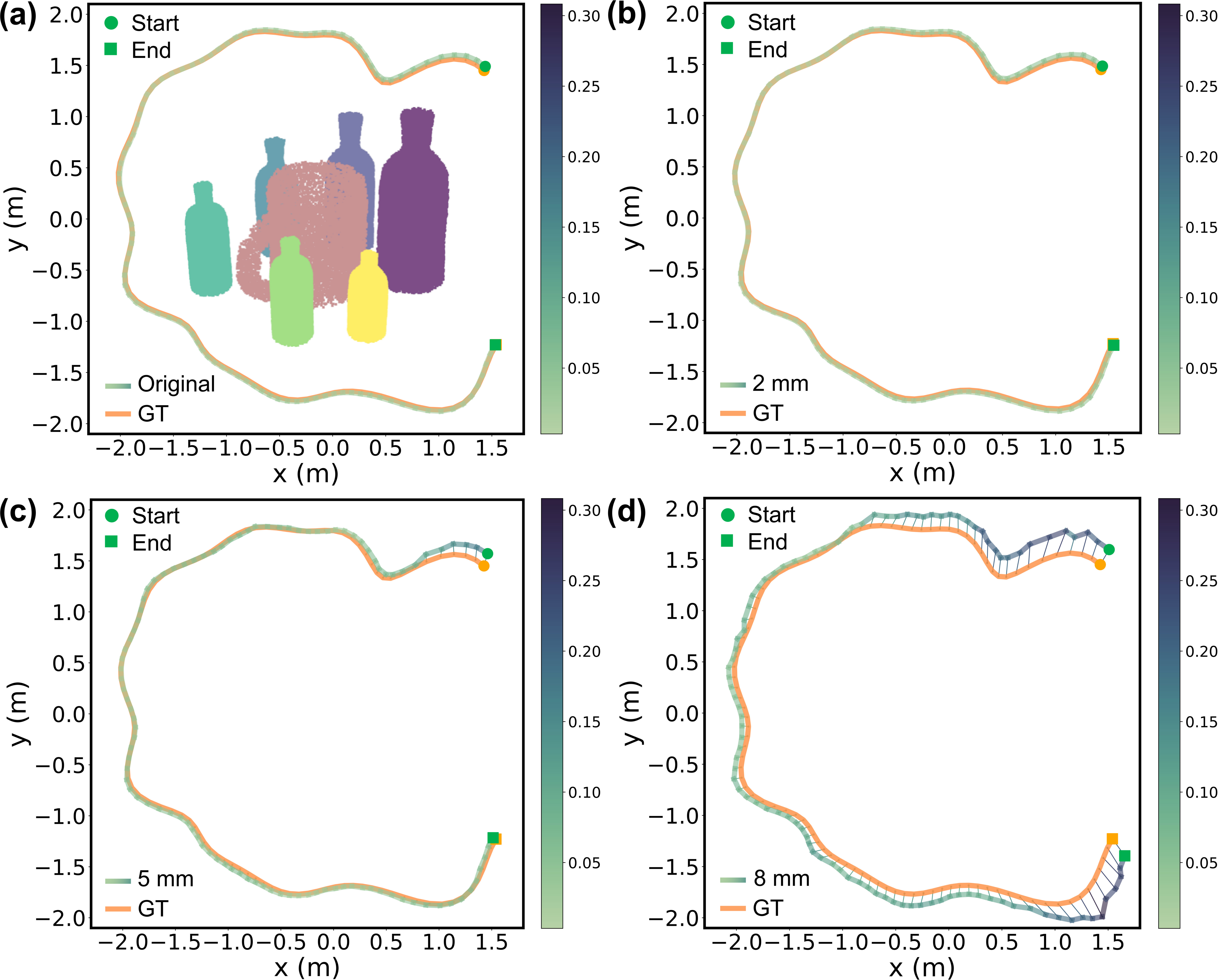}
    \caption{\textbf{Visualization of estimated trajectories against ground truth (GT) for the original and corrupted data at varying noise levels. }Trajectory color variation indicates ATE distribution.   \textbf{(a)} also shows object reconstructions, illustrating the camera’s approximately circular path around the object setup. From \textbf{(a)} to \textbf{(d)}, NeuSE remains robust at lower noise levels (2~mm and 5~mm), recovering camera trajectories that closely align with the ground truth. At the higher noise level (8~mm), the estimated trajectory becomes jagged, especially near the start and end where mug visibility is low, indicating NeuSE's sensitivity to occlusion under high depth noise.}
    \label{occ_ate_viz}
\end{figure}

In line with this goal, we simulated a tabletop scene in PyBullet with a central mug surrounded by six bottles of increasing size. A camera circled around the setup, generating a trajectory that started near the purple bottle and ended around the yellow one, while consistently keeping the mug's handle in view (see \autoref{occlusion_data_overview}(a)). The decreasing size of the bottles along the way results in varying degrees of occlusion, leading to changes in mug visibility ratio throughout the sequence. We plot the visibility ratio of the mug across the camera trajectory in \autoref{occlusion_data_overview}(b.2), where we use nearest-neighbor occupancy calculation between the mug observation and the mug model points to compute the visibility ratio. The trajectory starts near the biggest purple bottle with a low visibility ratio of around 16\% and ends around the smallest yellow bottle. Throughout the trajectory, as the camera moves, the mug visibility ratio first increases and then decreases due to changes in bottle size and viewing angle, covering various visibility levels ranging from 12\% to 41\%.

To examine the effects of depth noise on NeuSE’s performance, we simulate noise values according to the distance-dependent noise characteristics of commercial depth sensors, such as Azure Kinect and RealSense cameras~\citep{s21020413,s21227770}. We vary the sensors' maximum depth noise value, $\sigma_{max}$, at different levels and employ a simple distance-dependent Gaussian noise model to generate noisy depth measurements, $d_{\text{noisy}}$, as follows:
\begin{equation}
    \begin{aligned}
d_{\text{noisy}}(u,v) &= d(u,v) + \eta(u,v) \\
\eta(u,v) &\sim N(0, \sigma(u,v)) \\
\sigma(u,v) &= \sigma_{\text{max}} \cdot \frac{d(u,v)}{d_{\text{max}}},
\end{aligned}
\end{equation}
where $d(u,v)$ is the depth value at pixel location $(u,v)$, and $d_{max}$ is the maximum depth value of the objects observed in the current frame.  The depth noise $\eta (u,v)$ follows a zero-mean Gaussian distribution with a standard deviation $\sigma$, which scales proportionally with the depth value at the given location relative to the maximum depth, $d_{max}$.

Since depth measurement noise increases proportionally with distance, we corrupted the data based on noise levels across depth sensing ranges, following the $\sigma_{max}$ values of commercial depth sensors for close (within 3~m), medium (3-5~m), and far (beyond 5~m) ranges. In particular, we applied noise levels of $\sigma_{\text{max}}$ as follows: 2~mm and 5~mm to represent depth noise for close-range measurements, 8~mm for mid-range measurements, and 1~cm for far-range measurements.  In our testing data, $d_{\text{max}}$ typically fell within the close-range depth measurements with distances of around 2~m, making the 2~mm and 5~mm noise levels the most meaningful. 

\begin{table}
    \centering
    \caption{RMSE of ATE and translational RPE on synthetic sequences of different depth noise levels.}
    \label{occlusion_stats_transposed}
    \small
    \resizebox{\columnwidth}{!}{  
    \begin{tabular}{lcc}
        \toprule
        & \textbf{RMSE of ATE (m)} & \textbf{RMSE of Trans RPE (m/f)} \\
        \midrule
        \textbf{Original} & 0.019 & 0.003 \\
        \textbf{2~mm} & 0.021 & 0.004 \\
        \textbf{5~mm} & 0.055 & 0.009 \\
        \textbf{8~mm} & 0.136 & 0.034 \\
        \textbf{1~cm} & 37.398 & 7.697 \\
        \bottomrule
    \end{tabular}
    }
    \label{occ_rpe_ate_transposed}
\end{table}

Below, we report the performance of our approach under both object occlusion and depth noise.

\begin{figure}[!htb]
    \centering
    \includegraphics[width=\linewidth]{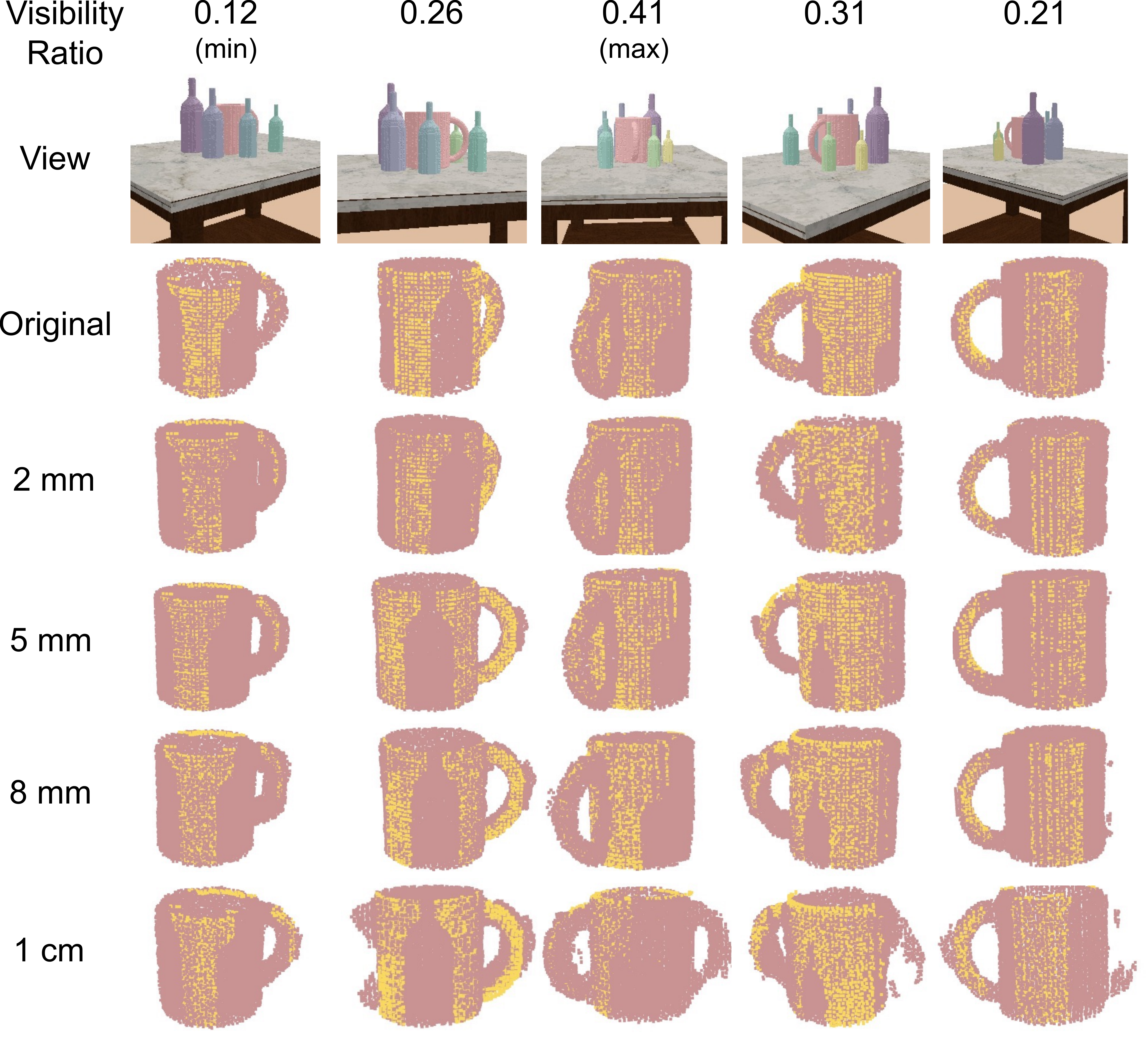}
    \caption{\textbf{Multi-view mug reconstruction with varying visibility and noise levels}. Ground truth views are arranged along the camera path, including frames with maximum, minimum, and intermediate mug visibility to highlight the visibility variation along the way. Yellow points represent the observed partial point clouds. NeuSE demonstrates strong robustness at lower noise levels (2~mm and 5~mm), accurately reconstructing the full mug shape in alignment with partial observations, regardless of visibility changes. At the 8~mm noise level, NeuSE's sensitivity to occlusion increases, resulting in artifacts like a falsely shaped square handle and residual points around the mug body in low-visibility frames (first and last columns). At the 1~cm level, where NeuSE fails to recover the camera trajectory, reconstructions degrade further, producing entirely incorrect mug shapes, as the impact of noise dominates over that of occlusion.}
    \label{occ_mug_rec}
\end{figure}

\paragraph{Occlusion.}Taking the visibility ratio curve in \autoref{occlusion_data_overview}(b.2) as the reference, in \autoref{occlusion_data_overview}(b.1), we observe that the RPE values remain relatively stable throughout the trajectory at low to medium noise levels, even as visibility fluctuates, which corresponds to the closely aligned camera trajectory estimates in \autoref{occ_ate_viz}(a)-(c). Higher RPE peaks occur primarily at the start and end of the trajectory, where the visibility ratios are the lowest. In the middle of the trajectory, where the camera moves through areas with smaller and more spaced-out bottles, visibility ratio increases, resulting in smaller and smoother RPE fluctuations.  The mug reconstructions shown in \autoref{occ_mug_rec} further illustrate NeuSE’s ability to recover the full mug shape despite varying levels of occlusion under close to medium range noise levels. This robustness can be attributed to NeuSE’s training data design, which incorporates diverse occlusion patterns, enabling it to handle partial observations effectively.

\paragraph{Depth noise.}While NeuSE handles occlusion well, noise has a greater impact on its performance. NeuSE shows strong robustness to close-range noise levels, which are the most relevant for our target applications. At a noise level of 2~mm,  NeuSE performs almost on par with the noise-free scenario. This is proven by the similarly low ATE values in \autoref{occ_rpe_ate_transposed} compared to that of Original, the well-aligned trajectory estimates in \autoref{occ_ate_viz}(b),  and the largely overlapping RPE curves between Original and 2~mm in \autoref{occlusion_data_overview}(b.1), all of which indicate that the added low-level noise has little impact on the trajectory estimation accuracy.

As the noise level increases to 5~mm, representing the higher end of close-range depth sensing noise,  NeuSE continues to perform well across most of the trajectory. However, it becomes affected in regions with low visibility, such as at the beginning of the trajectory around the least visible part, as shown in \autoref{occ_ate_viz}(c). At higher noise levels, such as 8~mm and 1~cm, NeuSE struggles to handle the noisy measurements. The once smooth camera pose estimates seen at lower noise levels become inconsistent, and NeuSE becomes more sensitive to occlusion, as reflected in the more erratic RPE curve in red (\autoref{occlusion_data_overview}(b.1)). This degradation is also evident in the jagged and drifted trajectory in \autoref{occ_ate_viz}(d) for the 8~mm noise level, while for the 1~cm noise level,  NeuSE fails to recover a valid trajectory.

NeuSE’s robustness to close-range noise is also demonstrated by the mug reconstructions,  which remain complete and accurate at noise levels up to 5~mm, as shown in the first three rows of \autoref{occ_mug_rec}. However, as noise levels increase, NeuSE’s sensitivity to occlusion becomes more apparent. For instance, at the 8~mm noise level, the handle shape of the mug is reconstructed incorrectly, and at 1~cm, the entire mug shape is falsely recovered, further showcasing the challenges posed by higher depth noise.
\subsection{Real-world sequences}
\begin{figure}[!htb]
    \centering
    \includegraphics[width=\linewidth]{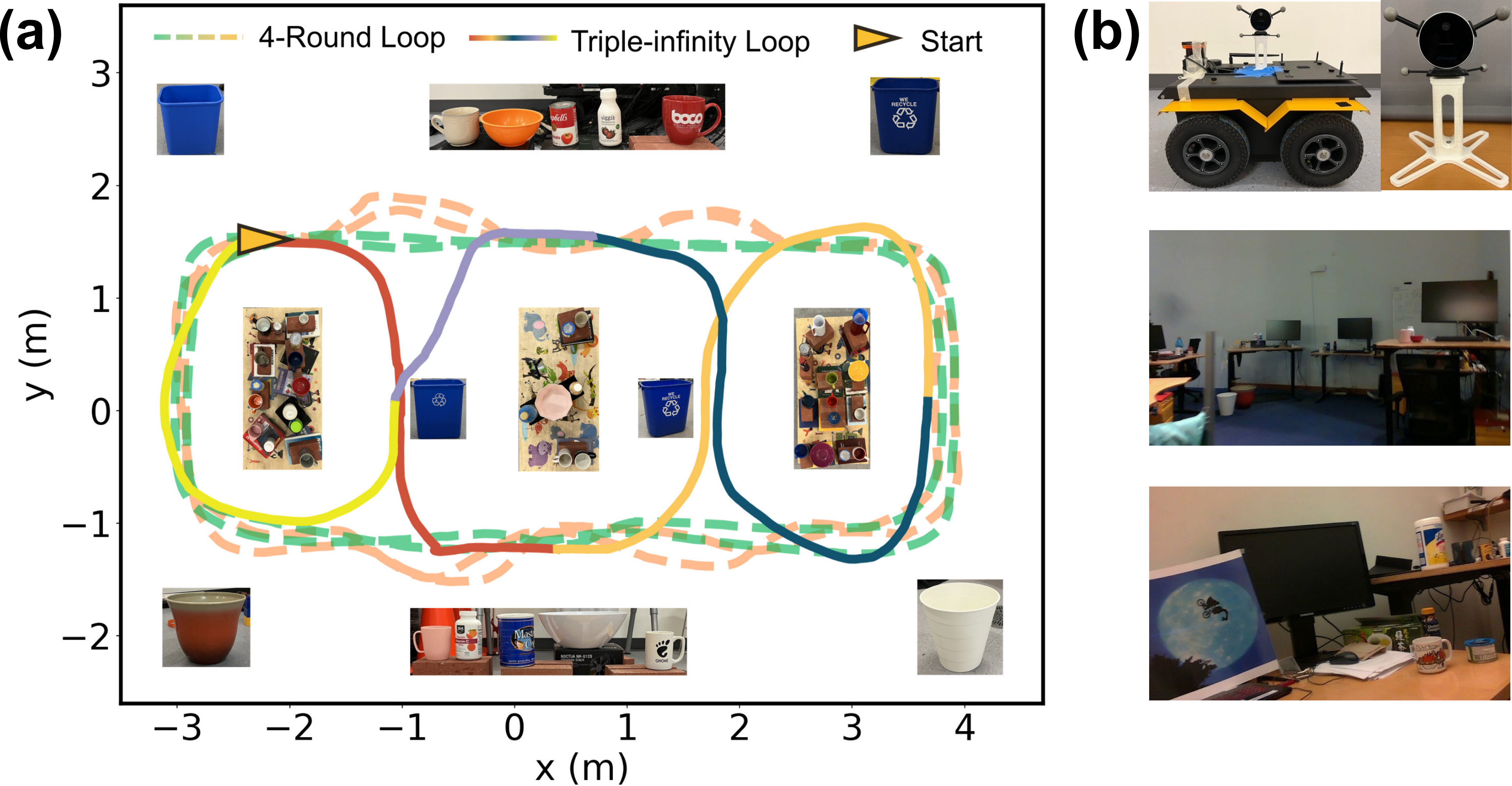}
    \caption{\textbf{Real-world data overview}. \textbf{(a)} Object layout and the ground truth camera trajectory of the controlled environment. Object changes occur at each junction of the colored trajectory segments. \textbf{(b)} Camera setup and open lab (second row) and indoor office (third row) environments for the hand-held sequences.}
    \label{real_data_overview}
\end{figure}
Having studied the properties of NeuSE in simulation, we now extend our evaluation to real-world robot motion, where diverse object occlusion patterns and varying camera angles are common. As the robot moves, objects of interest may intermittently go out of sight. Building on our previous results showing NeuSE's ability to handle noisy external odometry, this section further demonstrates its practical application in long-term, real-world scenarios. We emphasize its ability to \emph{complement} other SLAM systems in both controlled and uncontrolled settings, showcasing its smooth transition from simulation training to real-world deployment. NeuSE consistently produces globally accurate trajectory estimates, even in the presence of challenges such as temporal scene inconsistency, densely placed objects with varying shapes and sizes, freeform camera motion, and potential drift due to tracking loss.

\subsubsection{Data preparation and choice of baselines.}
The real-world data were collected in two different settings with object instances from five categories: mug, bottle, bowl, can, and trash bin. The settings were as follows: (1) a controlled environment containing 49 object instances densely arranged on five tables within a $7~\text{m} \times 5~\text{m}$ space (\autoref{real_data_overview}(a)), and (2) a less constrained lab-office environment consisting of two interconnected areas, a smaller indoor space and an open lab space connected by a corridor. In this setting, 45 object instances were distributed on the floor and across 16 tables of varying heights (\autoref{real_data_overview}(b)).

\begin{table*}
    \centering
    \caption{RMSE (m) of the Absolute Trajectory Error on real-world sequences.  Best results for each trajectory are marked in bold.}
    \label{rmse}
    \resizebox{\linewidth}{!}{
    \begin{tabular}{c|c|ccccccccc}
    \toprule
    \textbf{Method} & \textbf{Type} & \textbf{4-Round: $1^{\text{st}} - 2^{\text{nd}}$ round} 
    & \textbf{4-Round: $2^{\text{nd}}-3^{\text{rd}}$ round} 
    & \textbf{4-Round: $3^{\text{rd}}-4^{\text{th}}$ round} 
    & \textbf{4-Round: Full} 
    & \textbf{Triple-Infinity} \\
    \midrule
    \multirow{1}{*}{\textbf{CubeSLAM~\citep{yang2019cubeslam}}} 
    & %
    & 0.208 & 0.156 & 0.164 & 0.206 & 0.334 \\
    \multirow{1}{*}{\textbf{EM-Fusion~\citep{strecke2019_emfusion}}} & All Objects Detected 
    & 0.302 & 0.315 & 0.219 & 0.304 & 0.281 \\
    \multirow{1}{*}{\textbf{VOOM~\citep{wang2024icra}}} & 
    & 0.461 & 0.395 & 0.236 & 0.538 & 0.160 \\
    \midrule
    \multirow{4}{*}{\textbf{Obj-only}} 
    & \multirow{1}{*}{Raw Odometry} & 0.628 & 0.904 & 1.022 & 1.248 & 0.591 \\
    & \multirow{1}{*}{Mug-only} & 0.170 & 0.142 & 0.155 & 0.163 & 0.232 \\
    & \multirow{1}{*}{All-object (\sethree-only)} & 0.142 & 0.167 & 0.160 & 0.174 & 0.194 \\
    & \multirow{1}{*}{All-object} & 0.205 & 0.185 & 0.097 & 0.152 & 0.178 \\
    \midrule
    \multirow{3}{*}{\textbf{ORB3-NS}} 
    & \multirow{1}{*}{Base} & 0.105 & 0.138 & 0.116 & 0.128 & 0.161 \\
    & \multirow{1}{*}{+ \sethree-only} & 0.107 & 0.120 & 0.110 & 0.119 & 0.150 \\
    & \multirow{1}{*}{+ Ours} & 0.093 & \textbf{0.102} & 0.096 & \textbf{0.093} & \textbf{0.131} \\
    \midrule
    \multirow{3}{*}{\textbf{ORB3-PW}} 
    & \multirow{1}{*}{Base} & 0.094 & 0.141 & 0.118 & 0.138 & 0.275 \\
    & \multirow{1}{*}{+ \sethree-only} & 0.112 & 0.126 & 0.115 & 0.124 & 0.153 \\
    & \multirow{1}{*}{+ Ours} & \textbf{0.088} & 0.103 & \textbf{0.094} & 0.095 & 0.132 \\
    \bottomrule
    \end{tabular}
    }
    \end{table*}

In the first controlled setting, 14 to 16 objects were added, removed, or relocated to create two distinct object arrangements. An RGB-D dataset was recorded using a RealSense D515 camera mounted on a Clearpath Jackal robot, following two preset closed trajectories: (1) a four-round peripheral loop around three central tables, with the former two rounds recorded under one object arrangement and the latter two under the other arrangement, and (2) a more challenging triple-infinity loop traversing two central and two side tables, with seven object changes made along the way. Ground truth camera trajectories were obtained using a Vicon motion capture system.

In the second uncontrolled setting, 8 to 9 objects were added, removed, or relocated to create two distinct object layouts. The same RealSense D515 camera, mounted on a tripod, was carried by hand to record RGB-D data along the tables in the scene before and after object changes (please refer to the supplementary video for further details). Two trajectories were collected: (1) a two-round scan along eight indoor tables against the office walls, using top-down and near-far viewing angles, with each round corresponding to one object layout; and (2) a two-round ``freeform'' traversal among 16 tables spanning both indoor office and open lab spaces. The trajectory began indoors, passed through a corridor into the open lab space, and returned back to the office after object changes in both the indoor office and outside lab areas. Motion capture data were available for the indoor office portion, covering 24\% of the trajectory. No motion capture was available in the open lab space, and a ``blackout'' period occurred in the corridor where no objects of interest were visible.

To evaluate the global consistency of camera localization, we adopt ATE as the evaluation metric and compare our approach against the popular state-of-the-art ORB-SLAM3~\citep{campos2021orb} pipeline and three directly deployable object-based SLAM systems: CubeSLAM~\citep{yang2019cubeslam}, EM-Fusion~\citep{strecke2019_emfusion}, and VOOM~\citep{wang2024icra}. Both CubeSLAM and VOOM are built on top of ORB-SLAM2~\citep{murORB2} to conduct joint optimization of object representations and camera poses. CubeSLAM uses cuboids for object representation, while VOOM represents objects with dual quadrics in a hierarchical manner in combination with map points. CubeSLAM assumes a static operating environment (or objects with known motion models, which do not apply in our case), while VOOM is reported to have a certain level of robustness to scene dynamics. EM-Fusion, on the other hand, uses local Signed Distance Function (SDF) object volumes for tracking moving objects and performing camera localization, making it capable of handling dynamic scenes. These approaches serve as baselines to evaluate object SLAM performance and the potential impacts of object changes in the scene. For CubeSLAM, we used the implementation integrated with ORB-SLAM.

\subsubsection{Controlled environments: 4-Round and Triple-Infinity loops.}  The two sequences in controlled settings were captured with objects densely arranged on tabletops and positioned around distant corners. This led to a series of observations featuring variations in both near and far viewing angles, as well as rich occlusion patterns.

As ORB-SLAM3 does not address temporal scene inconsistency, to further explore the effect of object changes on localization performance, we generated two sets of ORB-SLAM3  odometry measurements from its camera trajectory estimates as starting points by running it: (1) non-stop (ORB3-NS) for the whole trajectory, and  (2) piecewise (ORB3-PW) for each trajectory segment with consistent object layout (as shown in \autoref{real_data_overview}(a)). 

Before showcasing NeuSE's ability to complement external SLAM systems, we first need to verify NeuSE’s ability to transfer from simulation to the real world.  We follow the object-only experiments for synthetic data and run Mug-only and All-object on these two real-world sequences. Raw Odometry, generated from RGBD images using Open3D~\citep{Zhou2018} based on photometric and geometric loss~\citep{8237287}, is adopted to sustain system operation when no objects are in sight or associated for generating pose constraints.

\begin{figure}[!htb]
    \centering
    \includegraphics[width=\linewidth]{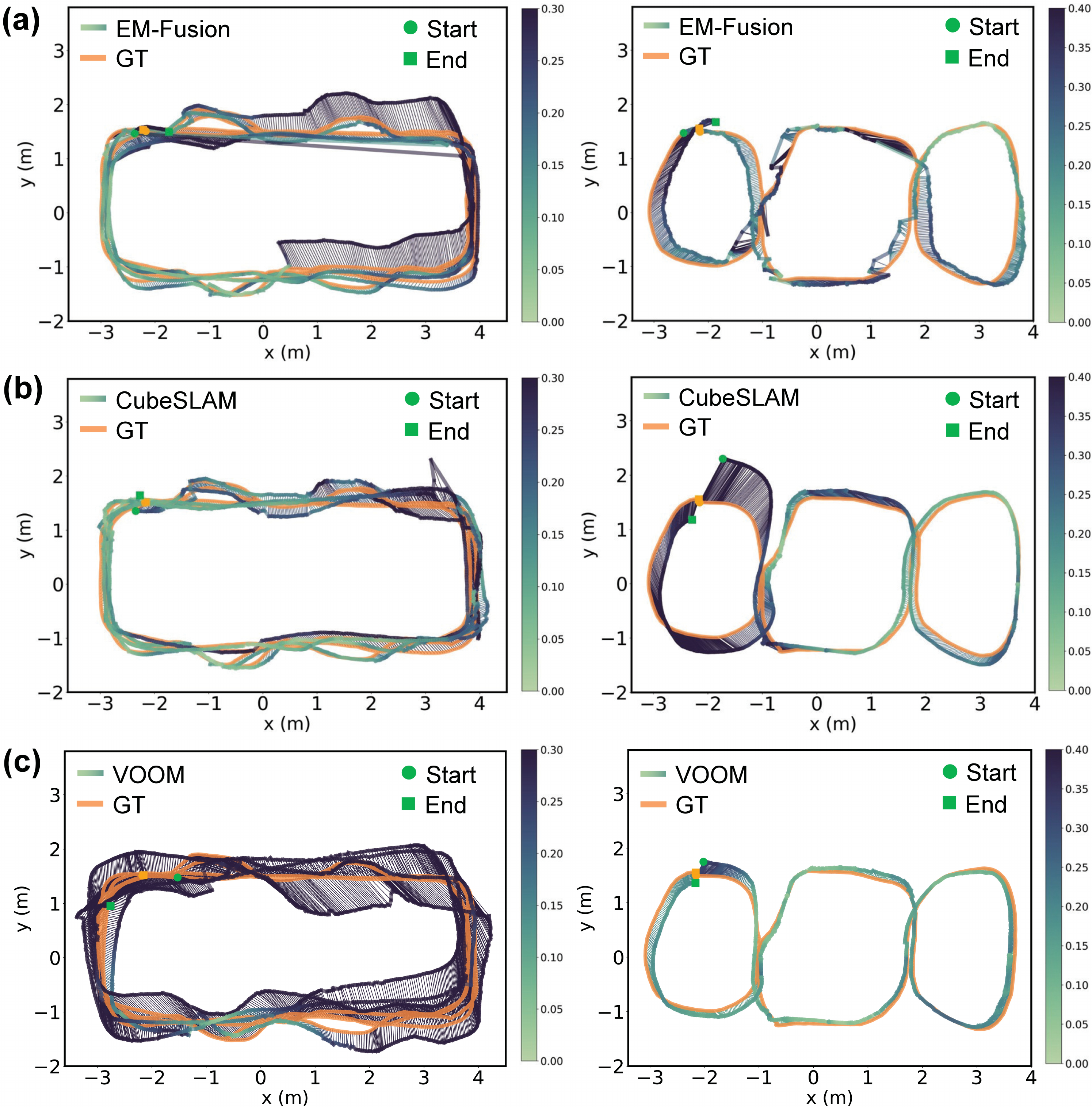}
    \caption{\textbf{Baseline trajectory estimation results against ground truth (GT) on the 4-Round (left) and Triple-Infinity loops (right).} \textbf{(a)} EM-Fusion; \textbf{(b)} CubeSLAM; \textbf{(c)} VOOM. All chosen baselines suffer from either localization drift in the 4-Round trajectory or failure to close the loop in the Triple-Infinity trajectory.}
    \label{traj_baseline}
\end{figure}

\begin{figure*}[!htbp]
    \centering
    \includegraphics[width=\linewidth]{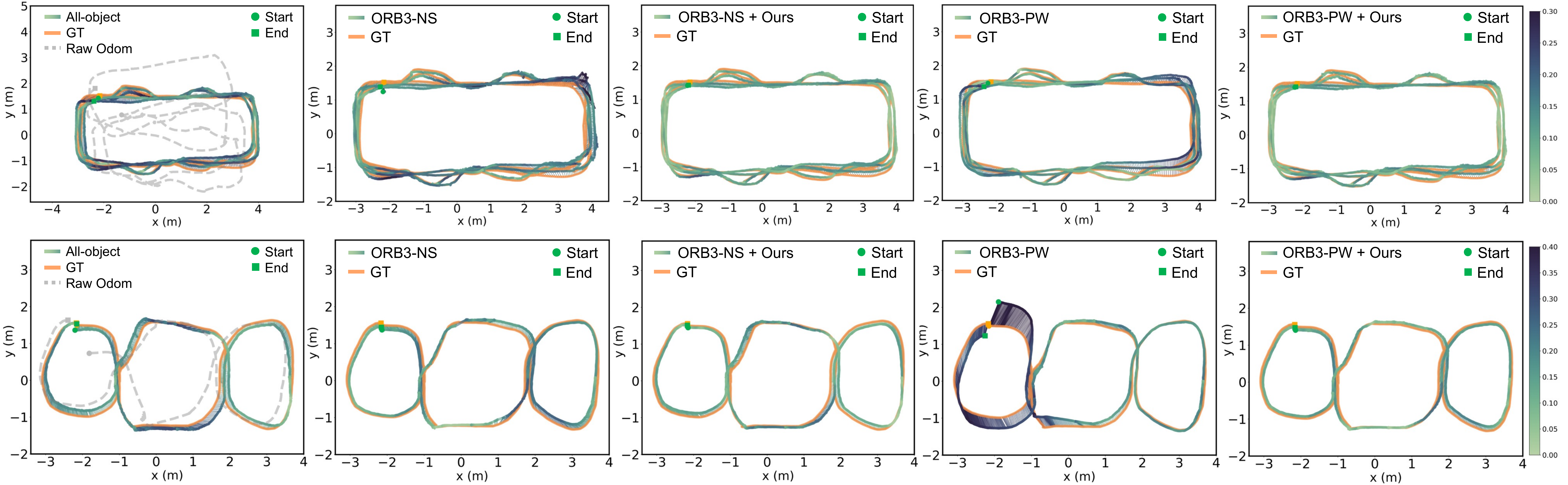}
    \caption{\textbf{Visualization of estimated trajectories against ground truth (GT) on controlled sequences.} Trajectory color variation indicates ATE distribution (color bar on the right).  Top: Estimated trajectories of the  4-Round loop. The integration of our strategy (columns 3 and 5) helps smooth unstable trajectory estimates and improves general estimation accuracy, as showcased by the absence of small spikes (upper right) seen in column 2 and the consistently lighter colors of lower ATE values along the way. Bottom: Estimated trajectories of the Triple-Infinity loop. Our strategy (column 5) successfully corrects the accumulated start and end point drift for ORB3-PW (column 4), leading to an improved and consistent trajectory estimate when revisiting the leftmost table.}
    \label{real_traj}
\end{figure*}

Building on the synthetic sequence experiments, we further investigate whether the advantage of scale equivariance in generalizing to objects of different sizes extends to real-world data. To this end, we also evaluate NeuSE's \sethree-only variant (\sethree-only) across all previously generated odometry sets.

\begin{figure}[!htb]
    \centering
    \includegraphics[width=0.8\linewidth]{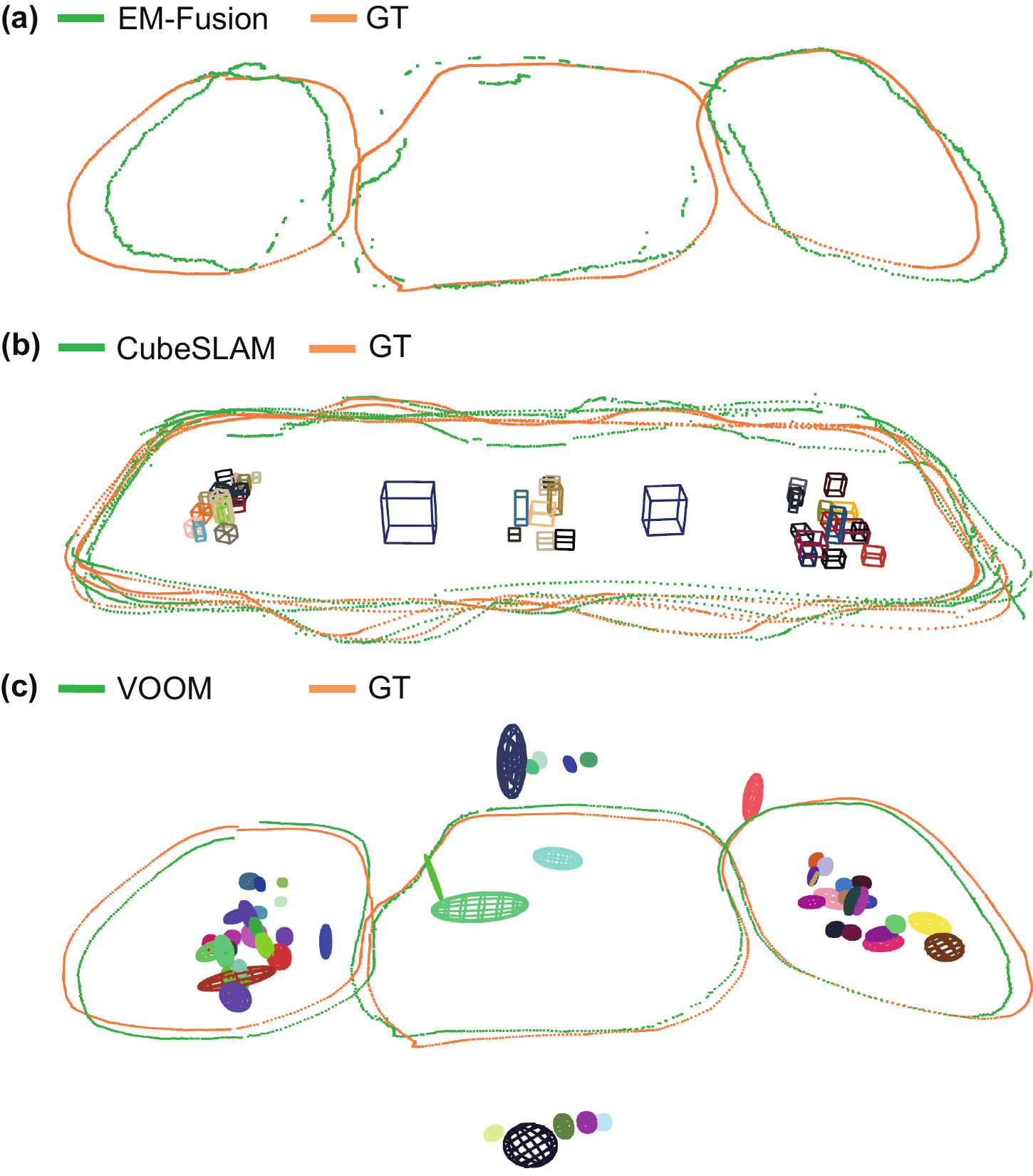}
    \caption{ \textbf{Trajectory estimation and object mapping artifacts in selected baselines.} \textbf{(a)} EM-Fusion undergoes heavy out-of-plane drift and intermittent loss of tracking in the Triple-Infinity loop due to the rapid rotations around corners. \textbf{(b)} CubeSLAM fails to handle object changes effectively, resulting in data association errors, missed detections, drift, and overlapping cuboids, which affect the joint optimization of cuboid and  camera trajectory estimates.  \textbf{(c)} VOOM produces an object map with incorrect object count, scale, and pose estimates, including a green object falsely intersecting with the camera trajectory. Its use of dual quadrics as the object representation struggles to capture the full size of objects from partially occluded point clouds, particularly in cluttered scenes.}
    \label{baseline_reconstruction}
\end{figure}

We present the RMSE of ATE for all estimated trajectories in \autoref{rmse} and visualize them in Figures~\ref{traj_baseline}--\ref{baseline_reconstruction}. 

Although relying on noisy Raw Odometry when objects are unavailable, NeuSE successfully transfers from simulation to the real world, generating accurate camera pose constraints that offset and correct accumulated drift from Raw Odometry, resulting in overall lower ATE RMSE values, as shown in the results of the ``Obj-only'' method in \autoref{rmse} and the first column of \autoref{real_traj}.  This confirms NeuSE's full functionality when applied to real data. Furthermore, the results of ``Obj-only'' in \autoref{rmse} indicate that the All-object approach outperforms the Mug-only approach in both sequences, verifying the effectiveness of our strategy in extracting useful camera pose constraints from covisible ambiguous objects.
 
We compare our approach to three selected object SLAM baselines that use all detected objects in the scene, as shown in \autoref{rmse}. Our approach, All-object, using only NeuSE-based camera pose constraints supported by noisy Raw Odometry, achieves performance nearly on par with VOOM on the Triple-Infinity loop and outperforms the rest of the baselines on both sequences, even though the baselines directly use existing powerful SLAM systems such as ORB-SLAM2. This demonstrates NeuSE's advantage in facilitating lightweight and robust localization in real-world sequences with scene inconsistency.

Notably, CubeSLAM fails to close the loop in \autoref{traj_baseline}(b) after returning to the leftmost table with object changes in the Triple-Infinity loop. In terms of object estimates, it also produces multiple missed, drifted, or falsely overlapped cuboid estimates shown in \autoref{baseline_reconstruction}(b). This could be attributed in part to the difficulty of accurate cuboid detection with diverse occlusion patterns. Moreover, assuming a static environment, CubeSLAM struggles to address object changes within the two sequences, leading to errors in cuboid association and estimation between old and new objects in neighboring areas.  This results in false camera-cuboid geometric constraints and ultimately affects the joint optimization of object cuboids and camera trajectory. 

 Meanwhile, EM-Fusion, as shown in \autoref{traj_baseline}(a) and \autoref{baseline_reconstruction}(a), gives subpar, bumpy, and drifted trajectory estimates. While it can handle scene layout changes at sequence segment intersections, EM-Fusion suffers from lower tracking accuracy due to the accumulated drift from limited object overlap. In addition, since EM-Fusion was initially tested on tabletop scenes, it requires a coarser SDF background volume resolution to prevent memory exhaustion in our larger multi-table scenario, which further reduces the accuracy of its camera tracking.

VOOM performs best on the Triple-Infinity loop, with the smallest start-to-end drift among the three baselines, but struggles significantly on the 4-Round loop, as shown in \autoref{traj_baseline}(c). The drift worsens with each round, especially after object changes, such as the removal of large objects (e.g., trash bins) between the central tables, which introduces confusion in object odometry and mapping due to the lack of mechanisms to handle such changes. VOOM’s use of dual quadrics for object representation makes it difficult to accurately capture full object sizes from partially occluded point clouds, particularly when objects are densely placed on the tables. This results in inaccurate object initialization and odometry, which further hinders the mapping process. An example of this is the green object in \autoref{baseline_reconstruction}(c), which falsely intersects with the camera trajectory.

We next illustrate how NeuSE can be directly combined with other SLAM systems for additional improvements. First, in the last column of \autoref{rmse}, NeuSE helps improve both ORB3-NS and ORB3-PW, with our combined object SLAM system outperforming all methods including the VOOM baseline on the Triple-Infinity loop. We then compare the ``Base'' and ``+ Ours'' variants of ORB3-NS and ORB3-PW methods. We observe consistent improvement in terms of the RMSE of ATE values along two sequences when integrating our proposed strategy using all objects of interest (``Ours'') with the vanilla ORB-SLAM3 measurements (``Base''), as also supported by the lighter colors in  columns 3 and 5 of \autoref{real_traj} compared to columns 2 and 4. Furthermore, NeuSE enables robust data association and manages to prevent the occurrence of unstable tracking, for example, the small spikes in column 2 of \autoref{real_traj} for the 4-Round trajectory.   

\begin{figure}[!htb]
    \centering
    \includegraphics[width=\linewidth]{figures/20.pdf}
    \caption{\textbf{Trajectory estimation against ground truth (GT) of NeuSE's \sethree-only variant on controlled sequences.}    The \sethree-only variant has difficulty handling objects with sizes outside the training data distribution, leading to less accurate camera pose constraints, such as in regions with the larger trash bins. This limitation is highlighted by the cyan blocks in \textbf{(a)} for the 4-Round loop and \textbf{(b)} for the Triple-Infinity loop.}
    \label{se3_traj_holodeck}
\end{figure}

The greatest RMSE improvement in \autoref{rmse} is observed from ORB3-PW + Ours on the Triple-Infinity trajectory. Our proposed strategy helps decrease the RMSE by 52\%  from 0.275~m to 0.132~m while aligning the start and end points with better trajectory accuracy when revisiting the leftmost table. In this way, with abundant objects visible for deriving camera pose constraints, ORB3-PW + Ours  outperforms ORB3-NS and is almost on par with ORB3-NS + Ours, despite receiving less global loop closing constraints from ORB3-PW than ORB3-NS. Considering the little scene overlap within each of the four trajectory segments, this notable improvement highlights the critical role of our strategy in constraining camera pose estimates in short and longer ranges, especially when insufficient loop closing (e.g., throughout ORB3-PW) is performed by the external SLAM system. 

Our strategy also demonstrates robustness in handling scene changes, despite the less significant improvement in the 4-Round loop due to the numerous loop closure opportunities offered by ORB-SLAM3. Columns 3-6 of \autoref{rmse} present the RMSE values of ORB3-NS at different stages of  the 4-Round loop, as the sequence proceeds with object layout transition. Note that ORB3-PW does not run between the second and third rounds, with the corresponding values included for comparison only. When object changes happen at the intersection of the second and third rounds, ORB3-NS is clearly affected and shows an RMSE of ATE value jump from 0.105~m to 0.138~m. On the contrary, our effective data association based on full object scale and shape similarity as well as spatial proximity allows ORB3-NS + Ours to maintain a relatively steady and lower RMSE of ATE around 0.10~m,  reducing the overall RMSE of ATE by 27.3\% from 0.128~m to 0.093~m.

We further compare our approach with its \sethree-only variant. We observe that while it still leads to more accurate camera trajectory estimates when used alongside other SLAM measurements, its improvements are less significant compared to the full \sethree-~and scale-equivariant version. This is indicated by the modest reduction in the RMSE of ATE values in \autoref{rmse} across various types of trajectories. This degraded performance can be attributed to its limited ability to handle objects whose sizes fall outside the distribution of the training data, leading to less accurate camera pose constraints. This limitation is further illustrated in \autoref{se3_traj_holodeck}, which shows greater deviations from the ground truth in both the 4-Round and Triple-Infinity loops, particularly in regions involving larger objects, such as trash bins, as highlighted by the two cyan blocks. This again demonstrates the advantage of incorporating scale equivariance, which enhances NeuSE's generalization ability to better transfer category-level geometric knowledge to instances that are smaller or larger than those encountered during training.

\begin{figure}[!htb]
    \centering
    \includegraphics[width=\linewidth]{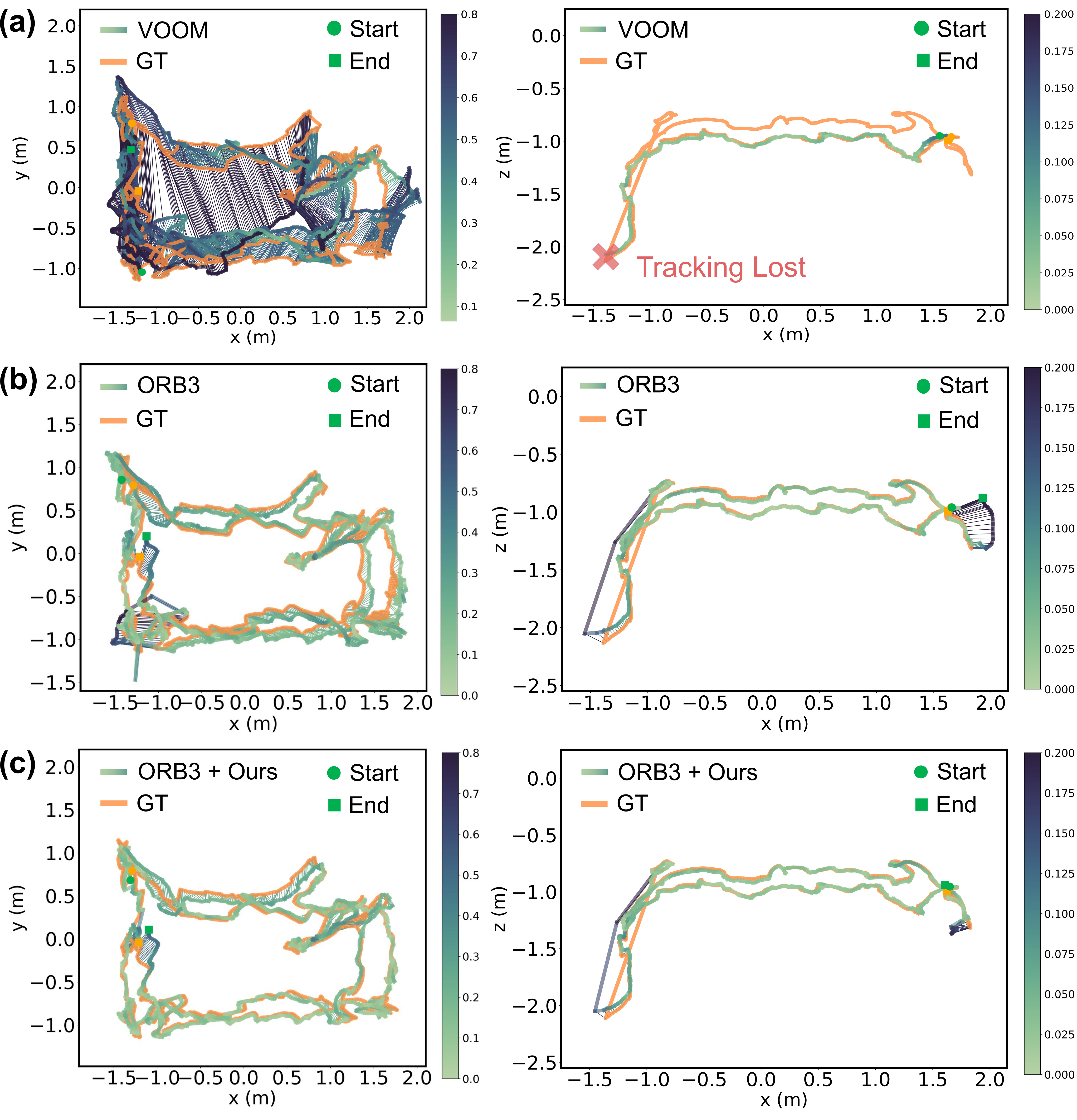}
    \centering
    \caption{\textbf{Comparison of estimated trajectories against ground truth on hand-held sequences.} Column 1: Top-down views of the full Indoor sequence. Column 2: Side views of the start and end segments of the Freeform sequence, where ground truth camera poses are available. \textbf{(a)} VOOM completes the Indoor loop but suffers from significant drift and loses tracking early in the Freeform loop, indicating its limited ability to handle more drastic variations in viewing angles. \textbf{(b)--(c)} NeuSE-derived camera pose constraints consistently enhance ORB-SLAM3's localization performance across both sequences, despite the more complex hand-held motion and varied viewing angles. The stronger alignment between ORB3 + Ours and the ground truth, particularly in the Freeform loop in \textbf{(c)}, highlights our method's effectiveness in facilitating relocalization after tracking loss, such as when the camera re-enters the office space following the corridor ``blackout.''  }
    \label{lab_traj_viz}
\end{figure}
\subsubsection{Uncontrolled environments: Indoor and Freeform loops.} While our strategy performs well in controlled settings, where viewing angles and occlusion patterns vary, the camera motion in previous sequences is relatively steady and constrained, and the environment contains abundant objects. To further evaluate the proposed approach in more dynamic real-world conditions, we recorded the more challenging Indoor and Freeform loops with a hand-held camera, allowing full 3D motion and variable speeds. The experimental setup mimicked a lab-office environment, with objects of varying sizes scattered across tables of different heights and on the floor. To better test NeuSE's object characterization capabilities with \sethree~and scale equivariance, we introduced objects not included in the controlled sequences to further increase size variation. For instance, two small white square jars were added to the trash bin category due to their shape similarity to a square trash bin (see the second column of \autoref{object_scale_comparison}(a)). Additionally, we incorporated a ``blackout'' corridor phase to test NeuSE's data association and relocalization capabilities.
\begin{table}[t]
    \centering
    \small
    \caption{RMSE (m) of the Absolute Trajectory Error on hand-held sequences. Best results for each trajectory are marked in bold.}
    \label{lab_res}
      \resizebox{\columnwidth}{!}{
    \begin{tabular}{@{}lccccc@{}}
        \toprule
          & VOOM & ORB3 & ORB3 + \sethree-Only & ORB3 + Ours \\
         \midrule
          \textbf{Indoor}\\
          \quad\quad\quad\quad Before & 0.724 &0.180& 0.174 & \textbf{0.130} \\
          \quad\quad\quad\quad After  & 0.179 &0.201& 0.183 & \textbf{0.168} \\
          \quad\quad\quad\quad Full   & 0.696 &0.192& 0.181 & \textbf{0.151} \\
          \midrule
          \textbf{Freeform (partial)}\\
          \quad\quad\quad\quad Before  & Tracking Lost & 0.032 & 0.041&\textbf{0.031}  \\
          \quad\quad\quad\quad After  & - &0.075 & 0.055&\textbf{0.043}  \\
           \quad\quad\quad\quad Full & -&  0.060 &0.053 &\textbf{0.045}  \\
        \bottomrule
    \end{tabular}
    }
    \end{table}
Here, CubeSLAM and EM-Fusion fail to complete both sequences while VOOM completes only the Indoor loop but with heavy drift. We base our integration and comparison on running ORB-SLAM3 non-stop, and present quantitative and qualitative results in \autoref{lab_res} and \autoref{lab_traj_viz}. From \autoref{lab_res}, we observe that by integrating NeuSE-based camera pose constraints, ORB3 + Ours consistently produces more accurate camera pose estimates compared to ORB-SLAM3-only throughout both rounds before and after object changes, improving overall trajectory estimation accuracy.

For the Indoor loop, we observe that ORB3 + Ours successfully prevents the tracking failure (the spike in the lower-left corner of \autoref{lab_traj_viz}(b)) that originally occurs during the independent execution of ORB-SLAM3. Additionally, the improved alignment of the trajectory with the ground truth from ORB3 + Ours demonstrates our approach's ability to facilitate localization in a confined space with full \sethree~motion.

In the Freeform loop, as illustrated in the second column of \autoref{lab_traj_viz}, we observe improved alignment and reduced trajectory drift from ORB3 + Ours compared to the ground truth. This improvement is particularly evident at the sequence's left and right ends, specifically when (1) navigating through the door after the corridor blackout, where camera tracking is lost, and (2) approaching the starting point with a sharp turn. NeuSE's effectiveness in robust relocalization, particularly in data association across diverse viewing angles, and our purely geometric approach, which does not rely on pixel information susceptible to motion blur, together contribute to this improvement. Hence, our strategy effectively aids localization in a larger motion space and supports reliable relocalization during potential blackout periods.

\begin{figure}[!htb]
    \centering
    \includegraphics[width=\linewidth]{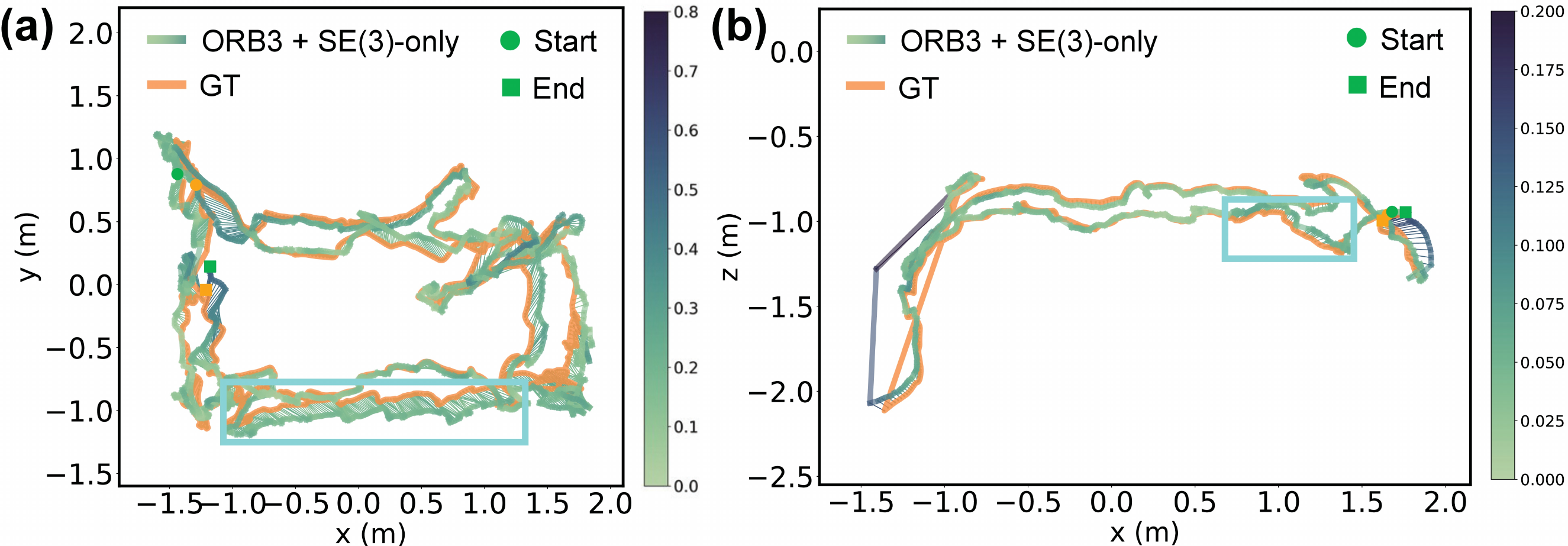}
    \centering
    \caption{\textbf{Trajectory estimation against ground truth (GT) for NeuSE's \sethree-only variant on handheld sequences.} The \sethree-only variant offers less noticeable improvements in camera pose estimation accuracy compared to the full \sethree-~and scale-equivariant version. Specifically, unlike NeuSE, this variant fails to process observations of the white square jar, an instance intentionally included  under the trash bin category with a similar shape but significantly smaller. The jar is used to test the approach's ability to handle objects of varying sizes. This failure to correctly handle objects of out-of-distribution sizes results in persistent drift in areas where the jar is observed, as highlighted by the cyan blocks in \textbf{(a)} for the Indoor loop and \textbf{(b)} for the Freeform loop.}
    \label{se3_lab_viz}
\end{figure}

For both loops, the \sethree-only variant improves the original ORB-SLAM3 results, but only slightly in terms of final trajectory accuracy, as indicated by the lower RMSE of ATE values. Notably, in the Freeform loop, the performance of the \sethree-only variant actually worsens during the first loop, prior to any object changes, as reflected by the RMSE of ATE values.  This degradation stems from its inability to effectively process the newly added white square jar, which is significantly smaller than the typical trash bin sizes seen during training. The failure to handle this object introduces a lingering effect on the localization performance, adversely impacting the quality of camera pose constraints generated jointly from the jar and  other objects in the frame. This results in a more remarkable drift in the trajectory estimate wherever the jar appears, as highlighted by the two cyan blocks in \autoref{se3_lab_viz}. In contrast, no significant drift is observed in these regions when using NeuSE with full scale equivariance (\autoref{lab_traj_viz}(c)), further demonstrating the capability of scale equivariance to generalize to objects of varying sizes.

\section{Spatial understanding with NeuSE: Change-aware object-centric mapping} 
In this section, we demonstrate how our approach constructs and maintains object-centric maps by utilizing equation~\eqref{se3} to generate consistent object reconstructions from each latent embedding. We further show how these maps can be updated to reflect changes in object layouts.

\subsection{Multi-view multi-scale object mapping}
Here, we showcase the robustness of NeuSE-based object mapping under varying viewing angles, object sizes, and shapes across categories. 

\begin{figure}[!htb]
    \centering
    \includegraphics[width=\linewidth]{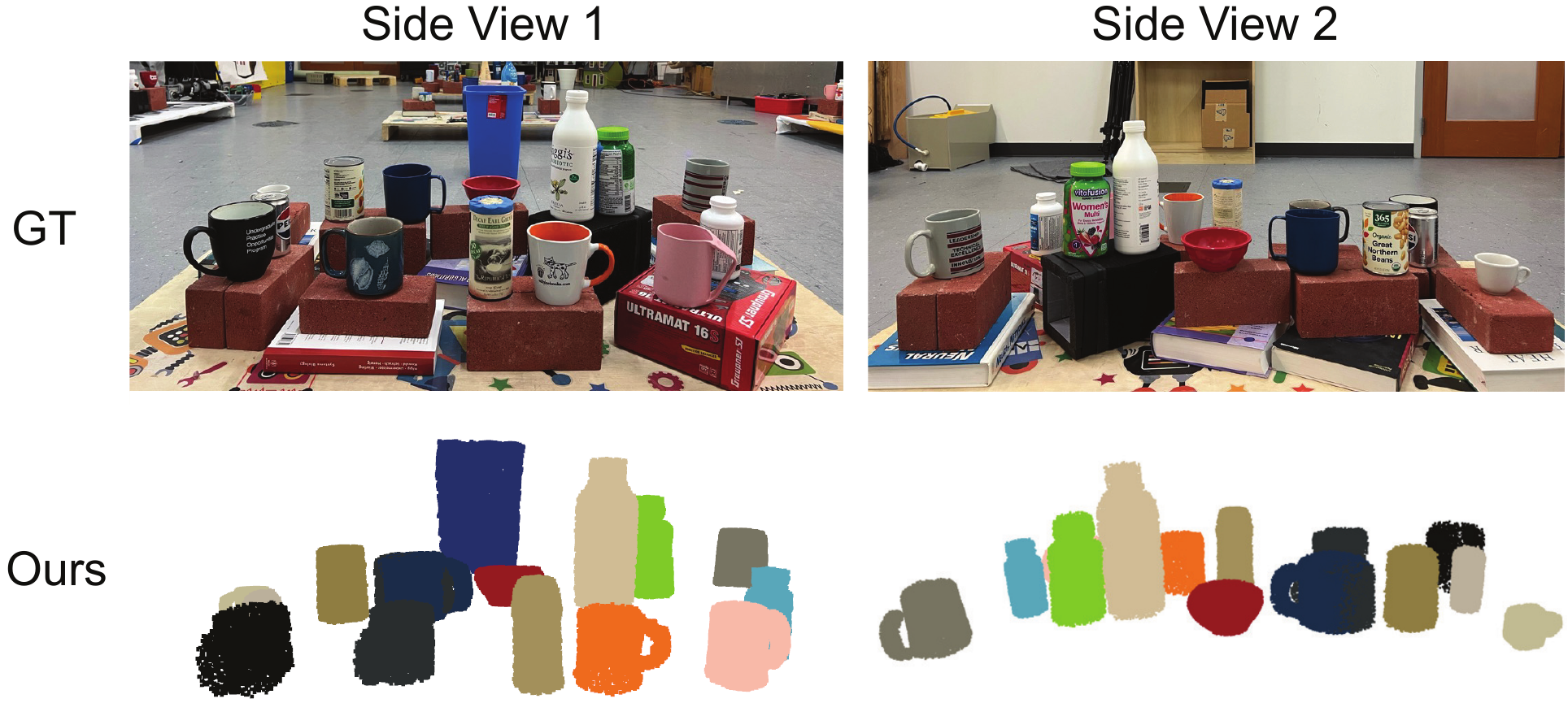}
    \caption{\textbf{Consistent multi-view object reconstruction. }We present our reconstruction (Ours) of the table with the highest object density across the four real-world testing sequences, including the trash bin visible in the first side view. Across all the five object categories present, the reconstruction maintains shape and spatial consistency, closely aligning with the ground truth (GT), as confirmed by the two side views.}
    \label{multi_view_reconstruction}
\end{figure}

In \autoref{multi_view_reconstruction}, we present the multi-view reconstruction of the leftmost table (referred to as table 1 in \autoref{holodeck_reconstruction}(a.1) and (d)) from the controlled environment, along with the blue trash bin placed just behind it (see ``Side View 1'' in \autoref{multi_view_reconstruction}). This table is selected as a representative example because it contains objects from all the five categories and is one of the tables with the highest object density across all four real-world sequences, thereby introducing more occlusion patterns. Despite the cluttered layout, the reconstructions demonstrate decent consistency with the ground truth from both side views. NeuSE successfully recovers the full shape of each object and accurately estimates their locations, yielding a spatial layout that closely aligns with the actual physical arrangement.

\begin{figure}[!htb]
    \centering
    \includegraphics[width=\linewidth]{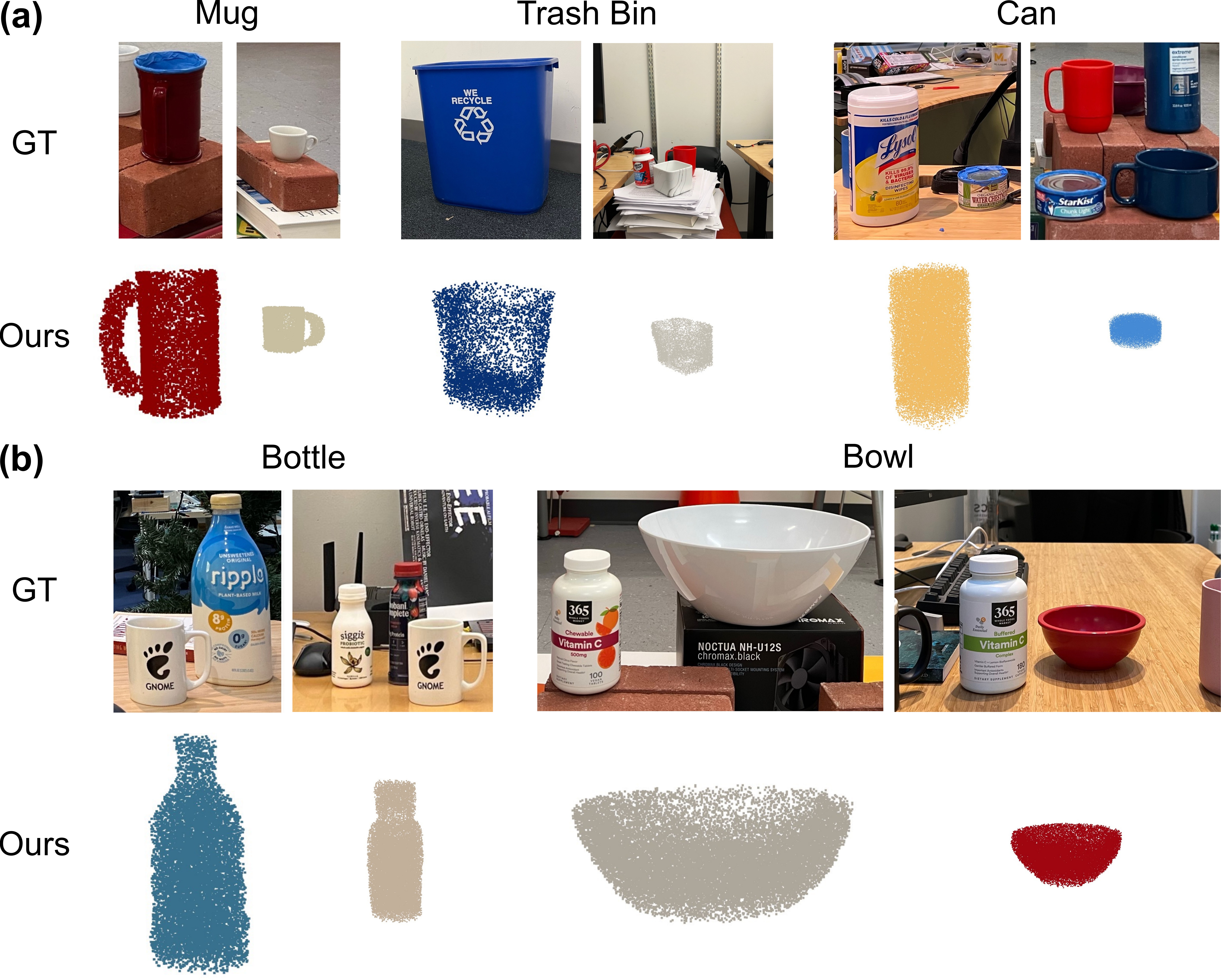}
    \caption{\textbf{Object reconstruction across scales}.
We present NeuSE’s reconstructions of the largest and smallest objects from real-world sequences across each of the five categories: mug, bottle, bowl, can, and trash bin. Reconstructions are visualized to preserve the approximate relative size ratios of the objects within each pair. \textbf{(a)} For object pairs without reference for relative scale ratio in ground truth images, the rough volumes from left to right are:  beer mug (600~mL) and espresso cup (60~mL);  trash bin (26.5~L) and small jar (0.3~L); yellow can (494~mL) and blue can (188~mL). From \textbf{(a)} through \textbf{(b)}, despite the considerable intra-category scale variation, NeuSE maintains consistent reconstruction performance across both categories and object sizes.}
    \label{object_scale_comparison}
\end{figure}
\begin{figure}[!htb]
    \centering
    \includegraphics[width=\linewidth]{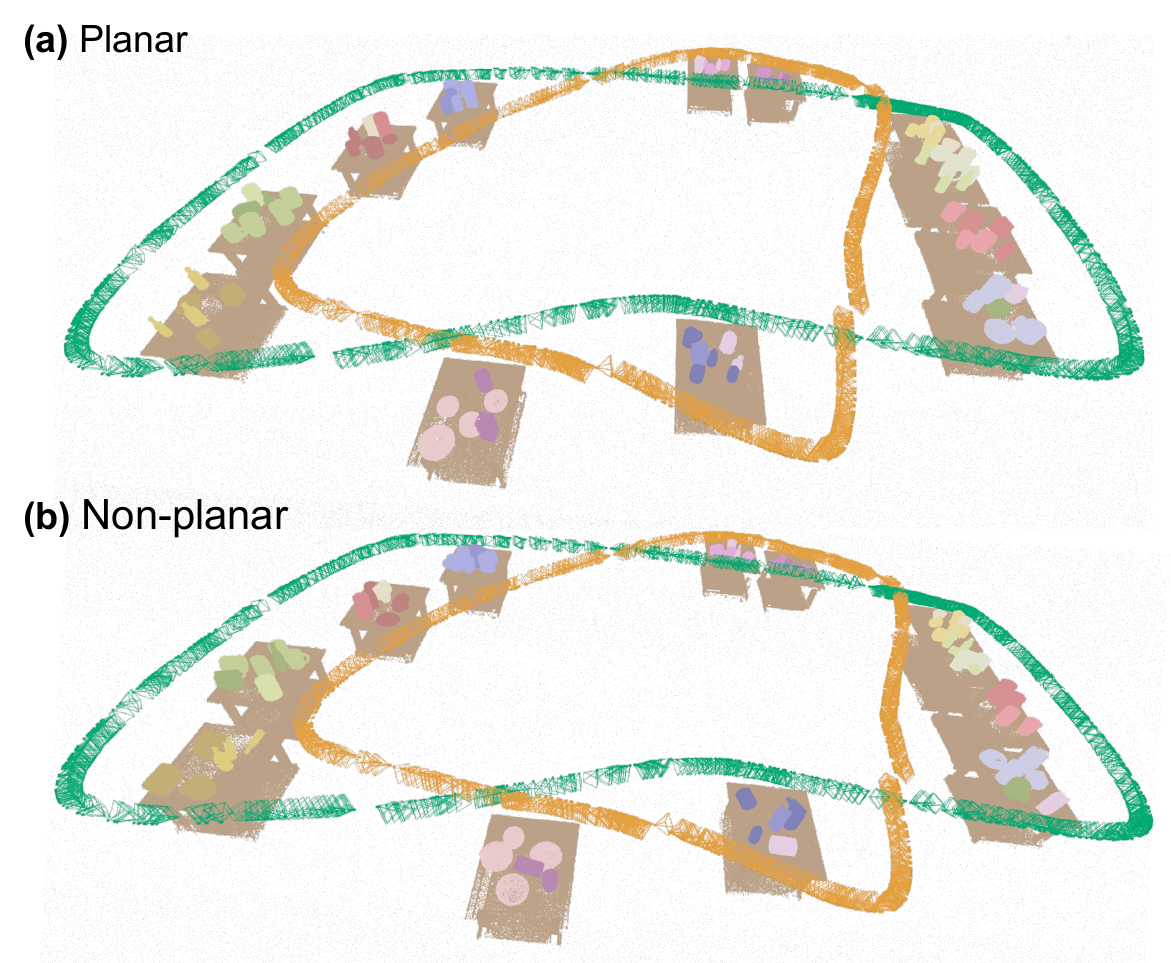}
    \caption{\textbf{Complete object reconstructions of the synthetic multi-category sequences with planar and non-planar object layouts}. \textbf{(a)} Planar layout. \textbf{(b)} Non-planar layout. Tables are rendered for visual clarity, with points back-projected to the world using camera pose estimates from NeuSE-predicted constraints, demonstrating the effectiveness of our localization strategy.}
    \label{sim_reconstruction}
\end{figure}

In terms of NeuSE's ability to generalize across object scales, \autoref{object_scale_comparison} displays the reconstructions of the largest and smallest objects from each of the five categories, mug, bottle, bowl, can, and trash bin in our real-world sequences. The reconstructions are rendered to approximate their actual size ratios within each object pair. For object pairs in \autoref{object_scale_comparison}(a), where no reference objects are present in the ground truth images for relative scale comparison, we provide the approximate volume details: beer mug (600~mL) and espresso cup (60~mL), trash bin (26.5~L) and small bin (0.3~L), yellow plastic can (494~mL) and tuna can (188~mL). Despite the considerable scale variations listed here, NeuSE effectively handles significant differences in object shapes and sizes, consistently demonstrating reliable reconstruction performance across these multi-category, multi-scale objects in real-world sequences.

\subsection{Object-centric mapping with changes}
Since there are no suitable object-based SLAM pipelines for direct comparison of mapping with temporal scene changes, we use the object-level mapping method with online change detection, panoptic multi-TSDFs (PMT) by ~\citet{9811877}, as our baseline. We feed PMT with our trajectory estimates that have the lowest RMSE of ATE values and compare the change detection results for synthetic and real-world sequences.

We quantify the performance of our system and PMT in \autoref{change_res} by comparing the numbers of correctly detected changes (true positives, TP), falsely detected changes (false positives, FP), and undetected changes (false negatives, FN). For the synthetic data, we choose the planar sequences of the Mug + Bottle and Multi-category object settings as they exhibit more typical upright object layouts compared to their \sethree~counterparts.  We further calculate precision (Pr) and recall (Re) rates based on these numbers. The results show that our system correctly detects most of the changes for both synthetic and real-world data, while PMT produces several false positives and false negatives due to localization errors and the inability to reason holistically over partial observations.
\begin{table}
    \centering
    \caption{Change detection results on the synthetic and real-world sequences. Best results are marked in bold.}
    \label{change_res}
    \centering
    \resizebox{\linewidth}{!}{
    \begin{tabular}{@{}lccccc@{}}
        \toprule
          &  TP & FP & FN &  Pr & Re \\
         \midrule
          \textbf{Synthetic: Mug + Bottle}\\
          \quad\quad\quad\quad PMT  &  7 & 2 & 2 & 77.8\% & 77.8\%\\
          \quad\quad\quad\quad Ours & 9 & 0 & 0 & \textbf{100.0\%} & \textbf{100.0\%}  \\
               \midrule
          \textbf{Synthetic: Multi}\\
          \quad\quad\quad\quad PMT  &  11 & 2 & 0 & 84.6\% & \textbf{100.0\%}\\
          \quad\quad\quad\quad Ours & 11 & 0 & 0 & \textbf{100.0\%} & \textbf{100.0\%}  \\
          \midrule
          \textbf{4-Round}\\
          \quad\quad\quad\quad PMT  & 14 & 0 & 2 & \textbf{100.0\%} & 87.5\% \\
          \quad\quad\quad\quad Ours & 16 & 0 & 0 & \textbf{100.0\%} & \textbf{100.0\%} \\
          \midrule
          \textbf{Triple-Infinity}\\
          \quad\quad\quad\quad PMT  & 14 & 2 & 0 & 87.5\% & \textbf{100.0\%} \\
          \quad\quad\quad\quad Ours &  14 & 1 & 0 & \textbf{93.3\%} & \textbf{100.0\%} \\
           \midrule
           \textbf{Indoor}\\
          \quad\quad\quad\quad PMT  & 9 & 2 & 0 & 81.8\% & \textbf{100.0\%} \\
          \quad\quad\quad\quad Ours &  9 & 0 & 0 & \textbf{100.0\%} & \textbf{100.0\%} \\
           \midrule
           \textbf{Freeform}\\
          \quad\quad\quad\quad PMT  & 5 & 0 & 3 & \textbf{100.0\%} & 62.5\% \\
          \quad\quad\quad\quad Ours &  8 & 1 & 0 & 88.9\% & \textbf{100.0\%} \\
        \bottomrule
    \end{tabular}
    }
\end{table}
Qualitatively, we present in \autoref{sim_reconstruction} reconstructions of all objects that have appeared in the synthetic multi-category planar and non-planar layouts, respectively. Figures~\ref{holodeck_reconstruction}--\ref{freeform_change_detection} display the object-centric maps and their updates from our method and PMT before and after changes for the tables in the controlled and uncontrolled settings, respectively. 
\subsubsection{Controlled environments: 4-Round and Triple-Infinity loops.}
\begin{figure*}[!htb]
    \centering
    \includegraphics[width=\linewidth]{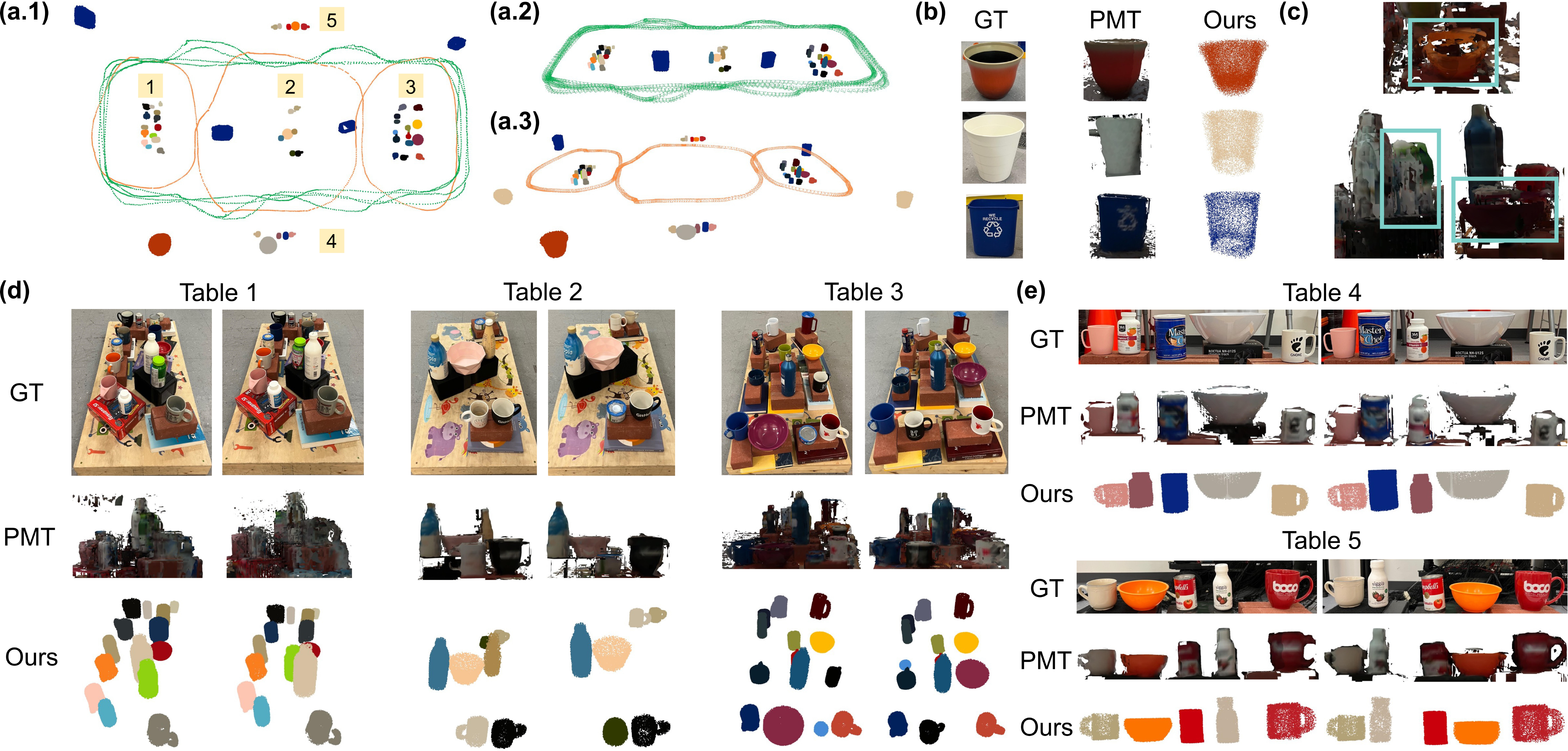}
    \caption{\textbf{Results of change-aware mapping for real-world controlled sequences.}  \textbf{(a)} \textbf{(a.1)} presents the reconstructed object-centric map with camera trajectory estimates, illustrating the spatial consistency between object reconstructions and actual camera motion. \textbf{(a.2)} and \textbf{(a.3)} display full object reconstructions and estimated trajectories for the \textcolor{green}{4-Round} and \textcolor{orange}{Triple-Infinity} sequences.
\textbf{(b)} Comparison of non-tabletop object reconstructions, i.e., various trash bins on the floor, across ground truth scenes (GT),  from Panoptic Multi-TSDF Mapping (PMT), and from our object-centric maps (Ours). \textbf{(d)--(e)} Reconstructed objects and layouts of all the five tables in the controlled environment, shown before (left) and after (right) changes, comparing GT, PMT, and Ours. Our approach provides a lightweight, object-centric map that better captures object changes and outperforms PMT in both reconstruction quality and change detection accuracy. \textbf{(c)} Failures in PMT’s change detection and reconstruction are highlighted in cyan blocks.  Top: The orange bowl on table 3 is falsely labeled as changed by PMT during the Triple-Infinity loop due to the little overlap between the nearly opposite camera views when the bowl is visible. Bottom: PMT fails to correctly detect the objects that switch locations, resulting in overlapping reconstructions of the white and green bottles (left) and the black mug in the red bowl(right).}
    \label{holodeck_reconstruction}
\end{figure*}

In \autoref{holodeck_reconstruction}, we present the full reconstructions of all the objects in the sequences of the controlled environment.  Our approach generates a lightweight, object-centric map that precisely captures changes (see \autoref{holodeck_reconstruction}(d) and (e) for the object-centric maps constructed for each table before and after changes). In contrast, PMT, being a traditional TSDF-based mapping technique, fails to perform accurate change detection and produces reconstructions with various defects. PMT is sensitive to limited viewing angle overlap, especially when coupled with localization errors, which can easily lead to false positive changes. This occurs when there is little overlap between the two partial object volumes due to localization drift, such as the orange bowl in the first row of \autoref{holodeck_reconstruction}(c), which is falsely marked as changed. In addition, PMT struggles to distinguish between objects that switch places due to its inability to perform full object shape comparison as NeuSE does. This is shown by the overlapping reconstructions of the white and green bottles, as well as the black mug in the red bowl, in the bottom row of \autoref{holodeck_reconstruction}(c).

\begin{figure*}[!htb]
    \centering
    \includegraphics[width=\linewidth]{figures/27.pdf} 
    \caption{\textbf{Overview of the reconstructed object-centric map and camera trajectory in the uncontrolled environment.} The reconstruction and the camera trajectory estimates cover two regions, the indoor office area and the open lab space, connected by a blackout corridor, and demonstrate spatial consistency both within each region and across them, well aligned with the overall floor plan. \textbf{(b)--(e)--(f)} show the reconstructed object  layout and camera trajectory estimates for the \textcolor{green}{Indoor} and \textcolor{orange}{Freeform} loops, with a gray segment in the corridor to be the ``blackout'' part of the Freeform loop, where no objects of interest are available and ORB-SLAM3 tracking fails to recover the accurate trajectory. \textbf{(a)} shows the  reconstructions for the unchanged objects  while \textbf{(d)} presents the full estimated camera trajectory with the object layout in the open lab space. For the indoor office area, \textbf{(c)} outlines the Indoor trajectory estimate with the reconstructed object-centric map, and \textbf{(g)} exhibits the unchanged objects in the indoor office room.}
    \label{lab_traj_reconstruction}
\end{figure*}

\begin{figure*}[!htb]
    \centering
    \includegraphics[width=\linewidth]{figures/28.pdf}
    \caption{\textbf{Object reconstruction and change detection in the indoor office area.} The map progression of the five tables is shown before and after the object changes (the left figure is the original state of the table, and the right figure is the new state of the table). Our approach maintains a clean and up-to-date object-centric map with all changes correctly detected. While for PMT, as its TSDF-based mapping method is limited by the parts of objects it has observed and the quality of camera trajectory estimates, the reconstructions can be incomplete and noisy, for example, the map for table 1 before changes. Moreover, as highlighted by the cyan blocks, PMT fails to remove the orange bowl on table 3 (right figure) after it is moved to the right side of the table. PMT also leaves a residual volume of the large blue bottle on table 5 at its previous location, and makes it overlap with the blue can after it is moved to the right side of the windowsill.}
    \label{indoor_change_detection}
\end{figure*} 

\begin{figure*}[!htb]
    \centering
    \includegraphics[width=\linewidth]{figures/29.pdf}
    \caption{\textbf{Object reconstruction and change detection in the open lab space.} The map transition of the five tables is shown before and after the object changes (the left figure is the original state of the table, and the right figure is the new state of the table). Our method accurately captures object changes and recovers complete object shapes across categories. In contrast, as highlighted by the cyan blocks, PMT produces incomplete reconstructions, such as the missing blue can on table 3, and fails to remove the dark bottle from table 1 after it is moved.}
    \label{freeform_change_detection}
\end{figure*}

\subsubsection{Uncontrolled environments: Indoor and Freeform loops.}
In \autoref{lab_traj_reconstruction}, we present the reconstructed object layouts and full trajectory estimates for the Indoor and Freeform sequences. The strong spatial consistency between the camera trajectory estimates and the object layouts, shown in the Indoor  (\autoref{lab_traj_reconstruction}(c), (f), and (g)) and Freeform (\autoref{lab_traj_reconstruction}(a), (b), and (d)--(f)) loops, demonstrates our approach’s ability to handle 3D motion and generate globally consistent trajectory estimates from surrounding object observations.

We also illustrate the spatial relationship between the open lab space (\autoref{lab_traj_reconstruction}(b)) and indoor office area (\autoref{lab_traj_reconstruction}(f)), connected by the blackout corridor (\autoref{lab_traj_reconstruction}(e)), in a floor plan style. Despite the tracking loss in the corridor, NeuSE-based relocalization allows the Freeform trajectory to align with the indoor space, achieving cross-region spatial consistency. This is proven by the alignment of the Freeform indoor trajectory (orange) with the indoor trajectory (green) in \autoref{lab_traj_reconstruction}(f), especially after the camera passes through the blackout period in the corridor and re-enters the office.

For object change detection in the Indoor (\autoref{indoor_change_detection}) and Freeform (\autoref{freeform_change_detection}) loops, we present the development of object maps before and after object changes for all the tables in both sequences. Along with the reconstructed objects, we also observe smooth and consistent TSDF reconstructions of several tabletop surfaces from PMT when using NeuSE-based camera pose estimates as the input for localization. Notably, accurate tabletop reconstructions are achieved for tables in both the indoor office area and the outer open lab space, which further validates the accuracy of our camera pose estimates.

We also highlight instances, marked in cyan blocks in Figures~\ref{indoor_change_detection} and \ref{freeform_change_detection}, where PMT struggles with change detection and reconstruction due to TSDF’s inherent limitations in shape reasoning and PMT’s sensitivity to localization errors. These artifacts include the residual orange bowl remaining in its original position after being moved to the right side of the table in table 3, as well as the duplicate volume of the large blue bottle in the middle after it is moved to the right of the windowsill in table 5 of \autoref{indoor_change_detection}. Furthermore, in \autoref{freeform_change_detection}, the blue can on table 3 is missing from the reconstruction, and there is a false negative detection of the dark yogurt bottle on table 1, which is not properly removed from the map.
\section{Limitations and future work}
Our evaluation on both synthetic and real-world sequences demonstrates NeuSE's effectiveness in object-based SLAM. With its \sethree~and scale equivariance, NeuSE facilitates spatial understanding across objects of varying shapes and sizes, effectively handling occlusion and varying viewing angles to achieve robust long-term scene consistency. However, there are a few limitations worth addressing.

First, NeuSE operates as a category-level representation, requiring category-specific training for each type of object before being applied to a scene. This limitation reduces its flexibility and poses challenges when scaling to larger environments, especially in ``open-set'' spatial understanding, where no prior knowledge of object categories is available. In future work, we aim to overcome this by moving beyond category-level features and incorporating representations based on canonical geometric shapes, such as spheres, cylinders, and cuboids. In this spirit, we hope to make NeuSE interpolable in the latent space, allowing it to approximate actual object shapes more effectively with combinations of canonical shapes, thereby enhancing NeuSE's capability to handle diverse object shapes in real-world scenarios.

Second, NeuSE performs best with high-quality partial object point clouds, which depend on reliable depth measurements and clean RGB segmentation masks. Few straightforward approaches to address this are to further train NeuSE with additional depth noise building on its demonstrated robustness and to leverage advances in the image segmentation community for higher-quality segmentation masks. Beyond this, we plan to extend NeuSE's \sethree~and scale equivariance to handle not only individual objects but also small clusters of objects, enhancing its ability to manage smaller objects where current segmentation methods may struggle.

Finally, NeuSE relies on prior information, such as ambiguity patterns (e.g., unambiguous, cylindrical, or cubical) and accurate semantic labels for effective object association. Preparing training data with these ambiguity labels requires additional annotation effort. Moreover, NeuSE's \sethree~equivariance is effective only when the latent codes belong to the same category, making it sensitive to semantic labeling errors. Introducing thresholding based on confidence scores for semantic labels could provide a simple yet effective solution to mitigate the impact of labeling inaccuracy. In future work, transforming NeuSE from a category-level into a shape-based representation, as mentioned earlier, could be another promising direction. Beyond learning the latent interpolation of different canonical shapes to approximate the shape of the query object, we also aim to reveal ambiguity patterns through latent interpolation from these canonical shapes. This approach could significantly reduce the effort required for labeling ambiguity patterns from labeling different shape instances to only a few canonical shapes, while also decreasing our dependence on semantic labels for object association.

Overall, our results demonstrate that NeuSE is a compact and effective object representation for object-based SLAM. NeuSE-based object SLAM provides a fundamental step in spatial understanding by localizing and constructing an object-centric map. A promising direction for future work is to explore how NeuSE, along with SLAM results, can be directly integrated into downstream tasks such as navigation and mobile manipulation. Compared to explicit spatial representations like point clouds or voxel grids, NeuSE offers a more semantically meaningful description of a scene based on its objects and their relationships, and can be efficiently queried to support navigation tasks, which favors holistic descriptions of the objects in the scene. Furthermore, NeuSE's ability to construct a lightweight change-aware object-centric map can be crucial in scenarios involving changes caused by external agents or the robot's interactions with its surroundings, such as for initializing and updating the scene representation each time the robot acts on an object during mobile manipulation tasks.

\FloatBarrier
\section{Conclusion} 
\label{sec:conclusion}
In this paper, we present NeuSE, a category-level neural latent embedding for objects, and demonstrate its ability to support object-based SLAM for consistent spatial understanding, even in the presence of long-term scene changes. NeuSE distinguishes itself from previous neural representations used in SLAM by allowing key SLAM components to be operating with a unified set of inferred latent embeddings. Camera pose constraints between objects can be derived directly from the corresponding latent representations, which are associated through scale, shape, and spatial information embedded in the representations. Furthermore, with its \sethree~and scale equivariance, NeuSE is able to handle a broader range of object shapes and sizes during inference compared to those seen during training, while also constructing flexible object-centric maps that can easily accommodate long-term scene changes. Our evaluation on synthetic and real-world data, in both controlled and uncontrolled settings, demonstrates the feasibility of our approach for change-aware localization and object-centric mapping, whether functioning independently or as a complement to traditional SLAM pipelines.
\section*{Acknowledgments}
The authors thank Shichao Yang for discussions and guidance in setting up CubeSLAM for the real-world testing sequences, as well as Tim Magoun and  Gabriel Margolis for help with data collection. 

\section*{Funding}
This work was supported by ONR MURI grant N00014-19-1-2571 and ONR grant N00014-18-1-2832. 

\section*{Declaration of conflicting interests}
The authors declare no potential conflicts of interest with respect to the research, authorship, and/or publication of this article.

\section*{ORCID iDs}
\noindent Jiahui Fu: \href{https://orcid.org/0000-0001-8298-6659}{0000-0001-8298-6659}\\
Yilun Du: \href{https://orcid.org/0000-0001-6792-5946}{0000-0001-6792-5946}\\
Kurran Singh: \href{https://orcid.org/0000-0002-8097-7940}{0000-0002-8097-7940}\\
John J. Leonard: \href{https://orcid.org/0000-0002-8863-6550}{0000-0002-8863-6550}

\bibliographystyle{SageH}
\bibliography{references}
\raggedcolsend
\raggedbottom
\clearpage
\raggedend
\makeatletter
\@left@column@spreadfalse
\@right@column@spreadfalse
\makeatother
\appendix
\section*{Appendix}
\setcounter{subsection}{0}
\renewcommand{\thesubsection}{A\arabic{subsection}}
\setcounter{table}{0}
\renewcommand{\thetable}{A\arabic{table}} 
Here, we provide additional details on our experiment setup and mapping procedures.
\subsection{Experimental details}
\textbf{Model details.}  We provide the network architecture of our encoder and decoder in Tables~\ref{encoder} and \ref{decoder}, respectively, which are derived from Neural Descriptor Fields~\citep{9812146}  using the VNNLinear, VNNResnetBlock, VNLeakyReLU blocks introduced in Vector Neurons~\citep{Deng_2021_ICCV}. Here, $z_{dim}$ refers to the size of the NeuSE latent code, ${\bf z}\in \mathbb{R}^{z_{dim}\times 3}$. \newline 
\textbf{Training details.} During training, we randomly drew 500 sample points from the observed partial point cloud for each sample to be fed into the encoder network. For training with $L_{occ}$, the query point size was 750 consisting of half object points and half off-the-object points from the given object model. We set the dimension of the latent code $\bf z$ to be $z_{dim}=512$ and the weight coefficient for each loss term to be one among $L_{occ}$, $L_{transform}$, $L_{scale}$, and $L_{shape}$. We trained separate models for each category: mug, bowl, bottle, trash bin, and can. We found that results were robust across a range of $z_{dim}$ values.

\par\smallskip
\noindent\begin{minipage}{\linewidth}
\begin{minipage}[t]{0.48\linewidth}
\centering
\captionsetup{position=top,hypcap=false}
\captionof{table}{Encoder architecture.}
\label{encoder}
\small
\begin{tabular}{c}
    \toprule
    \textbf{Layer type}\\
    \midrule
    VNLinear(128,256)\\
    VNLinear(256,128)\\
    VNLinear(256,128)\\
    VNLinear(256,128)\\
    VNLinear(256,128)\\
    VNLinear(256,128)\\
    Meanpool\\
    VNLinear(128,$z_{dim}$)\\
    $\bf z$ $\leftarrow$ Encode\\
    \bottomrule
\end{tabular}
\end{minipage}
\hfill
\begin{minipage}[t]{0.48\linewidth}
\centering
\captionsetup{position=top,hypcap=false}
\captionof{table}{Decoder architecture.}
\label{decoder}
\small
\begin{tabular}{c}
    \toprule
    \textbf{Layer type}\\
    \midrule
    VNLinear($z_{dim}$,$z_{dim}$)\\
    Linear(2*$z_{dim}$+1,128)\\
    ResnetBlockFC(128)\\
    ResnetBlockFC(128)\\
    ResnetBlockFC(128)\\
    ResnetBlockFC(128)\\
    ResnetBlockFC(128)\\
    Linear(128,1)\\
    Sigmoid\\
    \bottomrule
\end{tabular}
\end{minipage}
\end{minipage}
\par\smallskip

\subsection{Reconstruction and update of the object-centric map}
In this section, we elaborate on our choices and procedures in building and maintaining the object-centric map, which we adopt to address the noise in real-world data.

Our proposed approach depends on correct object segmentation masks to produce effective latent codes for objects. To avoid potential failures from false object latent codes, given the uncertainty of off-the-shelf object detectors and depth cameras, we only initialize a new object instance if it has been recognized robustly by the depth camera and the object detector, for example, an object sufficiently close to the camera with a dense point cloud. Here in our experiment for object instantiation, we only considered objects that were within 2.5~m away from the depth camera or with their pixel-level segmentation mask size above 4,000.  Ultimately, a new object is instantiated after it has been recognized as a ``new" object three consecutive times during data association. We used a pre-trained Mask R-CNN~\citep{8237584} model to obtain segmentation masks.

For map maintenance and update, we conduct the removal of ``residual" object instances based on the bounding box of each full object reconstruction. This ensures reduced redundant object instantiation from partial observations and avoids overlapping object reconstructions for objects that are detected to have changed locations.

We can use our learned decoding function to obtain a reasonable full-shape prediction from partial observations, for example, an upright bottle with its bottom half obscured.  We are, therefore, able to obtain the 3D bounding box of each observed object and conduct the object removal procedure as follows: (1) If a small object's bounding box has a high overlap with a large one (0.95 in our case), this small object is deemed as a partial instance belonging to the large one and will be removed; (2) A changed object's bounding box intersects (we set it as 20\% of its bounding box volume) with an older object, meaning the older object should no longer be in its original place. We hence remove the older object from the map and update the changed object to its new position.

\end{document}